%% file: metrics.tex
\documentclass[12pt]{article}
\pdfoutput=1

\usepackage{cite}
\usepackage{setspace}
\usepackage{xspace}
\usepackage{color}
\usepackage{soul}
\usepackage{amsmath, amsfonts, amssymb, amsthm, upgreek, amstext, bm, mathtools, mathrsfs, stmaryrd}
\usepackage[font=footnotesize,labelfont=bf]{caption}
\usepackage{slashed}
\usepackage{subfig}
\usepackage{multirow}
\usepackage{array} 
\usepackage{bbold}
\usepackage{physics}
\usepackage{booktabs}
\usepackage{fontawesome}
\usepackage[english]{babel}
\usepackage[utf8]{inputenc}
\usepackage[colorlinks=true,linkcolor=rossos,anchorcolor=black,citecolor=blue,filecolor=cyan,menucolor=red,runcolor=filecolor,urlcolor=blue,bookmarks=true,bookmarksnumbered=true]{hyperref}
\usepackage{cleveref}
\usepackage[dvipsnames]{xcolor}
\usepackage{doi}
\usepackage{multicol}

\usepackage{graphicx} 
\usepackage{pdflscape}

\newcommand{\keywords}[1]{\small\textbf{Keywords:} #1}

\definecolor{rosso}{cmyk}{0,1,1,0.4}
\definecolor{rossos}{cmyk}{0,1,1,0.55}
\definecolor{rossoc}{cmyk}{0,1,1,0.2}
\definecolor{blu}{cmyk}{1,1,0,0.3}
\definecolor{blus}{cmyk}{1,1,0,0.6}
\definecolor{bluc}{cmyk}{1,1,0,0.1}
\definecolor{verde}{cmyk}{0.92,0,0.59,0.25}
\definecolor{verdec}{cmyk}{0.92,0,0.59,0.15}
\definecolor{verdes}{cmyk}{0.92,0,0.59,0.4}

\newcommand{\be}{\begin{equation}}
\newcommand{\ee}{\end{equation}} 
\newcommand{\bry}{\begin{array}}
\newcommand{\ery}{\end{array}} 
\newcommand{\dst}{\displaystyle} 
\newcommand{\bit}{\begin{itemize}} 
\newcommand{\eit}{\end{itemize}} 
\newcommand{\ben}{\begin{enumerate}} 
\newcommand{\een}{\end{enumerate}}
\newcommand{\Reals}{\mathbb{R}}

\DeclareMathOperator*{\argmin}{arg\,min}

\begin{document}

\title{{\bf \vspace{-1cm}Refereeing the Referees:\\ Evaluating Two-Sample Tests for Validating Generators in Precision Sciences\vspace{5mm}}}

\author{
Samuele Grossi$^{a,b}$, Marco Letizia$^{b,c}$, and Riccardo Torre$^{a,b}$\\ \\
{\small\emph{$^a$ Department of Physics, University of Genova, Via Dodecaneso 33, I-16146 Genova, Italy}}\\
{\small\emph{$^b$ INFN, Sezione di Genova, Via Dodecaneso 33, I-16146 Genova, Italy}}\\
{\small\emph{$^c$ MaLGa-DIBRIS, University of Genova, Via Dodecaneso 35, I-16146 Genova, Italy}}\\
}

\maketitle

\begin{abstract}
    We propose a robust methodology to evaluate the performance and computational efficiency of non-parametric two-sample tests, specifically designed for high-dimensional generative models in scientific applications such as in particle physics.
    The study focuses on tests built from univariate integral probability measures: the sliced Wasserstein distance and the mean of the Kolmogorov-Smirnov statistics, already discussed in the literature, and the novel sliced Kolmogorov-Smirnov statistic. These metrics can be evaluated in parallel, allowing for fast and reliable estimates of their distribution under the null hypothesis.
    We also compare these metrics with the recently proposed unbiased Fréchet Gaussian Distance and the unbiased quadratic Maximum Mean Discrepancy, computed with a quartic polynomial kernel.
    We evaluate the proposed tests on various distributions, focusing on their sensitivity to deformations parameterized by a single parameter $\epsilon$. Our experiments include correlated Gaussians and mixtures of Gaussians in 5, 20, and 100 dimensions, and a particle physics dataset of gluon jets from the JetNet dataset, considering both jet- and particle-level features.
    Our results demonstrate that one-dimensional-based tests provide a level of sensitivity comparable to other multivariate metrics, but with significantly lower computational cost, making them ideal for evaluating generative models in high-dimensional settings. 
    This methodology offers an efficient, standardized tool for model comparison and can serve as a benchmark for more advanced tests, including machine-learning-based approaches.
\end{abstract}

\vspace{5mm}


\newpage
\tableofcontents



\newpage
\section{Introduction}
In recent years, generative models have become a powerful tool across a wide range of scientific and industrial fields, enabling the generation of high-dimensional synthetic data. Broadly, the applications of generative models can be divided into two categories, depending on the knowledge of the underlying model. The first category primarily includes industrial applications, where the underlying model is often unknown and only a set of data, typically large, but not always, is available for training. In contrast, the second category is more common in scientific domains, such as Particle and Astroparticle Physics, where a theoretical hypothesis for the underlying model is available, or, at least, synthetic data can be generated according to such hypothesis, for instance through Monte Carlo techniques. In this second case, generative models are frequently used to replace existing generators to improve computational efficiency, portability, or scalability. Validation poses critical challenges in both categories: in the first, precision is limited by the quality and quantity of available data, while in the second, the known underlying model offers a benchmark of potentially much higher precision, for instance through an arbitrarily small statistical uncertainty. This demands generative models capable of achieving comparable levels of accuracy, especially in fields where simulations require exact modeling of correlations and higher-order effects. The design and validation of such models fit within the paradigm of ``precision Machine Learning'', where high standards for fidelity and efficiency are critical.

The validation of generative models has become increasingly important with the rise of deep learning, leading to the development of numerous approaches for assessing the quality of generated data and optimizing model parameters \cite{heusel2017gans,karras2019style,borji2022pros}. However, many existing methods lack a rigorous statistical foundation, making it difficult to provide robust and reliable evaluations, particularly in high-stakes scientific applications. As already mentioned, this challenge is especially relevant in scientific domains that rely heavily on simulations, such as High Energy Physics (HEP). Generative models are indeed attracting considerable attention in HEP, where model-based Monte Carlo simulations are expected to become a serious computational bottleneck in the near future \cite{Aberle:2749422,Software:2815292,CERN-LHCC-2020-015}. Therefore, there is an ongoing effort to design generative models to take over, at least in part, the traditional simulation pipeline. The high precision required in HEP, where accurate modeling of features, correlations, and higher-order moments is essential, demands a robust performance assessment of generative models. Addressing this need demands advanced statistical techniques that can efficiently scale to high-dimensional data.

A natural statistical framework to evaluate generative models is two-sample hypothesis testing, which aims to determine whether two independent samples are drawn from the same distribution. These tests fall into two categories: \emph{parametric} and \emph{non-parametric} tests. Parametric tests, such as the (log-)likelihood ratio (LLR) test \cite{neyman1933ix}, identified by the Neyman-Pearson lemma as the most powerful for simple hypotheses, rely on specific assumptions about the underlying data distributions, like normality, and are highly effective when these assumptions hold.
In contrast, non-parametric tests make no such assumptions, providing greater robustness at the cost of a potentially lower power.
Although numerous non-parametric tests exist for both one- and multi-dimensional data, few have known asymptotic distributions under the null hypothesis. Moreover, most are not computationally efficient in high-dimensional settings, where the curse of dimensionality poses a significant challenge.

Machine learning (ML) techniques, particularly classifiers that approximate the LLR or a related function, have proved promising in designing two-sample tests for evaluating generative models in HEP \cite{DAgnolo:2018cun,DAgnolo:2019vbw,Krause:2021ilc,Letizia:2022xbe,Grosso:2023scl}. In particular, the model presented in Ref~\cite{DAgnolo:2018cun} and further developed and detailed in Ref.s~\cite{DAgnolo:2019vbw,Letizia:2022xbe,Grosso:2023scl} allows for a statistically robust evaluation as a goodness-of-fit test and has been recently proposed for evaluating of generative models \cite{Cappelli:nplm2024}.

However, assessing the performance of these classifiers remains challenging. Many generative models developed for HEP have not yet achieved high accuracy beyond the first few moments (e.g. mean and covariance), leading to tests that may be overly sensitive to minor, less significant differences. As a result, interpreting strong rejections of models in these tests can be difficult. This highlights the need for a set of robust, simple, and interpretable two-sample tests that can serve as benchmarks for evaluating more advanced ML-based approaches. Currently, such benchmarks are lacking in the literature, and this paper aims to fill the gap.

We propose a robust methodology for evaluating two-sample tests, focusing on non-parametric tests based on univariate integral probability measures (IPMs). Our approach builds on two well-established methods for comparing one-dimensional, in short 1D, distributions, both based on the empirical Cumulative Distribution Function (eCDF): the Kolmogorov-Smirnov (KS) test \cite{an1933sulla,smirnov1948table} and the Wasserstein (W) distance \cite{peyre2019computational,villani2021topics}. We extend these to higher dimensions using three key test statistics: the sliced-Wasserstein (SW) distance, the mean of the KS statistics computed on each 1D marginal distribution, and a sliced version of the KS statistic. These tests are computationally efficient, as they can be evaluated in parallel, allowing for a fast and reliable estimate of their distribution under the null hypothesis. 

For comparison with existing methods, we include in our analysis the recently proposed unbiased Fréchet Gaussian Distance (FGD)\cite{binkowski2018demystifying,chong2020effectively} and the unbiased quadratic Maximum Mean Discrepancy (MMD)\cite{gretton2006kernel,gretton2012kernel}, computed using a quartic polynomial kernel. Additionally, when the underlying densities are known, we use the LLR test as a baseline to assess the effectiveness of the other tests.

We evaluate the proposed tests on a diverse set of scenarios. First, we use toy distributions with known probability density functions (PDFs), specifically correlated Gaussians and mixtures of Gaussians in 5, 20, and 100 dimensions. These controlled benchmarks allow us to fully characterize the problem and compare the performance of the proposed non-parametric tests against the exact log-likelihood ratio test. Next, we apply our methodology to a particle physics dataset of gluon jets from the JetNet dataset~\cite{Kansal_JetNet_A_Python_2023}, which includes both jet- and particle-level features. This real-world dataset enables us to test the robustness and versatility of our approach in a practical case where the underlying PDF is unknown and the data is finite.

Our methodology proceeds as follows. We begin by fixing a reference distribution for each dataset. We then introduce well-defined deformations, parameterized by a single variable, $\epsilon$, to generate alternative hypotheses. Next, we estimate the distribution of each test statistic under the null hypothesis by sampling, or resampling in the case of finite data, only from the reference distribution. Finally, we assess the sensitivity of the tests by testing the alternative hypothesis versus the reference one, and determining the minimum value of $\epsilon$ that leads to the rejection of the null hypothesis at a given Confidence Level (CL).\footnote{This can be phrased as computing the upper value of $\epsilon$ excluded at a given CL.} All results are accompanied by uncertainty estimates.

Our results show that 1D-based tests generally perform comparably to the recently proposed multivariate metrics while offering significantly lower computational costs, particularly in high-dimensional settings and with large sample sizes. This makes these tests especially suitable for evaluating generative models, where both computational efficiency and accuracy are critical. We recommend these metrics for practitioners who require efficient and reliable methods for model selection. Furthermore, our proposed methodology provides a standardized tool for comparing various approaches to two-sample testing, including newer classifier-based methods.

The paper is organized as follows. In Section \ref{sec:2st}, we introduce two-sample testing and its connection to statistical goodness-of-fit inference. We also explore different applications, such as comparing numerical samples, generators, or a combination of the two. In Section \ref{sec:teststat}, we describe the various test statistics considered in this paper. Section \ref{sec:method} details our methodology for evaluating two-sample tests, while Section \ref{sec:distrib} describes the distributions, datasets, and deformations used in the analysis. In Section \ref{sec:results}, we present the numerical results along with sample figures and tables, and in Section \ref{sec:conclusion}, we offer concluding remarks and discuss future directions. Appendix \ref{app:figures} and \ref{app:tables} provide additional material, including a full list of figures and tables. 

Code, figures, and tables are available on GitHub \cite{GitHubGMetrics,GitHubGenerativeModelsMetrics,GitHubJetNetMetrics} and full results are available on Zenodo \cite{ZenodoGenerativeModelsMetrics,ZenodoJetNetMetrics}.


\newpage

\section{Two-sample hypothesis testing}\label{sec:2st}

Given two random variables $x$ and $y$, defined on a space $\mathcal{X}\subseteq\Reals^d$, let us consider two samples $X=\{x_i\}$, where $i=1,\ldots,n$, and $Y=\{y_j\}$, where $j=1,\ldots,m$. Using capital Latin letters $I, J = 1, \dots, d$ to denote the components of $d$-dimensional vectors, we can indicate with $x_{iI}$ the scalar value of the $i$-th element of sample $X$ along the $I$-th dimension. We also assume that the two samples are independent and identically distributed according to the distributions $p$ and $q$, respectively.

Two-sample testing aims at determining whether the null hypothesis at the population level,
\be\label{eq:null_hyp}
H_0: p = q,
\ee
can be rejected based on finite data. The alternative hypothesis is the negation of the null:
\be\label{eq:alt_hyp}
H_1: p \neq q.
\ee
The test proceeds by selecting a \emph{test statistic}, $t:(\mathcal{X})^n \times (\mathcal{X})^m \rightarrow \Reals$, and calculating its value on the observed data:
\be\label{eq:t_obs}
t_{\rm obs} = t(X, Y).
\ee
To define a binary test, the observed value of the test statistic is compared to a threshold, $t_\alpha$, where
\be\label{eq:pvalue}
\alpha = P(t \geq t_\alpha | H_0) = \int_{t_\alpha}^\infty f(t | H_0) dt,
\ee
with $f(t|H_{0})$ being the distribution of the test statistic under the null hypothesis $H_{0}$, represents a preselected probability of a type-I error, i.e.~the \emph{rate of false positives}. The null hypothesis is not rejected if $t_{\rm obs} < t_\alpha$. Alternatively, one can report the \emph{p-value}, defined as the probability of obtaining a test statistic as extreme as $t_{\rm obs}$ under $H_0$:
\be\label{eq:alpha}
\text{p}_{\rm obs} = P(t \geq t_{\rm obs} | H_0).
\ee
This quantity can also be mapped to a \emph{Z-score} using the quantile function of a standard Gaussian distribution:
\be\label{eq:zscore}
\text{Z}_{\rm obs}=\Phi^{-1}(1-\text{p}_{\rm obs}).
\ee
Given $\alpha$, the Neyman-Pearson construction provides a method to compare the performance of different tests. This consists in introducing a type-II error, or the \emph{rate of false negatives}
\be\label{eq:beta}
\beta = P(t < t_\alpha|H_{1}) = \int_{-\infty}^{t_\alpha} f(t|H_{1})dt,
\ee
where $f(t|H_{1})$ is the distribution of the test statistic under the alternative hypothesis. Then the \emph{power} of the test is defined as
\be\label{eq:power}
\textrm{power}=P(t \geq t_\alpha|H_1)=1-\beta.
\ee
The best test is usually chosen as the one with the highest power for a given $\alpha$, i.e., the one with the smallest rate of false negatives at fixed rate of false positives.

To compute the quantities in \cref{eq:alpha,eq:zscore,eq:pvalue,eq:beta,eq:power}, the distributions of the test statistic under both the null and alternative hypotheses, $f(t|H_0)$ and $f(t|H_1)$, must be known or estimated. In some cases, analytical approximations for finite sample sizes or asymptotic behavior can be used, as shown in Refs.\cite{an1933sulla,massey1951kolmogorov,smirnov1948table} for the KS test, and in Refs.\cite{Wilks1938,Cowan:2010js} for likelihood ratio-based tests. In general, however, non-parametric testing often relies on empirical estimates.

We identify two common scenarios:
\begin{itemize}
    \item The first case occurs when only two finite samples, $X$ and $Y$, representing two populations, are available. This scenario is common in the statistical literature on two-sample testing. To estimate $f(t|H_0)$ and $f(t|H_1)$, resampling methods, also known as randomization tests, such as permutation and bootstrap tests, are often used \cite{tibshirani1993introduction,edgington2007randomization,wasserman2013all}.

    \item In the second case, one or both distributions, while not known analytically, can still be sampled from. In other words, we have access to a \emph{generator} for at least one of the distributions. This scenario commonly arises when evaluating the fidelity of a generator against a \emph{reference} dataset, or when testing the compatibility of a dataset with a theoretical model represented by a Monte Carlo generator. Another common scenario involves comparing two data-generating methods, such as a highly accurate but computationally expensive Monte Carlo method against a faster but less accurate neural network-based model.
\end{itemize}

In practical applications, a generator might be available but limited by resource constraints, preventing the generation of sufficient data to estimate the distribution of the test statistic accurately. In such cases, a hybrid approach can be used: generating as much data as possible and then applying resampling methods to estimate the distribution of the test statistic.

In the standard two-sample testing framework, as summarized in this section, the two distributions in \cref{eq:null_hyp,eq:alt_hyp} are treated symmetrically, with no preference for $p$ or $q$. However, this methodology can also be adapted for goodness-of-fit testing by designating one distribution as the \emph{reference} one.
This approach is useful when one distribution is considered the true or best approximation and the other is tested against it. For instance, in the example above, the fast but less accurate generative model would represent the alternative, while the slower but more accurate physics-based Monte Carlo generator the reference. Similarly, a dataset, treated as the alternative distribution, can be tested against a generator based on a well-verified theoretical model, taken as reference.

In this context, the hypotheses can be rephrased slightly. Under the null hypothesis $H_0$, both samples $X$ and $Y$ are drawn from the reference distribution $p$. Under the alternative hypothesis $H_1$, sample $X$ is drawn from the reference $p$, while sample $Y$ is drawn from an alternative distribution, $q \neq p$.
The test statistic is then computed as in \cref{eq:t_obs}, but the distribution $f(t | H_0)$ is estimated by testing the reference distribution against itself. This is done through repeated evaluations of the test statistic on pairs $(X^{(i)}, X^{(j)})$ of samples drawn from $p$. If $p$ can be sampled, these samples are independent; otherwise, when only a single instance of data from the reference is available, resampling methods are used. Moreover, in this case the null hypothesis in \cref{eq:null_hyp} becomes asymmetric between $p$ and $q$, as we are specifically testing whether $p$, the distribution of the random variable $x$, also applies to $y$. The choice of which distribution to treat as the reference is part of the test design and can depend on specific biases or computational considerations.

In this work, we proceed by selecting a ``true model" as the reference hypothesis. This reference could be either a generator or a finite dataset. We then introduce a precise methodology to compare the reference with alternative hypotheses, generated by systematically deforming the reference to simulate potential imperfections in a generative model. In our toy examples, where the PDF is known, we directly sample from both the reference and the alternative distributions. For more realistic data, such as those from the HEP literature, we rely on resampling methods. 

\section{Test statistics}\label{sec:teststat}
In this section, we introduce the different test statistics considered in this work. With the exception of the FGD, all the test statistics are derived from or based on IPMs. An IPM can generally be expressed as:
\be\label{eq:ipm}
    d_\mathcal{F}(p,q)=\sup_{f\in \mathcal{F}}\left(\mathbb{E}_{x\sim p}[f(x)]-\mathbb{E}_{y\sim q}[f(y)]\right)\,,
\ee
where $\mathcal{F}$ is a class of real-valued measurable functions, $f: \mathcal{X} \rightarrow \mathbb{R}$.

One of the primary goals of this work is to compare the performance of these tests. To achieve this, we introduce several benchmark scenarios in Section~\ref{sec:distrib}, where the data-generating densities are fully known. In Section~\ref{subsec:likratio}, we also introduce an exact LLR test, which is the most powerful test for simple hypotheses, according to the Neyman-Pearson lemma \cite{neyman1933ix}. This serves as a baseline for the absolute performance against which other tests can be compared.
A summary of all the test statistics introduced in the following is given in Table \ref{tab:tssummary}.

\begin{table*}[t!]
    \small
    \begin{center}
    \begin{tabular}{lc}
    \toprule
    \large Test statistic & \large Section \\
    \midrule
    $t_{\text{SW}} = \dst\frac{1}{K} \sum_{\theta\in\Omega_K}\left(\frac{1}{n}\sum_{i=1}^{n} \mid \underline{x}_{i}^{\theta}-\underline{x}_{i}'^{\theta}\mid\right)\,$ & \ref{subsec:wdtest} \vspace{6mm}\\
    $t_{\overline{\text{KS}}}=\dst\frac{1}{d}\sum_{I=1}^{d}\,\sqrt{\frac{n m}{n+m}}\,\sup_u \mid F_n^{I}(u) - G_m^{I}(u)|$ & \ref{subsec:kstests} \vspace{6mm}\\
    $t_{\text{SKS}}=\dst\frac{1}{K}\sum_{\theta\in\Omega_K}\sqrt{\frac{n m}{n+m}} \sup_u \mid F_n^{\theta}(u) - G_m^{\theta}(u) \mid$ & \ref{subsec:kstests} \vspace{6mm}\\
    $\begin{aligned}t_{\text{MMD}}=&\frac{1}{n(n-1)}\sum_{i=1}^n\sum_{j\neq i}^n k(x_i,x_j)+\frac{1}{m(m-1)}\sum_{i=1}^m\sum_{j\neq i}^m k(y_i,y_j)-\frac{2}{nm}\sum_{i=1}^n\sum_{j=1}^m k(x_i,y_j)\,\,,\\ &\text{with: } k(x,x')=\left(\frac{1}{d}x^T x'+1\right)^4\end{aligned}$ & \ref{subsec:mmdtest} \vspace{6mm}\\
    $t_{\text{FGD}}=\dst\lim_{n,m\rightarrow\infty}\sum_{I=1}^d(\mu_{1,n}^I-\mu_{2,m}^I)^2+\mathrm{tr}\left(\Sigma_{1,n}+\Sigma_{2,m}-2\sqrt{\Sigma_{1,n}\Sigma_{2,m}}\right)$ & \ref{subsec:fgdtest}  \vspace{6mm}\\
    $\dst t_{\text{LLR}}=-2\log \frac{\mathcal{L}_{H_0}}{\mathcal{L}_{H_1}}$ & \ref{subsec:likratio} \vspace{6mm}\\ 
    \bottomrule
    \end{tabular}
    \caption{Summary of the test statistics used in this work.}
    \label{tab:tssummary}
    \end{center}
\end{table*}

\subsection{Sliced Wasserstein distance}\label{subsec:wdtest}

The SW distance \cite{Rabin2011WassersteinBA,Bonneel2014} is a computationally efficient variant of the W distance, derived from optimal transport theory \cite{santambrogio2015optimal, peyre2019computational}. It involves averaging 1D projections of the W distance over all directions on the unit $d$-dimensional sphere, and it is a natural choice for two-sample testing in high-dimensional settings. 

In this work, we focus on the 1-Wasserstein distance, commonly referred to as the \emph{earth mover’s distance}. In one dimension, it is defined as:
\be\label{eq:W1}
    W_{n,m}=\int_\Reals |F_n(u)-G_m(u)| du,
\ee
where $F_n$ and $G_m$ are the eCDFs. 
This distance measures the minimal cost of transforming one distribution into another and depends linearly on the Euclidean distance between data points. It can also be interpreted as an IPM, where the function class $\mathcal{F}$ is the space of 1-Lipschitz functions \cite{villani2021topics}. In the case of two 1D samples with an equal number of data points $m=n$, the quantity in eq.~\eqref{eq:W1} can be simply computed as
\begin{equation}
W_{n}=\frac{1}{n}\sum_{i=1}^{n} \mid \underline{x}_{i}-\underline{x}_{i}'\mid\,,
\end{equation}
where underlined variables represent elements in the set obtained by permuting the original sample with a permutation $\mathcal{P}$ that sorts the points:
\begin{equation}
\{\underline{x}\} =  \mathcal{P}(\{x\} \mid \underline{x}_{1}\leq\ldots\leq\underline{x}_{n})\,.
\end{equation}

The SW test statistic takes the following form:
\be\label{eq:t_sw}
t_{\text{SW}} = \frac{1}{K} \sum_{\theta\in\Omega_K} W_{n}^\theta=\frac{1}{K} \sum_{\theta\in\Omega_K}\left(\frac{1}{n}\sum_{i=1}^{n} \mid \underline{x}_{i}^{\theta}-\underline{x}_{i}'^{\theta}\mid\right)\,.
\ee
where $\Omega_K$ is a set of $K$ directions selected uniformly at random on the unit sphere $\Omega=\{\theta\in\mathbb{R}^d\setminus||\theta||=1\}$, and $\{\underline{x}_i^\theta\}_{i=1}^n=\{\theta^T x_i\}_{i=1}^n$ are the sorted data points projected on the direction $\theta$.
It is important to note that, in the asymptotic limit $m,n\to \infty$ with $m/n\to c \neq 0,\infty$, the distribution of the test statistic in \cref{eq:t_sw} under the null hypothesis will depend on the underlying data distribution. This means that the asymptotic behavior of the test statistic is not distribution-free \cite{ramdas2017wasserstein}.

\subsection{Kolmogorov-Smirnov inspired test statistics}\label{subsec:kstests}
The KS test is a widely used non-parametric method for both goodness-of-fit and two-sample testing \cite{an1933sulla,smirnov1948table}. It measures the largest absolute difference between the two eCDFs of the samples. The KS test can be viewed as an IPM, where the function class $\mathcal{F}$ consists of indicator functions $\mathbb{1}_{(-\infty,t]}$ for all $t\in\mathbb{R}$.

The KS test statistic is defined as
\be\label{eq:t_ks}
t_{\text{KS}} =
\sqrt{\frac{n m}{n+m}}\sup_u \mid F_n(u) - G_m(u) \mid \,.
\ee
where $F_n(u)$ and $G_m(u)$ are the eCDFs of the two samples. The prefactor ensures that, under the null hypothesis (i.e., the two samples are drawn from the same distribution) and as $m, n \to \infty$ with $m/n \to c \neq 0, \infty$, the test statistic follows the Kolmogorov distribution. The Kolmogorov distribution has the following CDF:
\begin{equation}
F_K(x) = 1 - 2 \sum_{k=1}^{\infty} (-1)^{k-1} e^{-2k^2x^2}\,,
\end{equation}
and PDF
\begin{equation}\label{eq:Kpdf}
f_K(x) = \frac{d}{dx} F_K(x) = 8x \sum_{k=1}^{\infty} (-1)^{k-1} k^2 e^{-2k^2x^2}\,.
\end{equation}

While the KS test is widely used for 1D data, its application in higher dimensions is limited due to the curse of dimensionality \cite{bickel1969distribution,justel1997multivariate}. To address this, we consider two efficient multivariate extensions: the mean KS test and the sliced KS test.

\paragraph{Mean KS}
The mean KS test extends the KS test to higher dimensions by averaging the KS statistics computed along each dimension of the data. This test was originally introduced in Ref.~\cite{Coccaro:2019lgs}, using the median, and subsequently applied in Ref.~\cite{Reyes-Gonzalez:2022rco}. Its variation based on the mean was later introduced in Ref.~\cite{Coccaro:2023vtb}.
The test statistic is defined by the average
\be
t_{\overline{\text{KS}}} = \frac{1}{d} \sum_{I=1}^{d} t_{\text{KS}}^{I} = \frac{1}{d} \sum_{I=1}^{d} \sqrt{\frac{n m}{n + m}} \sup_u |F_n^{I}(u) - G_m^{I}(u)|,
\label{eq:ksmeanscaled}
\ee
where $F_n^{I}(u)$ and $G_m^{I}(u)$ are the eCDFs of the 1D marginals along the $I$-th dimension. This approach makes the KS test computationally feasible in multivariate settings. Since the mean KS test is uniquely defined by the 1D marginals, it is not expected to be directly sensitive to correlations between dimensions. To improve the sensitivity to such correlations, we introduce the sliced KS test.

\paragraph{Sliced KS}
Similar to the SW distance introduced in \cref{eq:t_sw}, the sliced KS (SKS) test extends the KS test to higher dimensions by projecting the original $d$-dimensional data onto 1D subspaces. These projections are taken along $K$ random directions sampled from the unit sphere. For each direction $\theta$, the KS test statistic is computed as:
\be
t^\theta_{\text{KS}} = \sqrt{\frac{n m}{n+m}} \sup_u \mid F_n^\theta (u) - G_m^\theta (u) \mid\,.
\ee
where $F_n^\theta(u)$ and $G_m^\theta(u)$ are the eCDFs of the projected samples along direction $\theta$. The SKS test statistic is then defined as the average of the KS statistics across the $K$ random directions:
\be\label{eq:t_sks}
t_{\text{SKS}} = \frac{1}{K}\sum_{\theta\in\Omega_K} t^\theta_{\text{KS}}\,.
\ee
This approach leverages 1D KS tests over multiple projections, making it computationally feasible for higher-dimensional data and potentially sensitive to correlations between dimensions. To the best of our knowledge, the SKS test has not been previously introduced in the literature.

\subsection{Maximum Mean Discrepancy}\label{subsec:mmdtest}
MMD is a statistical measure of the distance between two probability distributions, introduced in Ref.s \cite{gretton2006kernel, gretton2012kernel}. It is an example of an IPM, where the function class $\mathcal{F}$ is the unit ball in a reproducing kernel Hilbert space (RKHS). 

An unbiased empirical estimate of MMD, following Ref.~\cite{gretton2012kernel}, is given by
\be\label{eq:empunbMMD}
\begin{split}
    t_{\text{MMD}}=&
    \frac{1}{n(n-1)}\sum_{i=1}^n\sum_{j\neq i}^n k(x_i,x_j)+\frac{1}{m(m-1)}\sum_{i=1}^m\sum_{j\neq i}^m k(y_i,y_j)\\
    &-\frac{2}{nm}\sum_{i=1}^n\sum_{j=1}^m k(x_i,y_j)\,,
\end{split}
\ee
where $k(x, x’)$ is the kernel function defining the RKHS.

In Ref.~\cite{Kansal:2022spb}, a fourth-order polynomial kernel was used:
\be\label{eq:polykernel}
    k(x,x')=\left(\frac{1}{d}x^T x'+1\right)^4.
\ee
This kernel is not \emph{characteristic}, meaning that $k(x, x’)$ is not a true metric on the space of probability measures. Specifically, for this kernel, the condition $p = q$ is sufficient for $t_{\text{MMD}}(p, q) = 0$, but it is not necessary. For example, this kernel cannot distinguish between distributions that differ beyond their fourth moment (see also example 3 in Ref.~\cite{sriperumbudur2010hilbert}).
Because the polynomial kernel is not characteristic, this instance of MMD is a \emph{pseudo-metric}. In contrast, characteristic kernels, such as Gaussian and Laplacian kernels, are capable of fully distinguishing between different distributions. However, these kernels require tuning hyperparameters, like the kernel bandwidth, based on the data.
Following Ref.~\cite{Kansal:2022spb}, we will use the fourth-order polynomial kernel described in \cref{eq:polykernel}, as it has the advantage of not requiring hyperparameter tuning. A comparison of its performance relative to characteristic kernels could be the subject of future work.
Finally, computing the MMD between two datasets of size $n$ has a computational cost of $\mathcal{O}(n^2)$ because it requires storing the full kernel matrix $K$, where each element is given by $K_{ij} = k(x_i, x_j)$. This makes MMD computationally expensive, especially in large-scale scenarios or when the test needs to be evaluated multiple times.

\subsection{Fr\'echet Gaussian Distance}\label{subsec:fgdtest}
The FGD is a pseudo-metric, specifically the Fréchet distance or 2-Wasserstein distance, between two multivariate Gaussian distributions. These distributions are characterized by means $\mu_1$, $\mu_2 \in \Reals^d$ and covariance matrices $\Sigma_1, \Sigma_2 \in \Reals^{d \times d}$.
The FGD between two samples of sizes $n$ and $m$ is given by:
\begin{equation}\label{eq:FGD}
    \mathrm{FGD}_{n,m}=\sum_{I=1}^d(\mu_{1,n}^I-\mu_{2,m}^I)^2+\mathrm{tr}\left(\Sigma_{1,n}+\Sigma_{2,m}-2\sqrt{\Sigma_{1,n}\Sigma_{2,m}}\right)\,,
\end{equation}
where $\mu_{1,n}^I$ and $\mu_{2,m}^I$ represent the $I$-th components of the sample means $\mu_{1,n}$ and $\mu_{2,m}$, respectively.

As noted in Ref.\cite{Kansal:2022spb}, the FGD is biased when computed on finite samples \cite{binkowski2018demystifying}. To mitigate this bias, an unbiased asymptotic extrapolation can be introduced, as proposed in Ref.\cite{chong2020effectively}. This asymptotic value, denoted as
\begin{equation}\label{eq:t_FGD}
    t_\mathrm{\text{FGD}} \coloneq \mathrm{FGD}_\infty =\lim_{n,m\to \infty} \mathrm{FGD}_{m,n}\,,
\end{equation}
is estimated by fitting a linear model to FGD values computed at different finite sample sizes. For simplicity, in the following we refer to FGD$_\infty$ simply as FGD.



\subsection{Likelihood-ratio}\label{subsec:likratio}
Detecting deviations from a reference model can be framed as a goodness-of-fit test between two competing statistical models. For simple hypotheses, the Neyman–Pearson lemma shows that the most powerful test in this scenario is the likelihood-ratio test \cite{neyman1933ix}. In this work, we introduce the likelihood-ratio test using a slightly non-standard approach, following the two-sample testing framework outlined in Section~\ref{sec:2st}.

The likelihood function for the datasets $X$ and $Y$ under the null hypothesis (where both samples follow the reference distribution $p$) is written as
\be
    \mathcal{L}_{H_0}=\prod_{x\in X} p(x) \prod_{y\in Y} p(y). 
\ee
Under the alternative hypothesis (where the sample $Y$ follows a different distribution $q$), the likelihood is:
\be
    \mathcal{L}_{H_1}=\prod_{x\in X} p(x) \prod_{y\in Y} q(y). 
\ee
The ratio of the likelihoods under the null and alternative hypotheses is then given by
\be\label{eq:likratio}
    \Lambda = \frac{\mathcal{L}_{H_0}}{\mathcal{L}_{H_1}}= \prod_{y\in Y} \frac{p(y)}{q(y)},
\ee
The test statistic for the LLR test is then defined as:
\be\label{LR_test}
    t_{\text{LLR}} = -2 \log \Lambda.
\ee
Notice that, from \cref{eq:likratio}, the value of the test statistic on the observed data does not depend on sample $X$, since $p$ was selected as the reference distribution. As discussed in Section~\ref{sec:2st}, this effectively transforms the problem into a goodness-of-fit test between the reference distribution $p$ and the sample $Y$ in a two-sample testing framework. To estimate the distribution of the test statistic under the null hypothesis, we evaluate \cref{LR_test} on several reference-distributed samples $\{X^{(i)}\}$. 

\section{Methodology}\label{sec:method}
This work builds on recent contributions, such as Ref.~\cite{Kansal:2022spb}, by providing an informative and fair comparison between evaluation metrics. Specifically, we establish a well-defined and reproducible methodology aimed at comparing test statistics close to a meaningful decision boundary.

We begin by using the true PDF of the data, denoted as $p$, as the reference distribution, which defines the null hypothesis. We then introduce various deformations of $p$ to generate alternative hypotheses with distinct characteristics. These deformations, described in Section~\ref{subsec:deform}, are controlled by a unique scalar parameter $\epsilon$. The PDF of each deformed distribution is referred to as $q_{\epsilon}$.

Following the procedure outlined in Section \ref{sec:2st}, to estimate the test statistic $t$ under the null hypothesis, denoted by $t_0$, we use a Monte Carlo approach based on repeated sampling. Multiple random pairs of samples are drawn from $p$, and $t_0$ is computed for each pair. This generates an empirical distribution with eCDF $F(t_0)$ and ePDF $f(t_0)$, which can be used to establish a threshold $t_0^{\alpha}$, corresponding to a preselected false positive rate.

We define two confidence levels corresponding to $1-\alpha = 0.95, 0.99$. 
For each null distribution $f(t_0)$, we identify a threshold $t_0^{\alpha}$ such that:
\begin{equation}
    \alpha= \int_{t_{0}^{\alpha}}^{\infty}dF(t_{0}) = \int_{t_{0}^{\alpha}}^{\infty}f(t_{0})dt_{0}.
\end{equation}
In the Monte Carlo approach, this integral is empirically estimated by repeated sampling:
\begin{equation}
    \alpha = 1-F(t_{0}^{\alpha}) = \frac{\text{Number of simulated } t_{0} \text{ values } \geq t_{0}^{\alpha}}{\text{Total number of simulations}}.
\end{equation}

In this setup and for all tests considered here, with the exception of the LLR, the distribution of the test statistic under the null hypothesis depends only on the reference distribution $p$, and not on the alternative $q_\epsilon$. Therefore, the null distribution needs to be estimated only once for each reference $p$ and sample size.

Once the threshold values $t_{0}^{\alpha}$ are determined, two-sample tests are conducted between samples generated by the true model $p$ and samples from the deformed models $q_\epsilon$. The basic idea is to perform the tests for different values of $\epsilon$ until the test statistic under the alternative hypothesis, $t(\epsilon)$, reaches the threshold $t_0^\alpha$, i.e.~$t(\epsilon) = t_{0}^{\alpha}$. The value of $\epsilon$ for which the null hypothesis is rejected at the CL $\alpha$ is denoted by $\epsilon_\alpha$. This is found by solving the following optimization problem:
\be
\epsilon_{\alpha} = \argmin_{\epsilon} |t(\epsilon)-t_{0}^{\alpha}|,
\ee
using a simple bisection method over a specified range of $\epsilon$. This iterative method repeatedly halves the range until a sufficiently accurate value of $\epsilon_\alpha$ is found.

In practice, to ensure a robust statistical interpretation, each test under the alternative hypothesis is repeated 100 times. For each test, we calculate the average test statistic, $\mu_{t(\epsilon)}$, and its standard deviation, $\sigma_{t(\epsilon)}$. We then evaluate $\mu_{t(\epsilon)}$, $\mu_{t(\epsilon)} - \sigma_{t(\epsilon)}$, and $\mu_{t(\epsilon)} + \sigma_{t(\epsilon)}$. For each of these quantities, if it is above the threshold, the middle of the $\epsilon$ range is shifted left; if it is below, it is shifted right. The range is halved at each step, and the procedure is repeated until both the range of $\epsilon$ and the range of the absolute difference between $t_0^\alpha$ and $\mu_{t(\epsilon)}$, $\mu_{t(\epsilon)} - \sigma_{t(\epsilon)}$, and $\mu_{t(\epsilon)} + \sigma_{t(\epsilon)}$ fall within a predefined tolerance. For convergence, we fixed a tolerance of $10^{-2}$. In cases where the test statistic is particularly noisy (affecting convexity of the optimization problem), we increase the tolerance to $5 \times 10^{-2}$, which does not affect the overall results. The value of $\epsilon$ for which $\mu_{t(\epsilon)}$ converges to $t_0^\alpha$ is identified with $\epsilon_\alpha$. The values of $\epsilon$ for which $\mu_{t(\epsilon)} - \sigma_{t(\epsilon)}$ and $\mu_{t(\epsilon)} + \sigma_{t(\epsilon)}$ converge to $t_0^\alpha$ are used to establish the one standard deviation uncertainty bounds on $\epsilon_\alpha$, denoted as $\epsilon_{\alpha-\text{low}}$ and $\epsilon_{\alpha-\text{up}}$.

The LLR test requires a different approach since it is a parametric test that explicitly depends on both the reference distribution $p$ and the alternative distribution $q_\epsilon$. As a result, also the test statistic under the null hypothesis, denoted by $t_0(\epsilon)$, now varies with $\epsilon$, and so does its ePDF $f(t_0(\epsilon))$. This means that for each value of $\epsilon$, we must compute the null distribution and re-evaluate the threshold for the chosen significance level $\alpha$. The same procedure used for the non-parametric tests is followed here, but with the additional step of generating the null distribution for each value of $\epsilon$. The optimization problem for the LLR then becomes
\be
\epsilon_{\alpha} = \argmin_{\epsilon} |t(\epsilon)-t_{0}^{\alpha}(\epsilon)|,
\ee
where $t(\epsilon)$ is the test statistic under the alternative hypothesis and $t_0^\alpha(\epsilon)$ is the threshold value for the same significance level $\alpha$ and value of $\epsilon$. This optimization is more computationally expensive than for non-parametric tests since the null distribution must be computed for each value of $\epsilon$. Nevertheless, it is carried out using the same bisection method. In cases where the LLR test is highly sensitive to the value of $\epsilon$, such as in high-dimensional scenarios or when the sample size is large, we apply a slightly larger tolerance than $10^{-2}$ on the function value during optimization.

It is important to note that the LLR test can only be applied when the PDFs of both the true and alternative distributions are known. While this holds true for our reference distributions, when it is known analytically, it is not always the case for deformed distributions. Specifically, for random deformations like $\Sigma_{i \neq j}$, $\mathcal{N}$, and $\mathcal{U}$, the analytical forms of the PDFs $q_\epsilon$ cannot be obtained. This situation mimics what often occurs in real-world applications, where samples can be drawn from a distribution even though the exact PDF is unknown.\footnote{A typical example is exactly the one in which the output of the true PDF is ``smeared" by some probability distribution determined, for instance, by a detector response.} For these cases, we only perform and compare the non-parametric tests.

Our methodology can be directly applied to scenarios where the full details of the problem are known (see Section~\ref{subsec:toy}). However, when working with the finite set of simulated gluon jets from the JetNet dataset~\cite{Kansal_JetNet_A_Python_2023}, we must make some adjustments. In this case, we do not have access to the underlying PDFs that generate the data, even for the reference model. Nevertheless, we can still apply the same deformations to the numerical samples from the dataset, as we do with the toy models. We therefore use a bootstrap-based resampling method, where multiple instances are samples with replacement from the full dataset, to estimate the PDF of the test statistic $f(t_0)$. This allows us to test the alternative hypothesis as described earlier. Obviously, since we do not know the analytical form of the PDFs for the physics datasets, we cannot perform the LLR test in these cases.

We applied this methodology across different dimensionalities (number of features) and sample sizes for each model (see Section~\ref{sec:distrib}). Each metric was tested under various conditions, allowing for a fair comparison of the performance of each test statistic in relation to the specific deformations and the complexity of the problem.

\section{Datasets and analysis setup}\label{sec:distrib}
This section describes the distributions, datasets, and deformations used in our analysis, along with details about the dimensionalities and sample sizes considered.

\subsection{Toy models}\label{subsec:toy}
We define the PDFs of the reference distributions analytically, using multivariate Gaussian distributions and mixtures of multivariate Gaussian distributions. Having direct access to these PDFs offers two key advantages: first, it allows us to apply the LLR test for most of the considered deformations, providing optimal benchmarks according to the Neyman-Pearson lemma; second, it enables us to generate as many samples as needed with minimal computational effort. This ensures that we can always build the distribution of the test statistic under the null hypothesis, $f(t_0)$, with the statistics required by the hypothesis we want to test. As a result, the tail regions corresponding to $\alpha=0.05$ and $\alpha=0.01$ are well populated, supporting a robust and fair comparison when evaluating the alternative hypothesis.

\paragraph{Mixture of Gaussians (MoG):}
We consider mixtures of $q$-component, $d$-dimensional multivariate Gaussian distributions, referred to as MoG models, each with diagonal covariance matrices. The reference model is constructed as follows: for a given $q$ and $d$, we randomly generate means from a uniform distribution in the range $[-5, 5]$ and standard deviations in the range $[0, 1]$. The components are then mixed according to a categorical distribution with randomly assigned probabilities $P_i$, where $i = 1, \dots, q$. Each component has a different probability, and all dimensions of a given multivariate Gaussian share the same probability. For our analysis, we consider three MoG models: $q = 3$ components in $d = 5$ dimensions, $q = 5$ components in $d = 20$ dimensions, and $q = 10$ components in $d = 100$ dimensions. It is important to note that, although the covariance matrix for each Gaussian component is diagonal, the overall mixture model exhibits order-one off-diagonal elements in the correlation matrix.

\paragraph{Correlated Gaussians (CG):}
We also consider correlated $d$-dimensional multivariate Gaussian distributions, referred to as CG models, defined by their mean vector and covariance matrix. For a given dimensionality $d$, the reference model is constructed by randomly generating means from a uniform distribution in the range $[-5, 5]$. Full covariance matrix is taken from the correlation matrix of the MoG model with the same dimensionality, with order-one off-diagonal elements. In our analysis, we consider $d=5,20,100$.

\vspace{0.5cm}
For both the MoG and CG models, we compute $f(t_0)$ for each metric by drawing $10^4$ pairs of samples and performing two-sample tests between them, with varying sample sizes $n = m = (1, 2, 5, 10) \times 10^4$. The alternative hypothesis is tested using the procedure outlined in Section~\ref{sec:method}, with 100 pairs of samples drawn from the reference and deformed distributions for each PDF, sample size, and deformation.

\subsection{JetNet Dataset}\label{subsec:Jet}

To explore a scenario where we lack access to the underlying PDFs, and to check our results on a dataset relevant for HEP, we also consider, in our experiments, a dataset of simulated gluon jets within the JetNet dataset. We consider two sets of data: one labeled as particle$\string_$features, which includes features of individual particles within each jet, and another labeled as jet$\string_$features, which contains the overall features of the jets.

Out of a dataset containing a total of $n_{\text{jet}} = 177252$ gluon jets, we use, in our analysis, the following dataset structures:

\begin{itemize}
    \item Particle features dataset: this consists of tensors with shapes $[n_{\text{jet}}, n_{\text{part}}, 3]$, where $n_{\text{part}}$ is the maximum number of particles per jet. The last dimension corresponds to three physical features: the transverse momentum $p_T$, the pseudorapidity $\eta$, and the azimuthal angle $\phi$ with respect to the jet direction;
    \item Jet features dataset: this consists of tensors with shapes $[n_{\text{jet}}, 3]$, where the last dimension contains the transverse momentum $p_T$, the pseudorapidity $\eta$, and the jet mass $m_j$.
\end{itemize}

In this study, for the particle features dataset, we focus on the case with $n_{\text{part}} = 30$, and we discuss why we do not extend the analysis beyond this limit.

Since we do not know the PDFs generating the JetNet datasets, we cannot generate unlimited samples to build $f(t_0)$. To address this, we use a bootstrap method for resampling on the available data: we shuffle the dataset and split it in half, extracting sub-samples of size $n$ from each half and performing two-sample tests between them.

This procedure is repeated until each metric is evaluated $10^3$ times, ensuring that there are enough points available in the $\alpha < 0.01$ region, which allows for a reasonable estimate of the $99\%$ CL threshold. To test the alternative hypothesis, we apply the same method, where one half of the dataset is deformed, and the other half is kept as the reference. This procedure is repeated $10^2$ times for each metric, sample size, and deformation. 

All deformations are performed by first standardizing the dataset, which involves subtracting the mean and dividing by the standard deviation. After applying the deformation, we multiply by the original standard deviation and add back the original mean to ensure consistency in scale.

As in the toy examples, to thoroughly understand the behavior of the test statistics in different complexity scenarios, we selected sample sizes of $n = m = (1, 2, 5) \times 10^4$. Additionally, while testing the particle features dataset, we performed tests both on data with the original scale, where $p_T$ and $m_j$ can be quite large when measured in GeV, and on scaled data, with zero mean and unit variance. This approach allows us to test the robustness of the metrics to changes in the data scale and helps us determine the best approach for each metric.

\subsection{Deformations}\label{subsec:deform}
The deformed models are obtained by applying parametric transformations to the reference models. 
Specifically, given a random variable $X$ distributed according to the reference PDF $p$, the deformation acts as a parametric function of $\epsilon$, transforming $X$ into a new random variable $Y = g(X; \epsilon)$, where $Y$ is distributed according to the deformed PDF $q_{\epsilon}$.

When the transformation function $g$ is bijective, we can express $q_{\epsilon}$ analytically in terms of $p$ through the change of variables formula:
\begin{equation}
    q_{\epsilon}(Y) = p(g^{-1}(Y; \epsilon)) \lvert\det J_{g^{-1}} \rvert = p(g^{-1}(Y; \epsilon)) \lvert\det J_g \rvert ^{-1},
\end{equation}
where $J_{g} = \partial g /\partial x$ is the Jacobian matrix of the transformation $g$. In these cases, the likelihood $q_{\epsilon}(Y)$ can be computed, and so can the LLR test statistic.

However, when $g$ is not bijective or not invertible (e.g.~random transformations, such as a smearing), the PDF $q_{\epsilon}$ cannot be expressed analytically, and the LLR cannot be computed. Nonetheless, we can still generate samples from $q_{\epsilon}$ by drawing samples from $p$ and applying the transformation $g(X; \epsilon)$ to obtain the deformed samples. 

When working with real datasets like JetNet, where only samples drawn from the reference distribution are available, the deformed points are obtained by applying the transformation $g$ directly to the sample points. In such cases, neither $p$ nor $q_{\epsilon}$ are known analytically, and the LLR test cannot be performed.

Below, we describe each deformation as a function of the transformation $g$ and its action on the design matrix $x_{iI}$ of a sample. We also highlight when the function is invertible, and when it is not.

\begin{enumerate}
    \item $\boldsymbol{\mu}${\bf -deformation}\\
        A random vector $\delta \mu_{I}$, with each entry drawn from a uniform distribution in the interval $[-\epsilon, \epsilon]$, is added to each point from the reference model, modifying its mean from $\mu_{I}$ to $\mu_{I}+\delta \mu_{I}$. The transformation $g$ is bijective and can be written as:
        \begin{equation}
            \begin{array}{l}
                \dst y_{iI} = g(x_{iI};\delta \mu_{I}(\epsilon)) = x_{iI} + \delta \mu_{I}(\epsilon)\,,\vspace{2mm}\\
                \dst x_{iI} = g^{-1}(y_{iI};\delta \mu_{I}(\epsilon)) = y_{iI} - \delta \mu_{I}(\epsilon)\,,\vspace{2mm}\\
                \dst \delta \mu_{I}(\epsilon) \sim \mathcal{U}_{[-\epsilon,\epsilon]}\,,
            \end{array}
        \end{equation}
        with $\epsilon \geq 0$.
    \item $\boldsymbol{\Sigma_{II}}${\bf -deformation}\\
        A random vector $\delta \sigma_{I}$, with each entry drawn from a uniform distribution in the interval $[1, 1 + \epsilon]$, is used to scale the standard deviations of the model from $\sigma_{I}$ to $\sigma_{I}\cdot \delta \sigma_{I}$, leaving the correlation matrix unchanged. The mean is kept fixed by subtracting the model mean before scaling and adding the mean back after scaling. In formulae:
        \begin{equation}
            \begin{array}{l}
                \dst y_{iI} = g(x_{iI};\delta \sigma_{I}) = \mu_I + \delta \sigma_{I}(x_{iI}-\mu_I)\,,\vspace{2mm}\\
                \dst x_{iI} = g^{-1}(y_{iI},\delta\sigma_{I}) = \mu_I + \delta \sigma_{I}^{-1}(y_{iI}-\mu_I)\,,\vspace{2mm}\\
                \dst \delta \sigma_{I} \sim \mathcal{U}_{[1,1+\epsilon]}\,,
            \end{array}
        \end{equation}
        with $\epsilon \geq 0$. The transformation is bijective, so that the LLR can be computed.
    \item $\boldsymbol{\Sigma_{I\neq J}}${\bf -deformation}\\
        A fraction $\epsilon$ of each component of the reference model is shuffled independently relative to the others. In formulae:
        \begin{equation}
            y_{iI} = g\left(x_{jI};P_{ij}^{I}(\epsilon)\right) =\sum_j P_{ij}^{I}(\epsilon)x_{jI}\,,
        \end{equation}
        with $\epsilon \geq 0.$ $P_{ij}^{I}$ is a tensor describing $I$ permutation matrices in the $(i,j)$ space (the sum in the formula above does not run on the index $I$). The $P_{ij}^{I}$ permutation matrix has all entries equal to zero but one entry equal to one for each row and column. A fraction $1-\epsilon$ of the ones are on the diagonal, while a fraction $\epsilon$ are off the diagonal. The positions of the ones are random, and different for each $I$.

        Shuffling points independently for each feature has the effect of modifying the covariance matrix without changing the marginal distributions. As $\epsilon$ increases, the correlations between features decreases, leading to a diagonal covariance matrix when $\epsilon = 1$. For $\epsilon > 1$, this deformation is followed by the $\Sigma_{II}$ deformation, which scales the standard deviations further. The inverse transformation and the PDF of the deformed distribution are not known, so that the LLR cannot be computed for this deformation.

    \item $\textbf{pow}^{\boldsymbol{+}}${\bf -deformation}\\
        Each component of each point is modified by raising its absolute value to the power $(1+\epsilon)$, while keeping the sign fixed. This deformation alters the model beyond just its mean and covariance matrix. The transformation is bijective, and can be written as
        \begin{equation}
            \begin{array}{l}
                \dst y_{iI} = g(x_{iI};\epsilon) = \text{sign}(x_{iI})|x_{iI}|^{1+\epsilon}\,,\vspace{2mm}\\
                \dst x_{iI} = g^{-1}(y_{iI};\epsilon) = \text{sign}(y_{iI})|y_{iI}|^{1/(1+\epsilon)}\,,
            \end{array}
        \end{equation}
        with $\epsilon\geq 0$.
    \item $\textbf{pow}^{\boldsymbol{-}}${\bf -deformation}\\
        Similar to the previous deformation, but with the power $(1-\epsilon)$. In formulae:
        \begin{equation}
            \begin{array}{l}
                \dst y_{iI} = g(x_{iI};\epsilon) = \text{sign}(x_{iI})|x_{iI}|^{1-\epsilon}\,,\vspace{2mm}\\
                \dst x_{iI} = g^{-1}(y_{iI};\epsilon) = \text{sign}(y_{iI})|y_{iI}|^{1/(1-\epsilon)}\,,
            \end{array}
        \end{equation}
        with $\epsilon\geq 0$.
    \item $\boldsymbol{\mathcal{N}}${\bf -deformation}\\
        Each component of each point is modified by adding a random value $\delta_{iI}$, drawn from a normal distribution $\mathcal{N}_{0,\epsilon}$ with zero mean and standard deviation $\epsilon$. The transformation can be written as:
        \begin{equation}\label{eq:normal_transform}
            \begin{array}{l}
                \dst y_{iI} = x_{iI} + \delta_{iI}(\epsilon)\,,\vspace{2mm}\\
                \dst \delta_{iI}(\epsilon) \sim \mathcal{N}_{0,\epsilon}\,,
            \end{array}
        \end{equation}
        with $\epsilon\geq 0$. The transformation is not invertible, and the PDF of the deformed distribution is not known, so that the LLR cannot be evaluated.
    \item $\boldsymbol{\mathcal{U}}${\bf -deformation}\\
        Each component of each point is modified by adding a random value $\delta_{iI}$, drawn from a uniform distribution in the $[-\epsilon,\epsilon]$ interval. In formulae:
        \begin{equation}\label{eq:uniform_transform}
            \begin{array}{l}
                \dst y_{iI} = x_{iI} + \delta_{iI}(\epsilon)\,,\vspace{2mm}\\
                \dst \delta_{iI}(\epsilon) \sim \mathcal{U}_{[-\epsilon,\epsilon]}\,,
            \end{array}
        \end{equation}
        with $\epsilon\geq 0$.
        Like the $\mathcal{N}$ deformation, the transformation is not invertible, and the PDF of $q_{\epsilon}$ is not known.
\end{enumerate}

\section{Results}\label{sec:results}
In this section we present the results for all the models discussed above. All results have been obtained with an implementation of the metrics in \textsc{TensorFlow2} available in the \textsc{GitHub} repository \cite{GitHubGMetrics}. All figures and tables documenting the results are available in the \textsc{GitHub} repositories \cite{GitHubGenerativeModelsMetrics, GitHubJetNetMetrics} and listed and linked in Appendices \ref{app:figures} and \ref{app:tables}.

For all the considered distributions, the figures include corner plots showing the 1D and 2D marginal probability distributions for the reference and deformed distributions, color plots that give a pictorial representation of the correlation matrices for the reference and deformed distributions, and the distributions of the test statistics under the null hypothesis for all metrics but the LLR and for all sample sizes. The available tables report the results in terms of the values of $\epsilon$ for which the null hypothesis is rejected (in short, the upper bounds on $\epsilon$) at the $95\%$ and $99\%$ Confidence Level (CL) for each metric, sample size, and deformation.

\subsection{Toy models}\label{sub:toy}
Let us start discussing results for the two classes of toy distributions that we consider in this work, namely the MoG and CG models. Since the full list of results for the different dimensionalities $d=5,20,100$, and sample sizes $n=m=(1,2,5,10)\cdot 10^{4}$ is reported in the Appendices, we show here only a small subset of figures and tables, which helps understanding the methodology and the overall picture of the results that we discuss below.

\begin{figure}[t!]
    \centering
    \includegraphics[width=0.45\textwidth]{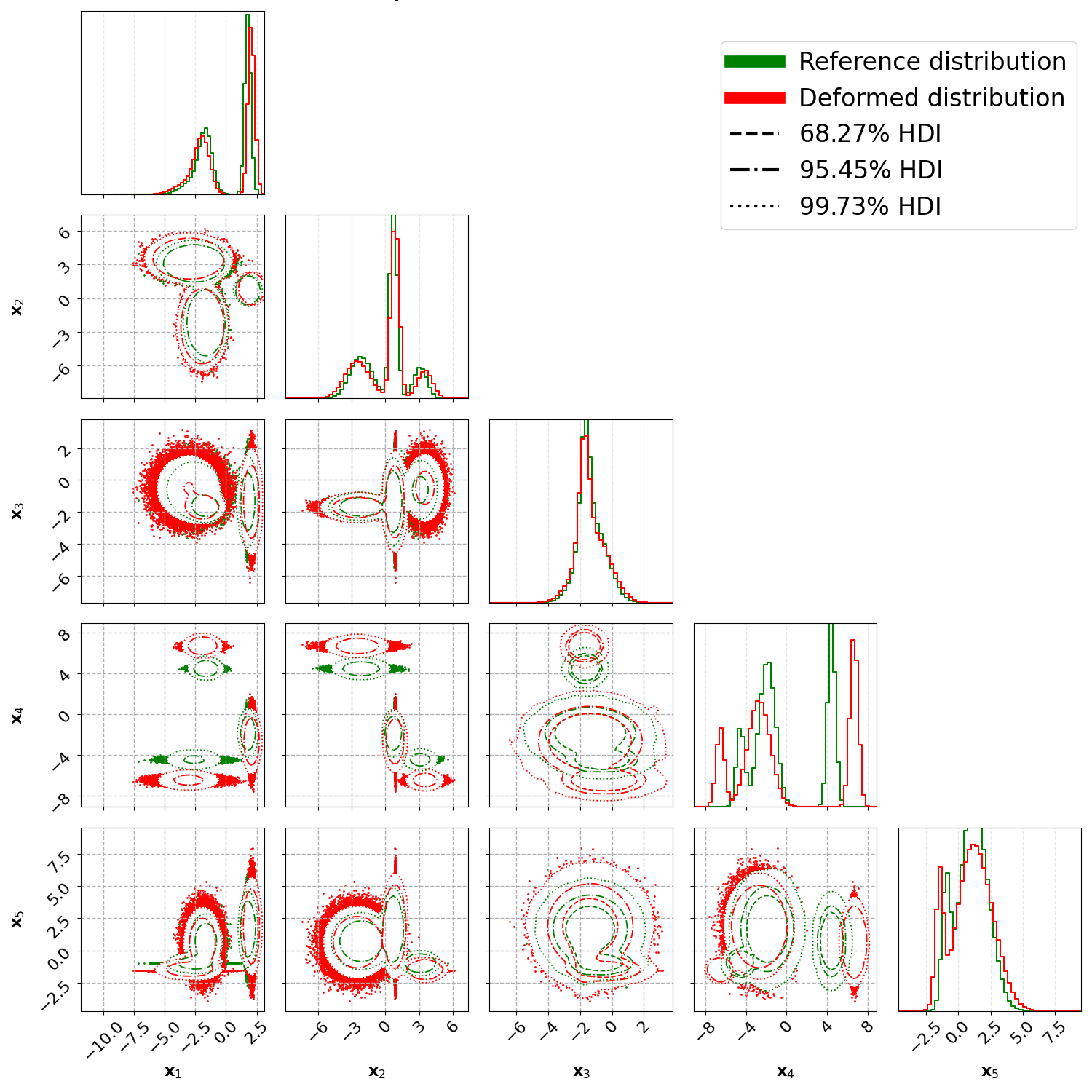}
    \includegraphics[width=0.45\textwidth]{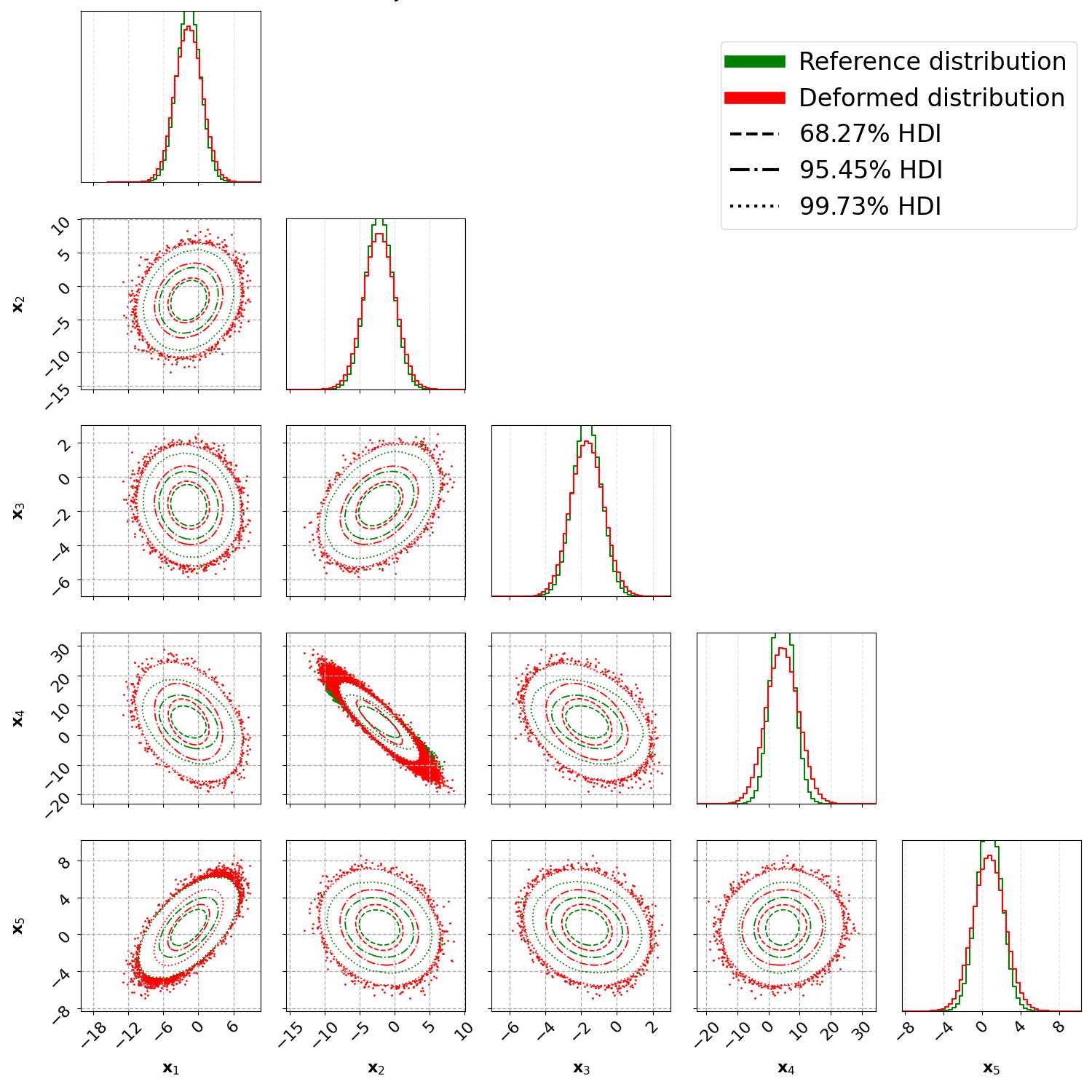}
    \caption{{\bf Left:} Corner plot showing the 1D and 2D marginal probability distributions for the reference and deformed distributions for the MoG model with $d=5$ and $q=3$, and $\Sigma_{ii}$-deformation with $\epsilon=0.5$. {\bf Right:} Same as the left plot but for the CG model with $d=5$. The plots are made with $10^{6}$ points per sample.}
    \label{fig:5D_corner_def_2}
\end{figure}

Figure \ref{fig:5D_corner_def_2} shows corner plots for the MoG and CG models with $d=5$ and $\Sigma_{ii}$-deformation with $\epsilon=0.5$. The plots show the 1D and 2D marginal probability distributions for the reference and deformed distributions. In the figure we clearly see the effect of the $\Sigma_{ii}$-deformation, which changes the vector of standard deviations, while leaving the vector of means and the correlation matrix unchanged. The plots show, in the 2D marginal distributions, contours for the highest density intervals corresponding to a probability of $68.27\%$, $95.45\%$, and $99.73\%$.\footnote{These probabilities correspond to one, two, and three standard deviations for a univariate Gaussian distribution.} Similar plots are available for all the other deformations and dimensionalities considered, and are referenced in Appendix \ref{app:figures}.

\begin{figure}[t!]
    \centering
    \includegraphics[width=0.99\textwidth]{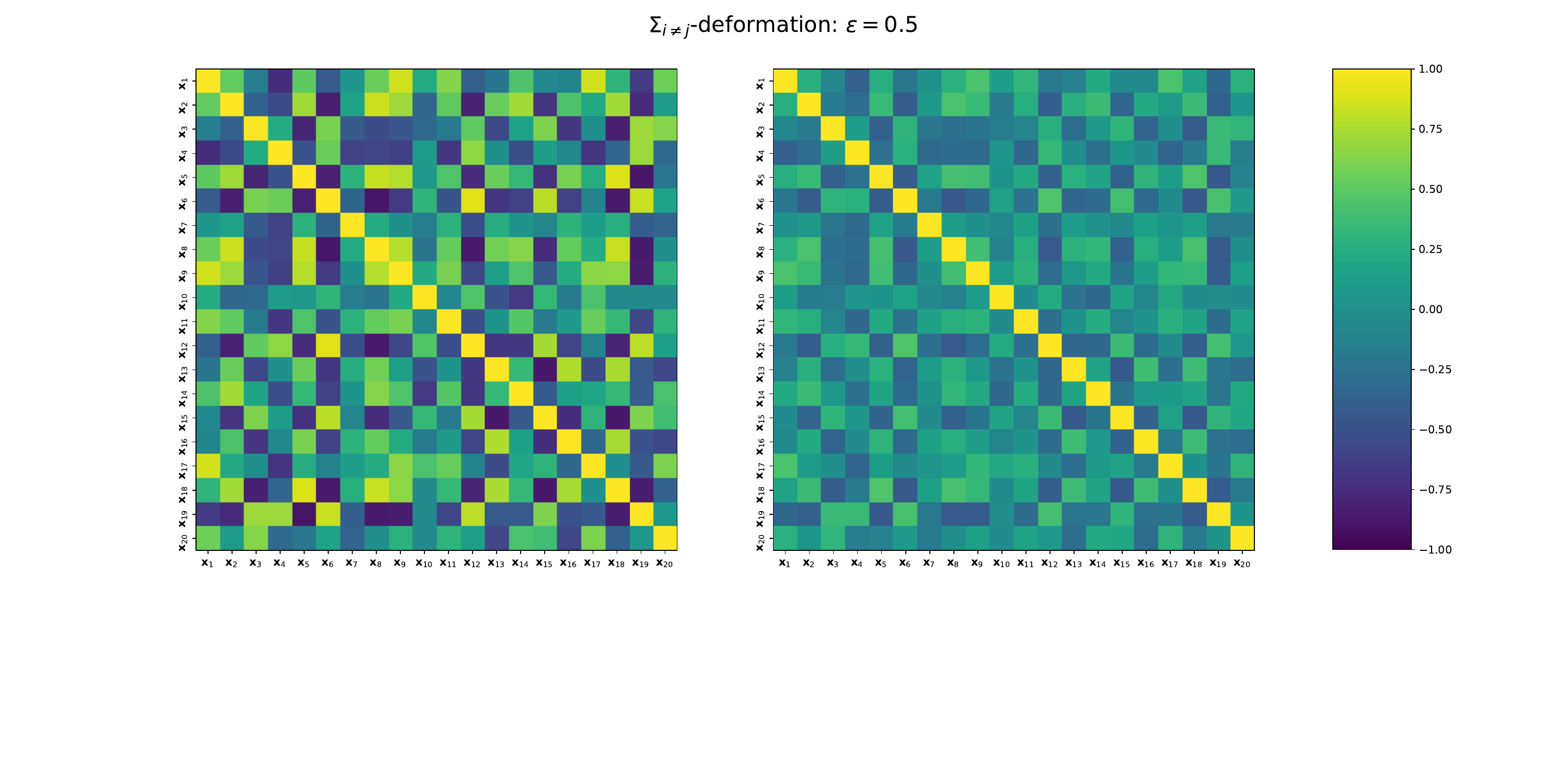}
    \caption{Color plot showing the correlation matrix for the reference (left) and deformed (right) distributions for the MoG model with $d=20$ and $q=5$, and $\Sigma_{i\neq j}$-deformation with $\epsilon=0.5$. The figure is identical in the case of the CG model, since the same correlation matrix is used for both models. The plots are made with $10^{6}$ points per sample.}
    \label{fig:20D_corr_def_3}
\end{figure}

Figure \ref{fig:20D_corr_def_3} shows a color plot for the correlation matrix for the MoG model with $d=20$ and $q=5$ (left panel), and the corresponding $\Sigma_{i\neq j}$ deformation with $\epsilon=0.5$ (right plot). The figure makes very intuitive the effect of the  $\Sigma_{i\neq j}$ deformation: it decreases the correlation between the features until the covariance matrix becomes diagonal (for $\epsilon > 1$ the covariance matrix is kept diagonal and the standard deviation is modified according to the $\Sigma_{ii}$-deformation). Similar plots are available for all the other deformations and dimensionalities considered and are refereced in Appendix \ref{app:figures}.

\begin{figure}[t!]
    \centering
    \includegraphics[width=0.48\textwidth]{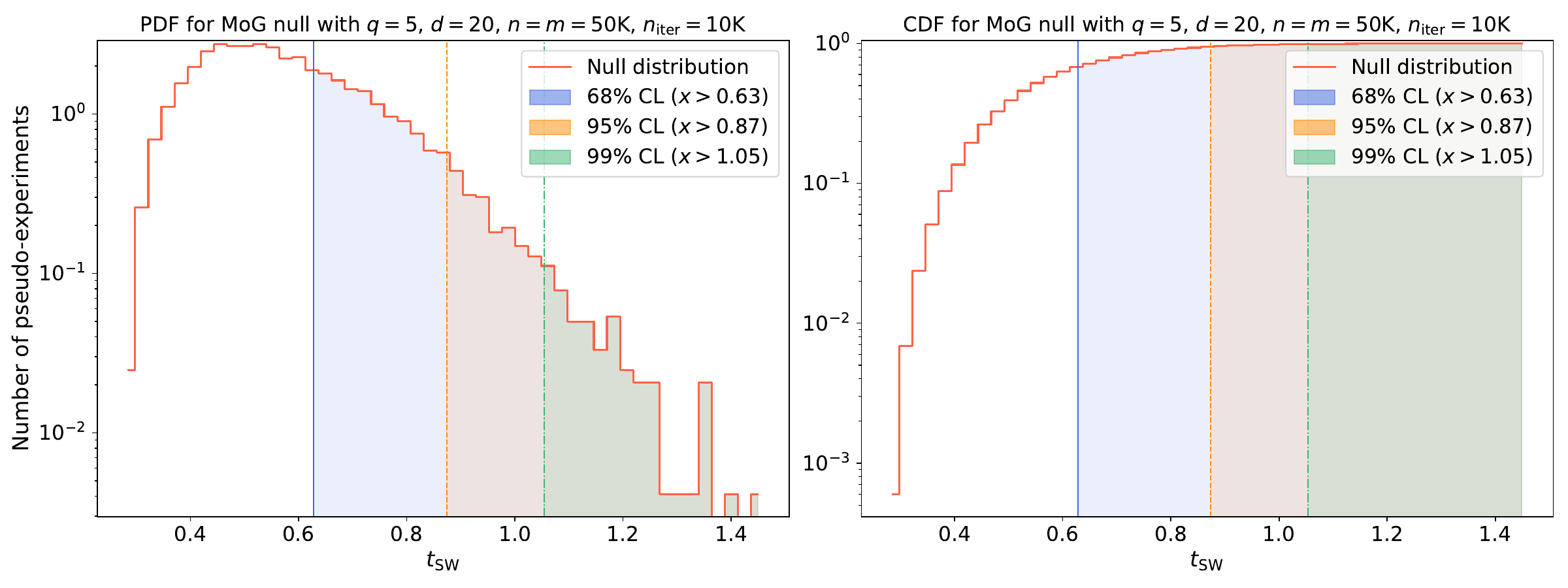}
    \includegraphics[width=0.48\textwidth]{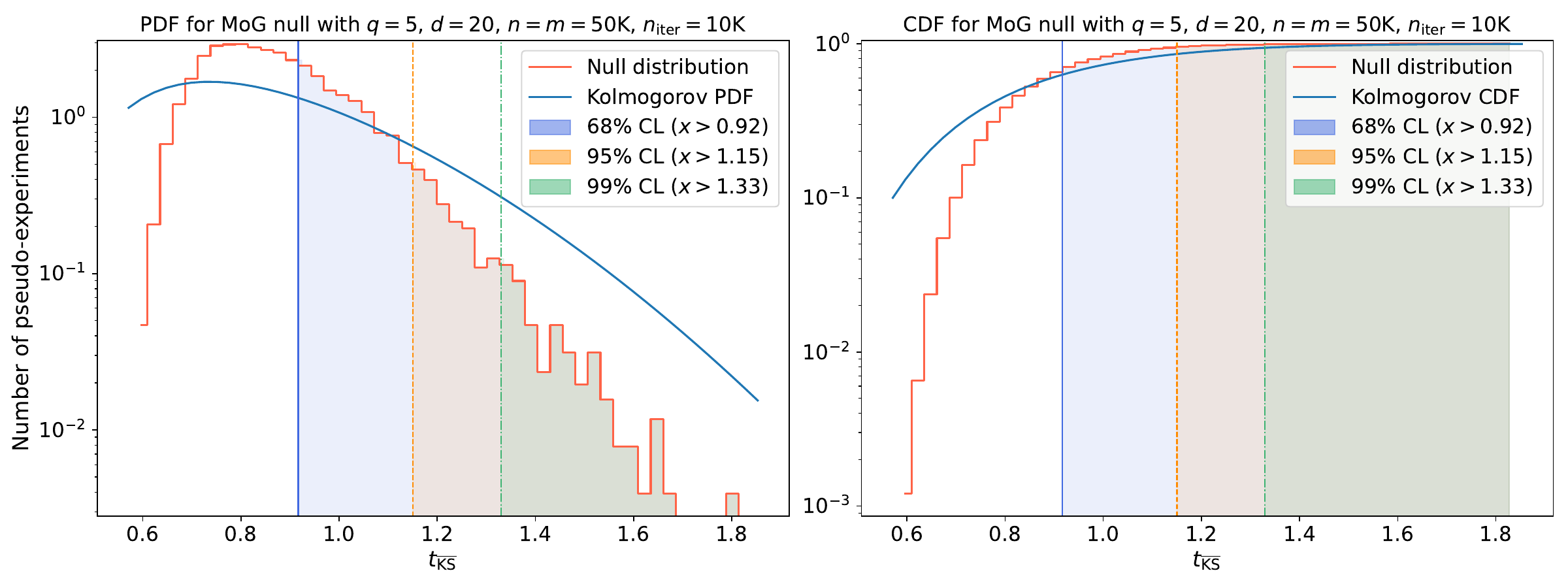}
    \includegraphics[width=0.48\textwidth]{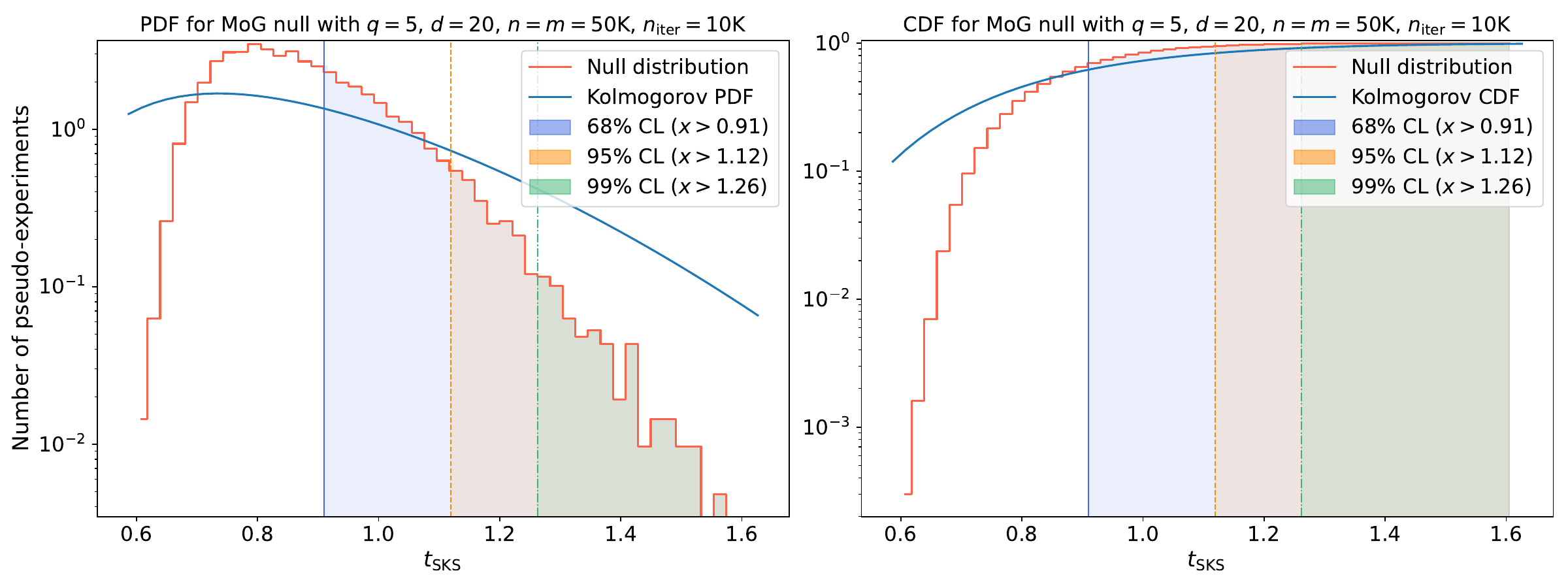}
    \includegraphics[width=0.48\textwidth]{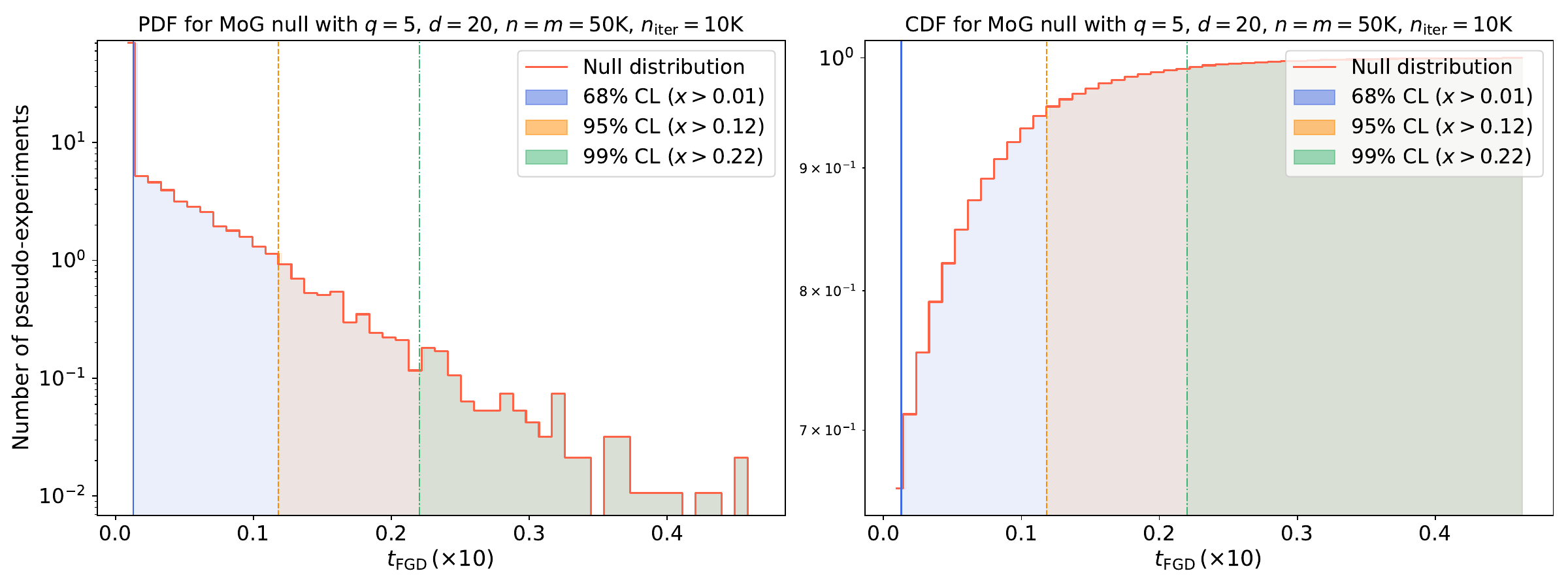}
    \includegraphics[width=0.48\textwidth]{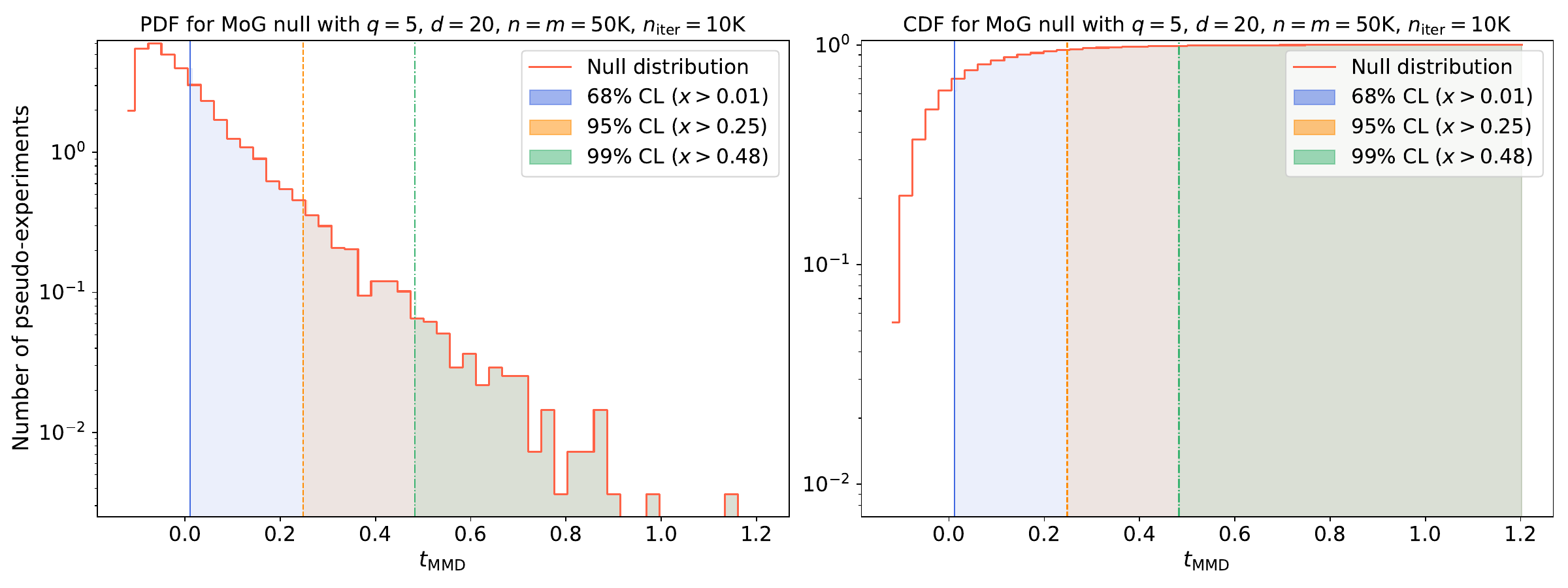}
    \caption{Each pair of plots represents the empirical PDF (left) and CDF (right) of the test statistic under the null hypothesis for the MoG model with $d=20$ and $q=5$, and $n=m=5\cdot 10^{4}$ samples. See the main text for a full description of the plots.}
    \label{fig:100D_table_def_4}
\end{figure}

Figure \ref{fig:100D_table_def_4} shows the empirical PDF and CDF of the various test statistics under the null hypothesis for the MoG model with $d=20$ and $q=5$, and $n=m=5\cdot 10^{4}$ samples. The histograms are built with $10^{4}$ iterations for each test statistic. In blue, red, and green, we highlight the thresholds corresponding to $68\%$, $95\%$, and $99\%$ CLs, respectively. The five panels represent the SW (upper left), $\overline{\mathrm{KS}}$ (upper right), SKS (middle left), FGD (middle right), and MMD (lower) metrics, respectively. The plots corresponding to KS-inspired tests, namely $\overline{\mathrm{KS}}$ and SKS, include a line showing, for comparison, the (parameter free) Kolmogorov distribution expected as asymptotic distribution for the univariate (original) KS test. Similar plots are available for all the other dimensionalities and sample sizes considered and are referenced in Appendix \ref{app:figures}.

Finally, results for each model are summarized in tables and referenced for all dimensionalities and sample sizes considered in Appendix \ref{app:tables}. As an instance, Table \ref{tab:20D_MoG_50K_results} shows the upper bound on $\epsilon$ at the $95\%$ and $99\%$ CL for the MoG model with $d=20$, $q=5$, and $n=m=5\cdot 10^{4}$ samples, while Table \ref{tab:20D_CG_50K_results} shows the same for the CG model with $d=20$ and $n=m=5\cdot 10^{4}$ samples. The tables report the results for all the metrics and all the deformations considered.

\begin{table}[t!]
    \centering
    \footnotesize
    \input{Tables/MoG_20D_50K_results_table}
    \caption{Upper bound on $\epsilon$ at the $95\%$ and $99\%$ CL for all the considered metrics and deformations in the MoG model with $d=20$, $q=5$, and $n=m=5\cdot 10^{4}$ samples.}
    \label{tab:20D_MoG_50K_results}
\end{table}

\begin{table}[t!]
    \centering
    \footnotesize
    \input{Tables/CG_20D_50K_results_table}
    \caption{Upper bound on $\epsilon$ at the $95\%$ and $99\%$ CL for all the considered metrics and deformations in the CG model with $d=20$ and $n=m=5\cdot 10^{4}$ samples.}
    \label{tab:20D_CG_50K_results}
\end{table}

Inspecting results for MoG and CG in different dimensions and with different sample sizes, we can draw a few general conclusions:
\begin{itemize}
    \item for the $\mu$ deformation, and for all dimensions and sample sizes, all metrics perform similarly, within uncertainties, for the CG models, with a slightly better performance of $\overline{\mathrm{KS}}$ for low dimensionalities, and of MMD for large dimensionalities; concerning the MoG models, the situation changes, with $\overline{\mathrm{KS}}$ always performing better than the other metrics by around one order of magnitude; the LLR is always the most sensitive for the $\mu$-deformation, with at least one order 
    of magnitude better performance than the best of the other metrics;
    \item for the $\Sigma_{ii}$ deformation, all metrics perform similarly, within uncertainties, for the CG models, with a slightly better performance of FGD and SW; concerning the MoG models, the situation changes, with $\overline{\mathrm{KS}}$ always performing better than the other metrics by around one order of magnitude; the LLR outperforms the other metrics by at least one, and often two, orders of magnitude;
    \item for the $\Sigma_{i\neq j}$ deformation, FGD is the clear winner, with sensitivities better than other metrics by up to one order of magnitude across all dimensions and sample sizes, and both for the CG and the MoG models; notice that the $\overline{\mathrm{KS}}$ is not sensitive, by construction, to this deformation;
    \item for the pow$^{+}$ and pow$^{-}$ deformations, $\overline{\mathrm{KS}}$ is again the most sensitive for the MoG models, often by one order of magnitude, while MMD is the leader for the CG models, even thought the sensitivity is often only about one or two standard deviations better than that of other metrics; indeed, for the CG models, all metrics perform similarly, within uncertainties, for the pow$^{+}$ and pow$^{-}$ deformations; the LLR is again around one order of magnitude more sensitive than the best metric;
    
    \item finally, for the random deformations $\mathcal{N}$ and $\mathcal{U}$, $\overline{\mathrm{KS}}$ is the most sensitive for the MoG models, with a gap of around one order of magnitude or even more, while FGD is the most sensitive for the CG models, up to about a factor of two compared to the other metrics.
\end{itemize}

The overall picture singles out the $\overline{\mathrm{KS}}$ as generally the most sensitive metric for the MoG models, and the FGD as the most sensitive for the CG models. The great sensitivity of the $\overline{\mathrm{KS}}$ for the MoG models is likely due to the fact that the KS-test is very sensitive for multi-modal distributions since a slight relative difference between the modes reflects very strongly on the difference in the cumulative distribution functions. The FGD, on the other hand, is likely the most sensitive for the CG models because it is designed as a distance between multivariate Gaussians, with sensitivity to differences in both the mean and the covariance matrix. 

Even though $\overline{\mathrm{KS}}$ and FGD sometimes outperform other metrics, in most cases the results are comparable within uncertainties, which suggests, depending on the case, the use of the metric that is most efficient to compute, especially in optimization and model selection phases where many models need to be confronted with a reference or among each others. The computational efficiency can be read from the time needed to estimate the test statistic distribution under the null hypothesis, while the time reported for each deformation is the time needed for the optimization in $\epsilon$, that means to estimate the upper bound on $\epsilon$ and its uncertainty. From the results for the null hypotheses we see that, in the case of the toy models, for which the data generating PDF is known and can be efficiently sampled, the $\overline{\mathrm{KS}}$ and SW metrics are generally the fastest for the computation of the distribution under the null hypothesis, respectively in low and high dimensionality, followed closely by the SKS. In contrast, the FGD and MMD are typically much slower. In particular, FGD slows down considerably with increasing dimensionality (large $d$), while MMD suffers more from increasing the sample size (large $n,m$).

The toy MoG and CG models are designed to highlight the differences that we discussed above. To better understand the performance of the metrics in a real-world scenario, where the data generating PDFs are not known, we now move to the JetNet dataset, with features corresponding to kinematic distributions of jets or particles within jets.

\begin{figure}[t!]
    \centering
    \includegraphics[width=0.49\textwidth]{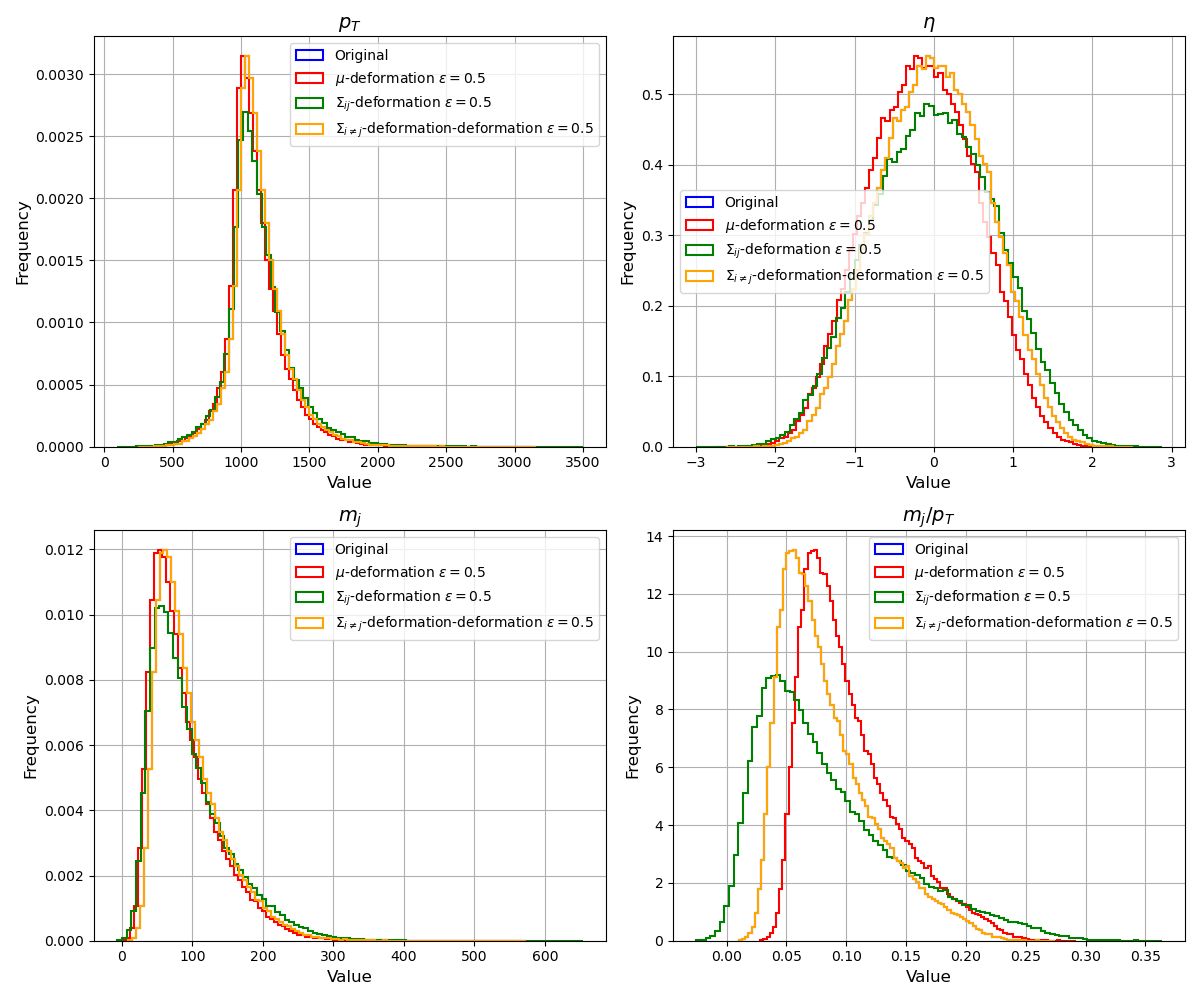}
    \includegraphics[width=0.49\textwidth]{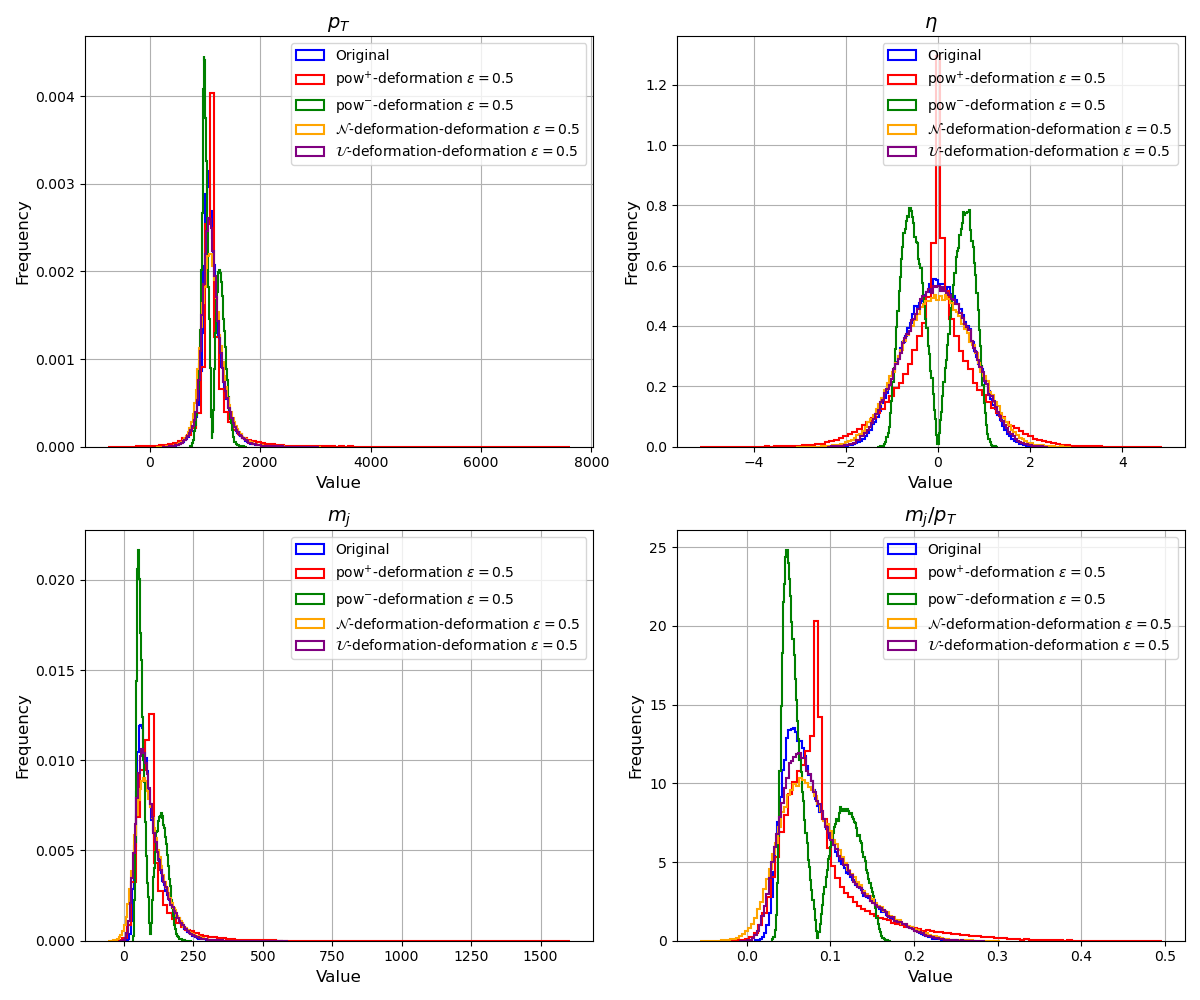}
    \caption{Original jet kinematic distributions compared with the $\mu$, $\Sigma_{ii}$, and $\Sigma_{i\neq j}$ (left), and pow$^{+}$, pow$^{-}$, $\mathcal{N}$, and $\mathcal{U}$ (right) deformations with $\epsilon = 0.5$. The plots are made with $10^{6}$ points per sample.}
    \label{fig:jet_features_deform}
\end{figure}

\subsection{JetNet Dataset}\label{subsec:JetNet}
We now present the results for the JetNet dataset, focusing on the particle and jet features datasets introduced in Section \ref{subsec:Jet}. As in the case of the toy models, we show only a small subset of figures and tables, which helps understanding the methodology and the overall picture of the results. In order to also check the dependence of the metrics performance on the scale of data, in this case we repeated the tests with original data and with data scaled to have zero mean and unit variance.

Before discussing the results, let us mention that, in the particle level dataset, we consider $n_{\text{part}}=30$ particles per jet and we do not go above this number, even though the dataset contains up to 150 particles per jet. This is because as the number of particles increases, so does the number of soft particles, with a $p_{T}$ distribution more and more similar to a $\delta$-function peaked at zero momentum (likely also due to padding when the constituents generated by the Monte Carlo is less than $150$). As some of the features become so narrow, then some metrics become more and more sensitive to smaller and smaller deformations that modify the features. This makes our numerical procedure ineffective. Indeed, if the reference distribution is a $\delta$-function, than even a metric like KS is theoretically infinitely sensitive to any deformation. We find that stopping at $n_{\text{part}}=30$ is a good compromise between a large number of features (three for each particle gives a total of $90$ features) and robustness with respect to the aforementioned issue.

In Figure \ref{fig:jet_features_deform} we show the effect of the various deformations on the 1D marginal distributions in the jet kinematic variables for the jet level dataset. The plots show the distributions for the reference dataset and for the dataset deformed with all the deformations considered and with $\epsilon = 0.5$. The figures give an idea of how the kinematic distributions are affected by the different deformations. 

\begin{figure}[t!]
    \centering
    \includegraphics[width=0.49\textwidth]{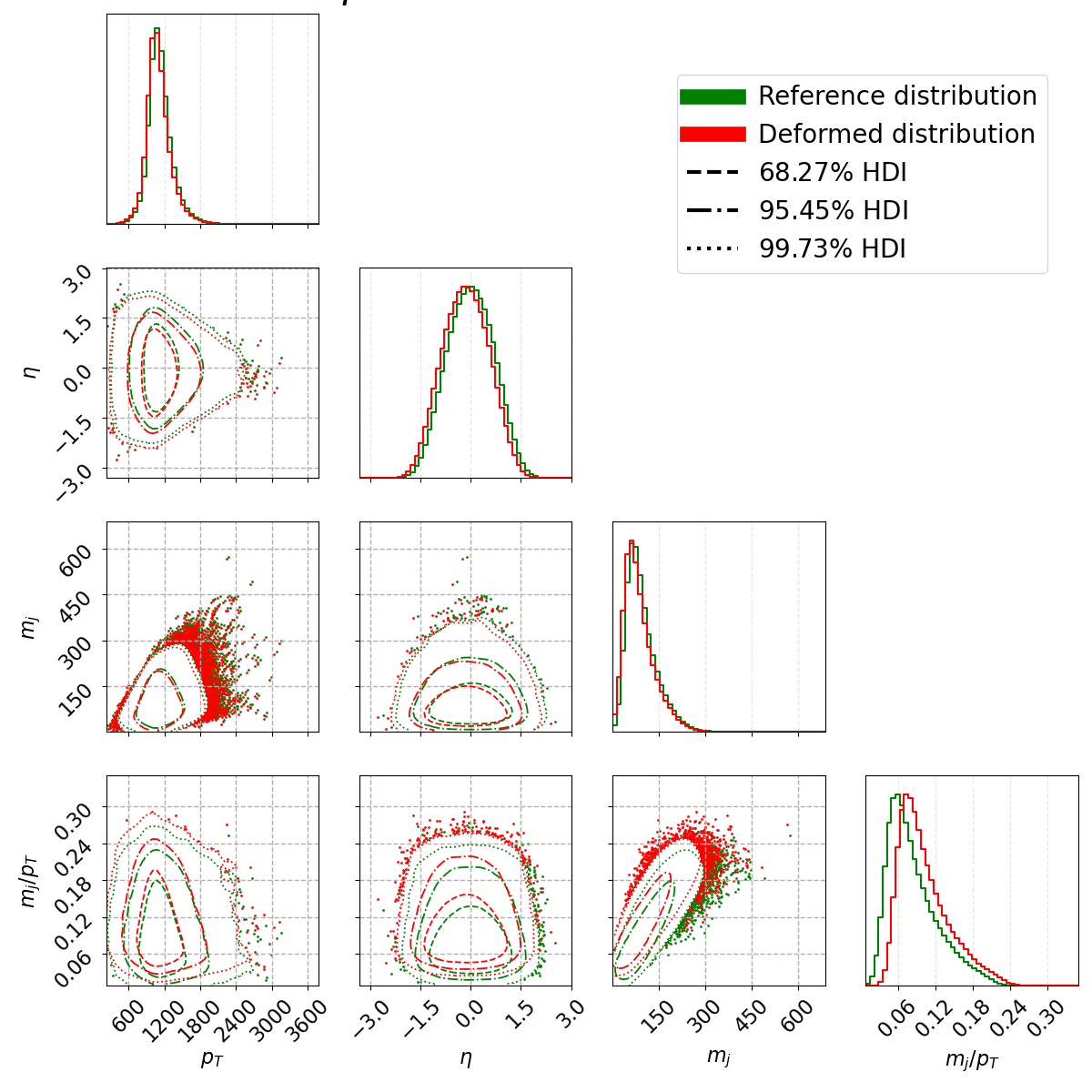}
    \includegraphics[width=0.49\textwidth]{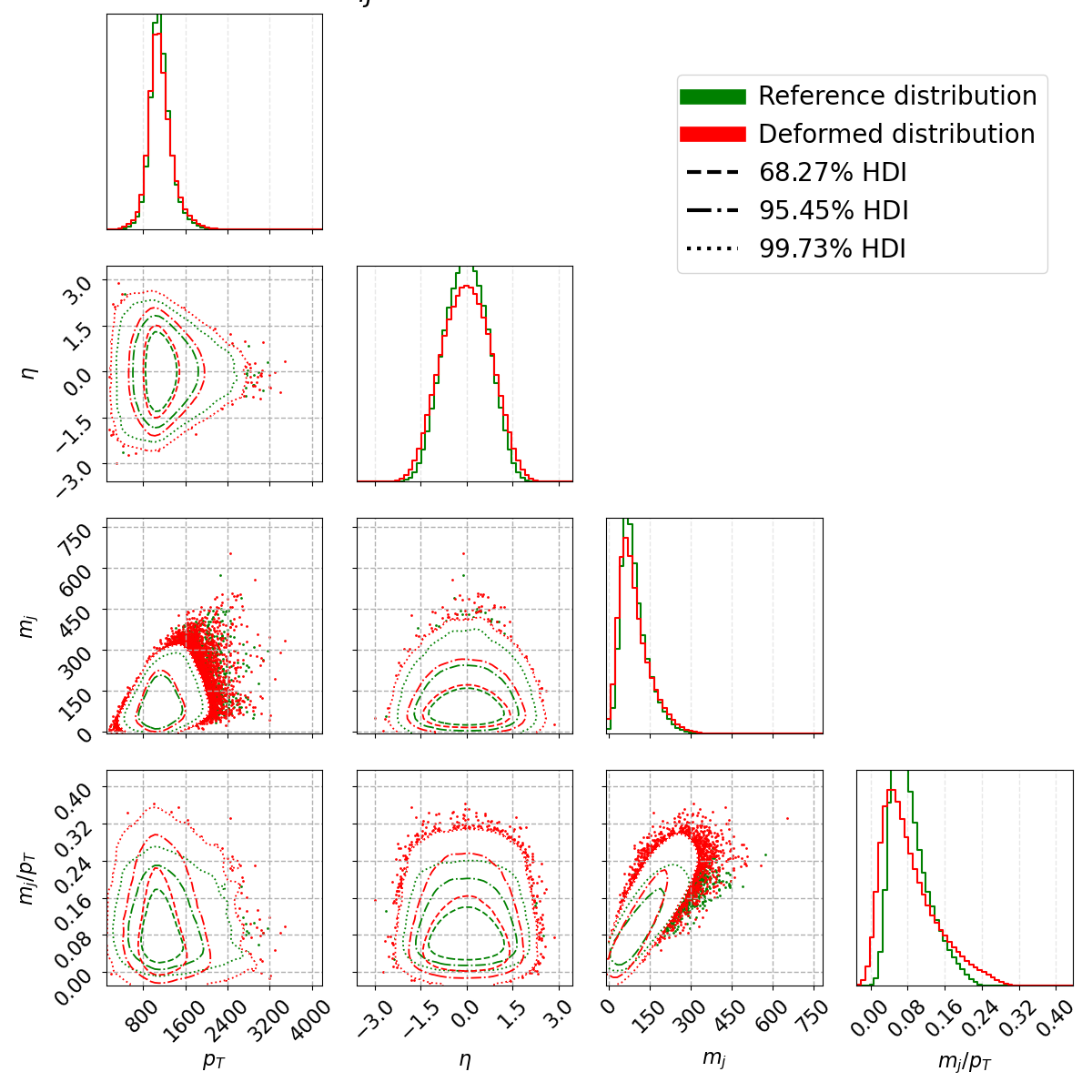}
    \caption{Corner plots of the original jet kinematic distributions compared with the $\mu$ (left) and $\Sigma_{ii}$ (right) deformations with $\epsilon = 0.5$. The plots are made with $10^{6}$ points per sample.}
    \label{fig:jet_features_deform_corner}
\end{figure}

To give an idea of how the correlations are also affected, we show in Figure \ref{fig:jet_features_deform_corner} corner plots for the 1D and 2D marginal distributions corresponding to the $\mu$ and $\Sigma_{ii}$ deformations with $\epsilon = 0.5$ for the jet level dataset.

Even though figures are too large to be shown here, and are references in Appendix \ref{app:figures}, the behavior of the kinematic variables and of the deformations in the case of the particle level dataset is very similar to what we have seen for the jet level dataset.

As for the toy models, the results for the JetNet dataset are summarized in tables and referenced in Appendix \ref{app:tables}. We show here only a small subset of the result tables. Starting from the jet level dataset, Table \ref{tab:jet_features_50K_results} shows the upper bound on $\epsilon$ at the $95\%$ and $99\%$ CL for the jet level dataset with $n=m=5\cdot 10^{4}$ samples. 

\begin{table}[t!]
    \centering
    \footnotesize
    \input{Tables/jet_features_50K_results_table}
    \caption{Upper bound on $\epsilon$ at the $95\%$ and $99\%$ CL for all the considered metrics and deformations in the jet level dataset with $n=m=5\cdot 10^{4}$ samples using original features.}
    \label{tab:jet_features_50K_results}
\end{table}

From the table we see that, except for the $\Sigma_{i\neq j}$-deformation, to which $\overline{\mathrm{KS}}$ is not sensitive by construction, $\overline{\mathrm{KS}}$ is the most sensitive to all other deformations. As expected from the toy results, for the $\Sigma_{i\neq j}$ deformation FGD is the most sensitive test. Despite $\overline{\mathrm{KS}}$ being generally the most sensitive metric, one can see that, within uncertainties, results from all metrics are comparable and the gap among them is not particularly significant. 

\begin{table}[t!]
    \centering
    \footnotesize
    \input{Tables/jet_features_50K_scaled_results_table}
    \caption{Upper bound on $\epsilon$ at the $95\%$ and $99\%$ CL for all the considered metrics and deformations in the jet level dataset with $n=m=5\cdot 10^{4}$ samples using features scaled to zero mean and unit variance.}
    \label{tab:jet_features_50K_scaled_results}
\end{table}

In contrast with the results in Table \ref{tab:jet_features_50K_results}, we show in Table \ref{tab:jet_features_50K_scaled_results} the results obtained performing the tests on data scaled to zero mean and unit variance. As expected, the results for $\overline{\mathrm{KS}}$ remain unchanged, since the KS test is ``scale invariant", while the results for the other metrics are affected. Moreover, while FGD continues to be the most sensitive metric for the $\Sigma_{i\neq j}$ deformation, now with a large gap (one order of magnitude better than the best of the other metrics), the other metrics become now more sensitive and compete with $\overline{\mathrm{KS}}$ for the other deformations. Nevertheless, results for the other deformations are comparable for the different metrics within uncertainties.\\

We can now proceed to discuss the results for the particle level dataset. Table \ref{tab:particle_features_50K_results} shows the upper bound on $\epsilon$ at the $95\%$ and $99\%$ CL for the particle level dataset with $n=m=5\cdot 10^{4}$ samples. Zeros in the table represent values smaller than $10^{-5}$ obtained from the optimization within a specified tolerance.

\begin{table}[t!]
    \centering
    \footnotesize
    \input{Tables/particle_features_50K_results_table}
    \caption{Upper bound on $\epsilon$ at the $95\%$ and $99\%$ CL for all the considered metrics and deformations in the particle level dataset with $n=m=5\cdot 10^{4}$ samples using original features.}
    \label{tab:particle_features_50K_results}
\end{table}

From the table we start to observe the effect discussed at the beginning of the section, where $\overline{\mathrm{KS}}$ becomes much more sensitive than the other tests (for all deformations except $\Sigma_{i\neq j}$). This increased sensitivity is due to the presence of several features with a narrow distribution, for which $\overline{\mathrm{KS}}$ is particularly sensitive. Moreover, despite a few exceptions, all other metrics generally perform equally well within uncertainties.

\begin{table}[t!]
    \centering
    \footnotesize
    \input{Tables/particle_features_50K_scaled_results_table}
    \caption{Upper bound on $\epsilon$ at the $95\%$ and $99\%$ CL for all the considered metrics and deformations in the particle level dataset with $n=m=5\cdot 10^{4}$ samples features scaled to zero mean and unit variance.}
    \label{tab:particle_features_50K_scaled_results}
\end{table}

Finally, Table \ref{tab:particle_features_50K_scaled_results} shows the results obtained using features from the particle level dataset scaled to zero mean and unit variance. As in the case of the jet level dataset, the results for $\overline{\mathrm{KS}}$ remain unchanged, while the results for the other metrics are affected by the scaling of the data. The FGD metric becomes very sensitive to the $\Sigma_{i\neq j}$ deformation, outperforming the other metrics by more than one order of magnitude. Despite all other tests show an increased sensitivity, they still struggle to compete with $\overline{\mathrm{KS}}$, except in the case of the random deformations, where FGD is the most sensitive. Even in these cases, the gap is not particularly significant, and, at least compared to $\overline{\mathrm{KS}}$, lies within uncertainties.

In summary, also for the JetNet dataset, the $\overline{\mathrm{KS}}$ test and the FGD test are the most sensitive for most of the deformations and for most values of the sample size. Despite some special cases, though, the SW test gives similar results, within uncertainties, with a testing time that is up to fifty times smaller than that of the other tests, and always below one hour, even for sample sizes of $5\cdot 10^{4}$.

Notice that, while the $\overline{\mathrm{KS}}$ metric is extremely fast to evaluate in the case of the toy models, for most of the configurations in terms of dimensions and number of samples, it becomes much less efficient for the JetNet dataset. This is due to the fact that even though $\overline{\mathrm{KS}}$ is highly parallelizable, we could obtain a very efficient implementation only when the data generating distribution is available, while we had more troubles getting the most out of parallel computation with resampling techniques. This is not the same for the SW test, which remains extremely efficient also in the case of resampling.

\section{Conclusions and Outlook}\label{sec:conclusion}
In this work, we have developed and thoroughly tested a comprehensive framework for comparing non-parametric two-sample tests in the context of evaluating high-dimensional generative models. Our analysis has focused on several key metrics, including the sliced Wasserstein distance, the mean of 1D Kolmogorov-Smirnov tests, a sliced variation of the Kolmogorov-Smirnov test, the unbiased quadratic Maximum Mean Discrepancy, and the unbiased Fréchet Gaussian Distance. Where possible, we have also benchmarked these non-parametric methods against the parametric log-likelihood-ratio test.

Our results demonstrate that non-parametric tests based on 1D marginal distributions or 1D projections can achieve sensitivity comparable to more complex multivariate metrics. This is particularly significant in high-dimensional settings where multivariate tests often struggle due to the curse of dimensionality, in contrast with the need for computational efficiency. Tests such as the sliced Wasserstein distance and the Kolmogorov-Smirnov-inspired tests show strong performance while maintaining lower computational costs, making them well suited for tasks involving large sample sizes and high-dimensional data, especially in the phase of model selection and optimization of generative models.

As expected, the log-likelihood-ratio test consistently outperforms non-parametric methods when the underlying probability densities of both the reference and deformed distributions are known. However, in practical applications where exact probability densities are not available, such as for the JetNet dataset we considered, non-parametric approaches are indispensable.

Our methodology has proven to be flexible, scalable, and capable of accommodating various deformations across a wide range of dimensions and sample sizes. The combination of toy models and real-world datasets used in this study provides a valuable benchmark, reflecting the theoretical and practical challenges faced in generative model validation.

Furthermore, with few exceptions, each metric exhibits similar sensitivity to the value of $\epsilon$ across different datasets for the same deformation. This consistency demonstrates that our procedure provides a robust evaluation of the tests themselves, rather than merely assessing the specific models and deformations considered.

Looking ahead, there are several promising directions for future research and applications of this methodology:

\begin{enumerate}
	\item Extension to classifier-based metrics: the methodology and results presented here can serve as a foundation for designing, implementing, and validating classifier-based two-sample tests, which have shown substantial potential in HEP applications. Extending our work to include these methods could significantly enhance the evaluation of generative models, a crucial aspect in scientific domains.
	\item Enhancement of computational efficiency: while our current implementation of the proposed metrics in \textsc{TensorFlow2} is computationally efficient, there remains room for further optimization. Improvements in computational efficiency would make these methods more accessible and scalable for even larger datasets, facilitating their application in real-world problems.
	\item Benchmarking on state-of-the-art generative models: our methodology offers a robust framework for benchmarking the performance of advanced generative models such as Generative Adversarial Networks, Variational Autoencoders, and Normalizing Flows, particularly in scientific applications where precision and reliability are critical.
	\item Uncertainty quantification: although our methodology provides a well-defined procedure for assigning uncertainty to test results, future work could explore more comprehensive approaches. For instance, in the case of generative models where deformations are not parameterized by a single parameter $\epsilon$, the full distribution of the test statistic under the alternative hypothesis could be estimated, enabling a more detailed assessment of the test power. This could also allow for the computation of the complete confusion matrix, providing deeper insights into the statistical properties of the tests.
\end{enumerate}

In conclusion, we believe that the methodology developed in this work represents a step forward in the key challenge of evaluating generative models, particularly in scientific applications where precision and reliability are key requirements.

\subsection*{Acknowledgments}\label{sec:ackn}
M.L. acknowledges the financial support of the European Research Council (grant SLING 819789).

\appendix


\section{List of Figures}\label{app:figures}

\subsection{List of Figures for the CG models}\label{app:figures_CG}
\subsubsection{List of Figures for CG results in $5D$}
The following Figures show the 1D and 2D marginal probability distributions for two samples $\mathbf{X}_{1}$, drawn from the reference distribution of the CG model with $d=5$, and $\mathbf{X}_{2}$, corresponding to the specified deformation. The plots are made with $10^{6}$ points per sample.
    
\begin{itemize}
    \item \href{https://github.com/TwoSampleTests/GenerativeModelsMetrics/blob/main/results/5D_unimodal/corner_plot_def_1.png}{CG\_5D\_corner\_def\_1}: $\mu$-deformation with $\epsilon=0.5$.
    \item \href{https://github.com/TwoSampleTests/GenerativeModelsMetrics/blob/main/results/5D_unimodal/corner_plot_def_2.png}{CG\_5D\_corner\_def\_2}: $\Sigma_{ii}$-deformation with $\epsilon=0.5$.
    \item \href{https://github.com/TwoSampleTests/GenerativeModelsMetrics/blob/main/results/5D_unimodal/corner_plot_def_3.png}{CG\_5D\_corner\_def\_3}: $\Sigma_{i\neq j}$-deformation with $\epsilon=0.5$.
    \item \href{https://github.com/TwoSampleTests/GenerativeModelsMetrics/blob/main/results/5D_unimodal/corner_plot_def_4.png}{CG\_5D\_corner\_def\_4}: $\text{pow}^{+}$-deformation with $\epsilon=0.1$.
    \item \href{https://github.com/TwoSampleTests/GenerativeModelsMetrics/blob/main/results/5D_unimodal/corner_plot_def_5.png}{CG\_5D\_corner\_def\_5}: $\text{pow}^{-}$-deformation with $\epsilon=0.1$.
    \item \href{https://github.com/TwoSampleTests/GenerativeModelsMetrics/blob/main/results/5D_unimodal/corner_plot_def_6.png}{CG\_5D\_corner\_def\_6}: $\mathcal{N}$-deformation with $\epsilon=0.5$.
    \item \href{https://github.com/TwoSampleTests/GenerativeModelsMetrics/blob/main/results/5D_unimodal/corner_plot_def_7.png}{CG\_5D\_corner\_def\_7}: $\mathcal{U}$-deformation with $\epsilon=0.5$.
\end{itemize}

\noindent The following Figures report a color plot representation of the correlation matrices for two samples $\mathbf{X}_{1}$, drawn from the reference distribution of the CG model with $d=5$, and $\mathbf{X}_{2}$, corresponding to the specified deformation. The plots are made with $10^{6}$ points per sample.

\begin{itemize}
    \item \href{https://github.com/TwoSampleTests/GenerativeModelsMetrics/blob/main/results/5D_unimodal/corre_matrix_plot_def_1.pdf}{CG\_5D\_corr\_matrix\_def\_1}: $\mu$-deformation with $\epsilon = 0.5$.
    \item \href{https://github.com/TwoSampleTests/GenerativeModelsMetrics/blob/main/results/5D_unimodal/corre_matrix_plot_def_2.pdf}{CG\_5D\_corr\_matrix\_def\_2}: $\Sigma_{ii}$-deformation with $\epsilon = 0.5$.
    \item \href{https://github.com/TwoSampleTests/GenerativeModelsMetrics/blob/main/results/5D_unimodal/corre_matrix_plot_def_3.pdf}{CG\_5D\_corr\_matrix\_def\_3}: $\Sigma_{i\neq j}$-deformation with $\epsilon = 0.5$.
    \item \href{https://github.com/TwoSampleTests/GenerativeModelsMetrics/blob/main/results/5D_unimodal/corre_matrix_plot_def_4.pdf}{CG\_5D\_corr\_matrix\_def\_4}: $\text{pow}^{+}$-deformation with $\epsilon = 0.1$.
    \item \href{https://github.com/TwoSampleTests/GenerativeModelsMetrics/blob/main/results/5D_unimodal/corre_matrix_plot_def_5.pdf}{CG\_5D\_corr\_matrix\_def\_5}: $\text{pow}^{-}$-deformation with $\epsilon = 0.1$.
    \item \href{https://github.com/TwoSampleTests/GenerativeModelsMetrics/blob/main/results/5D_unimodal/corre_matrix_plot_def_6.pdf}{CG\_5D\_corr\_matrix\_def\_6}: $\mathcal{N}$-deformation with $\epsilon = 0.5$.
    \item \href{https://github.com/TwoSampleTests/GenerativeModelsMetrics/blob/main/results/5D_unimodal/corre_matrix_plot_def_7.pdf}{CG\_5D\_corr\_matrix\_def\_7}: $\mathcal{U}$-deformation with $\epsilon = 0.5$.
\end{itemize}

\noindent The following Figures show the empirical PDF of the specified test statistic under the null hypothesis (both samples come from the same reference distribution) for the CG model with $d=5$ and $n=m=(1,2,5,10)\cdot 10^{4}$. The test statistics distributions are obtained with $10^{4}$ iterations.

\begin{multicols}{3}
\begin{itemize}
    \item \href{https://github.com/TwoSampleTests/GenerativeModelsMetrics/tree/main/results/5D_unimodal/5D_10K/null_hypothesis/SWD.pdf}{CG\_5D\_10K\_null\_SW}\\
    \href{https://github.com/TwoSampleTests/GenerativeModelsMetrics/tree/main/results/5D_unimodal/5D_20K/null_hypothesis/SWD.pdf}{CG\_5D\_20K\_null\_SW}\\
    \href{https://github.com/TwoSampleTests/GenerativeModelsMetrics/tree/main/results/5D_unimodal/5D_50K/null_hypothesis/SWD.pdf}{CG\_5D\_50K\_null\_SW}\\
    \href{https://github.com/TwoSampleTests/GenerativeModelsMetrics/tree/main/results/5D_unimodal/5D_100K/null_hypothesis/SWD.pdf}
    {CG\_5D\_100K\_null\_SW} 
    \item \href{https://github.com/TwoSampleTests/GenerativeModelsMetrics/tree/main/results/5D_unimodal/5D_10K/null_hypothesis/KS.pdf}{CG\_5D\_10K\_null\_$\overline{\mathrm{KS}}$}\\
    \href{https://github.com/TwoSampleTests/GenerativeModelsMetrics/tree/main/results/5D_unimodal/5D_20K/null_hypothesis/KS.pdf}{CG\_5D\_20K\_null\_$\overline{\mathrm{KS}}$}\\
    \href{https://github.com/TwoSampleTests/GenerativeModelsMetrics/tree/main/results/5D_unimodal/5D_50K/null_hypothesis/KS.pdf}{CG\_5D\_50K\_null\_$\overline{\mathrm{KS}}$}\\
    \href{https://github.com/TwoSampleTests/GenerativeModelsMetrics/tree/main/results/5D_unimodal/5D_100K/null_hypothesis/KS.pdf}{CG\_5D\_100K\_null\_$\overline{\mathrm{KS}}$}
    \item \href{https://github.com/TwoSampleTests/GenerativeModelsMetrics/tree/main/results/5D_unimodal/5D_10K/null_hypothesis/SKS.pdf}{CG\_5D\_10K\_null\_SKS}\\
    \href{https://github.com/TwoSampleTests/GenerativeModelsMetrics/tree/main/results/5D_unimodal/5D_20K/null_hypothesis/SKS.pdf}{CG\_5D\_20K\_null\_SKS}\\
    \href{https://github.com/TwoSampleTests/GenerativeModelsMetrics/tree/main/results/5D_unimodal/5D_50K/null_hypothesis/SKS.pdf}{CG\_5D\_50K\_null\_SKS}\\
    \href{https://github.com/TwoSampleTests/GenerativeModelsMetrics/tree/main/results/5D_unimodal/5D_100K/null_hypothesis/SKS.pdf}{CG\_5D\_100K\_null\_SKS}\\\\
    \item \href{https://github.com/TwoSampleTests/GenerativeModelsMetrics/tree/main/results/5D_unimodal/5D_10K/null_hypothesis/FGD.pdf}{CG\_5D\_10K\_null\_FGD}\\ 
    \href{https://github.com/TwoSampleTests/GenerativeModelsMetrics/tree/main/results/5D_unimodal/5D_20K/null_hypothesis/FGD.pdf}{CG\_5D\_20K\_null\_FGD}\\
    \href{https://github.com/TwoSampleTests/GenerativeModelsMetrics/tree/main/results/5D_unimodal/5D_50K/null_hypothesis/FGD.pdf}{CG\_5D\_50K\_null\_FGD}\\ 
    \href{https://github.com/TwoSampleTests/GenerativeModelsMetrics/tree/main/results/5D_unimodal/5D_100K/null_hypothesis/FGD.pdf}{CG\_5D\_100K\_null\_FGD}
    \item \href{https://github.com/TwoSampleTests/GenerativeModelsMetrics/tree/main/results/5D_unimodal/5D_10K/null_hypothesis/MMD.pdf}{CG\_5D\_10K\_null\_MMD}\\
    \href{https://github.com/TwoSampleTests/GenerativeModelsMetrics/tree/main/results/5D_unimodal/5D_20K/null_hypothesis/MMD.pdf}{CG\_5D\_20K\_null\_MMD}\\
    \href{https://github.com/TwoSampleTests/GenerativeModelsMetrics/tree/main/results/5D_unimodal/5D_50K/null_hypothesis/MMD.pdf}{CG\_5D\_50K\_null\_MMD}\\
    \href{https://github.com/TwoSampleTests/GenerativeModelsMetrics/tree/main/results/5D_unimodal/5D_100K/null_hypothesis/MMD.pdf}{CG\_5D\_100K\_null\_MMD} \\\\
\end{itemize}
\end{multicols}

\subsubsection{List of Figures for CG results in $20D$}
The following Figures show the 1D and 2D marginal probability distributions for two samples $\mathbf{X}_{1}$, drawn from the reference distribution of the CG model with $d=20$, and $\mathbf{X}_{2}$, corresponding to the specified deformation. The plots are made with $10^{6}$ points per sample.

\begin{itemize}
    \item \href{https://github.com/TwoSampleTests/GenerativeModelsMetrics/blob/main/results/20D_unimodal/corner_plot_def_1.png}{CG\_20D\_corner\_def\_1}: $\mu$-deformation with $\epsilon=0.5$.
    \item \href{https://github.com/TwoSampleTests/GenerativeModelsMetrics/blob/main/results/20D_unimodal/corner_plot_def_2.png}{CG\_20D\_corner\_def\_2}: $\Sigma_{ii}$-deformation with $\epsilon=0.5$.
    \item \href{https://github.com/TwoSampleTests/GenerativeModelsMetrics/blob/main/results/20D_unimodal/corner_plot_def_3.png}{CG\_20D\_corner\_def\_3}: $\Sigma_{i\neq j}$-deformation with $\epsilon=0.5$.
    \item \href{https://github.com/TwoSampleTests/GenerativeModelsMetrics/blob/main/results/20D_unimodal/corner_plot_def_4.png}{CG\_20D\_corner\_def\_4}: $\text{pow}^{+}$-deformation with $\epsilon=0.1$.
    \item \href{https://github.com/TwoSampleTests/GenerativeModelsMetrics/blob/main/results/20D_unimodal/corner_plot_def_5.png}{CG\_20D\_corner\_def\_5}: $\text{pow}^{-}$-deformation with $\epsilon=0.1$.
    \item \href{https://github.com/TwoSampleTests/GenerativeModelsMetrics/blob/main/results/20D_unimodal/corner_plot_def_6.png}{CG\_20D\_corner\_def\_6}: $\mathcal{N}$-deformation with $\epsilon=0.5$.
    \item \href{https://github.com/TwoSampleTests/GenerativeModelsMetrics/blob/main/results/20D_unimodal/corner_plot_def_7.png}{CG\_20D\_corner\_def\_7}: $\mathcal{U}$-deformation with $\epsilon=0.5$.
\end{itemize}

\noindent The following Figures report a color plot representation of the correlation matrices for two samples $\mathbf{X}_{1}$, drawn from the reference distribution of the CG model with $d=20$, and $\mathbf{X}_{2}$, corresponding to the specified deformation. The plots are made with $10^{6}$ points per sample.

\begin{itemize}
    \item \href{https://github.com/TwoSampleTests/GenerativeModelsMetrics/blob/main/results/20D_unimodal/corre_matrix_plot_def_1.pdf}{CG\_20D\_corr\_matrix\_def\_1}: $\mu$-deformation with $\epsilon = 0.5$.
    \item \href{https://github.com/TwoSampleTests/GenerativeModelsMetrics/blob/main/results/20D_unimodal/corre_matrix_plot_def_2.pdf}{CG\_20D\_corr\_matrix\_def\_2}: $\Sigma_{ii}$-deformation with $\epsilon = 0.5$.
    \item \href{https://github.com/TwoSampleTests/GenerativeModelsMetrics/blob/main/results/20D_unimodal/corre_matrix_plot_def_3.pdf}{CG\_20D\_corr\_matrix\_def\_3}: $\Sigma_{i\neq j}$-deformation with $\epsilon = 0.5$.
    \item \href{https://github.com/TwoSampleTests/GenerativeModelsMetrics/blob/main/results/20D_unimodal/corre_matrix_plot_def_4.pdf}{CG\_20D\_corr\_matrix\_def\_4}: $\text{pow}^{+}$-deformation with $\epsilon = 0.1$.
    \item \href{https://github.com/TwoSampleTests/GenerativeModelsMetrics/blob/main/results/20D_unimodal/corre_matrix_plot_def_5.pdf}{CG\_20D\_corr\_matrix\_def\_5}: $\text{pow}^{-}$-deformation with $\epsilon = 0.1$.
    \item \href{https://github.com/TwoSampleTests/GenerativeModelsMetrics/blob/main/results/20D_unimodal/corre_matrix_plot_def_6.pdf}{CG\_20D\_corr\_matrix\_def\_6}: $\mathcal{N}$-deformation with $\epsilon = 0.5$.
    \item \href{https://github.com/TwoSampleTests/GenerativeModelsMetrics/blob/main/results/20D_unimodal/corre_matrix_plot_def_7.pdf}{CG\_20D\_corr\_matrix\_def\_7}: $\mathcal{U}$-deformation with $\epsilon = 0.5$.
\end{itemize}

\noindent The following Figures show the empirical PDF of the specified test statistic under the null hypothesis (both samples come from the same reference distribution) for the CG model with $d=20$, and $n=m=(1,2,5,10)\cdot 10^{4}$. The test statistics distributions are obtained with $10^{4}$ iterations.

\begin{multicols}{3}
\begin{itemize}
    \item \href{https://github.com/TwoSampleTests/GenerativeModelsMetrics/tree/main/results/20D_unimodal/20D_10K/null_hypothesis/SWD.pdf}{CG\_20D\_10K\_null\_SW}\\
    \href{https://github.com/TwoSampleTests/GenerativeModelsMetrics/tree/main/results/20D_unimodal/20D_20K/null_hypothesis/SWD.pdf}{CG\_20D\_20K\_null\_SW}\\
    \href{https://github.com/TwoSampleTests/GenerativeModelsMetrics/tree/main/results/20D_unimodal/20D_50K/null_hypothesis/SWD.pdf}{CG\_20D\_50K\_null\_SW}\\
    \href{https://github.com/TwoSampleTests/GenerativeModelsMetrics/tree/main/results/20D_unimodal/20D_100K/null_hypothesis/SWD.pdf}{CG\_20D\_100K\_null\_SW}
    \item \href{https://github.com/TwoSampleTests/GenerativeModelsMetrics/tree/main/results/20D_unimodal/20D_10K/null_hypothesis/KS.pdf}{CG\_20D\_10K\_null\_$\overline{\mathrm{KS}}$}\\
    \href{https://github.com/TwoSampleTests/GenerativeModelsMetrics/tree/main/results/20D_unimodal/20D_20K/null_hypothesis/KS.pdf}{CG\_20D\_20K\_null\_$\overline{\mathrm{KS}}$}\\
    \href{https://github.com/TwoSampleTests/GenerativeModelsMetrics/tree/main/results/20D_unimodal/20D_50K/null_hypothesis/KS.pdf}{CG\_20D\_50K\_null\_$\overline{\mathrm{KS}}$}\\
    \href{https://github.com/TwoSampleTests/GenerativeModelsMetrics/tree/main/results/20D_unimodal/20D_100K/null_hypothesis/KS.pdf}{CG\_20D\_100K\_null\_$\overline{\mathrm{KS}}$}
    \item \href{https://github.com/TwoSampleTests/GenerativeModelsMetrics/tree/main/results/20D_unimodal/20D_10K/null_hypothesis/SKS.pdf}{CG\_20D\_10K\_null\_SKS}\\
    \href{https://github.com/TwoSampleTests/GenerativeModelsMetrics/tree/main/results/20D_unimodal/20D_20K/null_hypothesis/SKS.pdf}{CG\_20D\_20K\_null\_SKS}\\
    \href{https://github.com/TwoSampleTests/GenerativeModelsMetrics/tree/main/results/20D_unimodal/20D_50K/null_hypothesis/SKS.pdf}{CG\_20D\_50K\_null\_SKS}\\
    \href{https://github.com/TwoSampleTests/GenerativeModelsMetrics/tree/main/results/20D_unimodal/20D_100K/null_hypothesis/SKS.pdf}{CG\_20D\_100K\_null\_SKS}
    \item \href{https://github.com/TwoSampleTests/GenerativeModelsMetrics/tree/main/results/20D_unimodal/20D_10K/null_hypothesis/FGD.pdf}{CG\_20D\_10K\_null\_FGD}\\ 
    \href{https://github.com/TwoSampleTests/GenerativeModelsMetrics/tree/main/results/20D_unimodal/20D_20K/null_hypothesis/FGD.pdf}{CG\_20D\_20K\_null\_FGD}\\
    \href{https://github.com/TwoSampleTests/GenerativeModelsMetrics/tree/main/results/20D_unimodal/20D_50K/null_hypothesis/FGD.pdf}{CG\_20D\_50K\_null\_FGD}\\ 
    \href{https://github.com/TwoSampleTests/GenerativeModelsMetrics/tree/main/results/20D_unimodal/20D_100K/null_hypothesis/FGD.pdf}{CG\_20D\_100K\_null\_FGD} 
    \item \href{https://github.com/TwoSampleTests/GenerativeModelsMetrics/tree/main/results/20D_unimodal/20D_10K/null_hypothesis/MMD.pdf}{CG\_20D\_10K\_null\_MMD}\\
    \href{https://github.com/TwoSampleTests/GenerativeModelsMetrics/tree/main/results/20D_unimodal/20D_20K/null_hypothesis/MMD.pdf}{CG\_20D\_20K\_null\_MMD}\\
    \href{https://github.com/TwoSampleTests/GenerativeModelsMetrics/tree/main/results/20D_unimodal/20D_50K/null_hypothesis/MMD.pdf}{CG\_20D\_50K\_null\_MMD}\\
    \href{https://github.com/TwoSampleTests/GenerativeModelsMetrics/tree/main/results/20D_unimodal/20D_100K/null_hypothesis/MMD.pdf}{CG\_20D\_100K\_null\_MMD}\\\\
\end{itemize}
\end{multicols}

\subsubsection{List of Figures for CG results in $100D$}
The following Figures show the 1D and 2D marginal probability distributions for two samples $\mathbf{X}_{1}$, drawn from the reference distribution of the CG model with $d=100$, and $\mathbf{X}_{2}$, corresponding to the specified deformation. The plots are made with $10^{6}$ points per sample.

\begin{itemize}
    \item \href{https://github.com/TwoSampleTests/GenerativeModelsMetrics/blob/main/results/100D_unimodal/corner_plot_def_1.png}{CG\_100D\_corner\_def\_1}: $\mu$-deformation with $\epsilon=0.5$.
    \item \href{https://github.com/TwoSampleTests/GenerativeModelsMetrics/blob/main/results/100D_unimodal/corner_plot_def_2.png}{CG\_100D\_corner\_def\_2}: $\Sigma_{ii}$-deformation with $\epsilon=0.5$.
    \item \href{https://github.com/TwoSampleTests/GenerativeModelsMetrics/blob/main/results/100D_unimodal/corner_plot_def_3.png}{CG\_100D\_corner\_def\_3}: $\Sigma_{i\neq j}$-deformation with $\epsilon=0.5$.
    \item \href{https://github.com/TwoSampleTests/GenerativeModelsMetrics/blob/main/results/100D_unimodal/corner_plot_def_4.png}{CG\_100D\_corner\_def\_4}: $\text{pow}^{+}$-deformation with $\epsilon=0.1$.
    \item \href{https://github.com/TwoSampleTests/GenerativeModelsMetrics/blob/main/results/100D_unimodal/corner_plot_def_5.png}{CG\_100D\_corner\_def\_5}: $\text{pow}^{-}$-deformation with $\epsilon=0.1$.
    \item \href{https://github.com/TwoSampleTests/GenerativeModelsMetrics/blob/main/results/100D_unimodal/corner_plot_def_6.png}{CG\_100D\_corner\_def\_6}: $\mathcal{N}$-deformation with $\epsilon=0.5$.
    \item \href{https://github.com/TwoSampleTests/GenerativeModelsMetrics/blob/main/results/100D_unimodal/corner_plot_def_7.png}{CG\_100D\_corner\_def\_7}: $\mathcal{U}$-deformation with $\epsilon=0.5$.
\end{itemize}

\noindent The following Figures report a color plot representation of the correlation matrices for two samples $\mathbf{X}_{1}$, drawn from the reference distribution of the CG model with $d=100$, and $\mathbf{X}_{2}$, corresponding to the specified deformation. The plots are made with $10^{6}$ points per sample.

\begin{itemize}
    \item \href{https://github.com/TwoSampleTests/GenerativeModelsMetrics/blob/main/results/100D_unimodal/corre_matrix_plot_def_1.pdf}{CG\_100D\_corr\_matrix\_def\_1}: $\mu$-deformation with $\epsilon = 0.5$.
    \item \href{https://github.com/TwoSampleTests/GenerativeModelsMetrics/blob/main/results/100D_unimodal/corre_matrix_plot_def_2.pdf}{CG\_100D\_corr\_matrix\_def\_2}: $\Sigma_{ii}$-deformation with $\epsilon = 0.5$.
    \item \href{https://github.com/TwoSampleTests/GenerativeModelsMetrics/blob/main/results/100D_unimodal/corre_matrix_plot_def_3.pdf}{CG\_100D\_corr\_matrix\_def\_3}: $\Sigma_{i\neq j}$-deformation with $\epsilon = 0.5$.
    \item \href{https://github.com/TwoSampleTests/GenerativeModelsMetrics/blob/main/results/100D_unimodal/corre_matrix_plot_def_4.pdf}{CG\_100D\_corr\_matrix\_def\_4}: $\text{pow}^{+}$-deformation with $\epsilon = 0.1$.
    \item \href{https://github.com/TwoSampleTests/GenerativeModelsMetrics/blob/main/results/100D_unimodal/corre_matrix_plot_def_5.pdf}{CG\_100D\_corr\_matrix\_def\_5}: $\text{pow}^{-}$-deformation with $\epsilon = 0.1$.
    \item \href{https://github.com/TwoSampleTests/GenerativeModelsMetrics/blob/main/results/100D_unimodal/corre_matrix_plot_def_6.pdf}{CG\_100D\_corr\_matrix\_def\_6}: $\mathcal{N}$-deformation with $\epsilon = 0.5$.
    \item \href{https://github.com/TwoSampleTests/GenerativeModelsMetrics/blob/main/results/100D_unimodal/corre_matrix_plot_def_7.pdf}{CG\_100D\_corr\_matrix\_def\_7}: $\mathcal{U}$-deformation with $\epsilon = 0.5$.
\end{itemize}

\noindent The following Figures show the empirical PDF of the specified test statistic under the null hypothesis (both samples come from the same reference distribution) for the CG model with $d=100$, and $n=m=(1,2,5,10)\cdot 10^{4}$. The test statistics distributions are obtained with $10^{4}$ iterations.

\begin{multicols}{3}
    \begin{itemize}
        \item \href{https://github.com/TwoSampleTests/GenerativeModelsMetrics/tree/main/results/100D_unimodal/100D_10K/null_hypothesis/SWD.pdf}{CG\_100D\_10K\_null\_SW}\\
        \href{https://github.com/TwoSampleTests/GenerativeModelsMetrics/tree/main/results/100D_unimodal/100D_20K/null_hypothesis/SWD.pdf}{CG\_100D\_20K\_null\_SW}\\
        \href{https://github.com/TwoSampleTests/GenerativeModelsMetrics/tree/main/results/100D_unimodal/100D_50K/null_hypothesis/SWD.pdf}{CG\_100D\_50K\_null\_SW}\\
        \href{https://github.com/TwoSampleTests/GenerativeModelsMetrics/tree/main/results/100D_unimodal/100D_100K/null_hypothesis/SWD.pdf}{CG\_100D\_100K\_null\_SW}
        \item \href{https://github.com/TwoSampleTests/GenerativeModelsMetrics/tree/main/results/100D_unimodal/100D_10K/null_hypothesis/KS.pdf}{CG\_100D\_10K\_null\_$\overline{\mathrm{KS}}$}\\
        \href{https://github.com/TwoSampleTests/GenerativeModelsMetrics/tree/main/results/100D_unimodal/100D_20K/null_hypothesis/KS.pdf}{CG\_100D\_20K\_null\_$\overline{\mathrm{KS}}$}\\
        \href{https://github.com/TwoSampleTests/GenerativeModelsMetrics/tree/main/results/100D_unimodal/100D_50K/null_hypothesis/KS.pdf}{CG\_100D\_50K\_null\_$\overline{\mathrm{KS}}$}\\
        \href{https://github.com/TwoSampleTests/GenerativeModelsMetrics/tree/main/results/100D_unimodal/100D_100K/null_hypothesis/KS.pdf}{CG\_100D\_100K\_null\_$\overline{\mathrm{KS}}$}
        \item \href{https://github.com/TwoSampleTests/GenerativeModelsMetrics/tree/main/results/100D_unimodal/100D_10K/null_hypothesis/SKS.pdf}{CG\_100D\_10K\_null\_SKS}\\
        \href{https://github.com/TwoSampleTests/GenerativeModelsMetrics/tree/main/results/100D_unimodal/100D_20K/null_hypothesis/SKS.pdf}{CG\_100D\_20K\_null\_SKS}\\
        \href{https://github.com/TwoSampleTests/GenerativeModelsMetrics/tree/main/results/100D_unimodal/100D_50K/null_hypothesis/SKS.pdf}{CG\_100D\_50K\_null\_SKS}\\
        \href{https://github.com/TwoSampleTests/GenerativeModelsMetrics/tree/main/results/100D_unimodal/100D_100K/null_hypothesis/SKS.pdf}{CG\_100D\_100K\_null\_SKS}
        \item \href{https://github.com/TwoSampleTests/GenerativeModelsMetrics/tree/main/results/100D_unimodal/100D_10K/null_hypothesis/FGD.pdf}{CG\_100D\_10K\_null\_FGD}\\ 
        \href{https://github.com/TwoSampleTests/GenerativeModelsMetrics/tree/main/results/100D_unimodal/100D_20K/null_hypothesis/FGD.pdf}{CG\_100D\_20K\_null\_FGD}\\
        \href{https://github.com/TwoSampleTests/GenerativeModelsMetrics/tree/main/results/100D_unimodal/100D_50K/null_hypothesis/FGD.pdf}{CG\_100D\_50K\_null\_FGD}\\ 
        \href{https://github.com/TwoSampleTests/GenerativeModelsMetrics/tree/main/results/100D_unimodal/100D_100K/null_hypothesis/FGD.pdf}{CG\_100D\_100K\_null\_FGD} 
        \item \href{https://github.com/TwoSampleTests/GenerativeModelsMetrics/tree/main/results/100D_unimodal/100D_10K/null_hypothesis/MMD.pdf}{CG\_100D\_10K\_null\_MMD}\\
        \href{https://github.com/TwoSampleTests/GenerativeModelsMetrics/tree/main/results/100D_unimodal/100D_20K/null_hypothesis/MMD.pdf}{CG\_100D\_20K\_null\_MMD}\\
        \href{https://github.com/TwoSampleTests/GenerativeModelsMetrics/tree/main/results/100D_unimodal/100D_50K/null_hypothesis/MMD.pdf}{CG\_100D\_50K\_null\_MMD}\\
        \href{https://github.com/TwoSampleTests/GenerativeModelsMetrics/tree/main/results/100D_unimodal/100D_100K/null_hypothesis/MMD.pdf}{CG\_100D\_100K\_null\_MMD}\\\\
    \end{itemize}
    \end{multicols}

\subsection{List of Figures for the MoG models}
\subsubsection{List of Figures for MoG results in $5D$}
The following Figures show the 1D and 2D marginal probability distributions for two samples $\mathbf{X}_{1}$, drawn from the reference distribution of the MoG model with $d=5$ and $q=3$, and $\mathbf{X}_{2}$, corresponding to the specified deformation. The plots are made with $10^{6}$ points per sample.
    
\begin{itemize}
    \item \href{https://github.com/TwoSampleTests/GenerativeModelsMetrics/blob/main/results/5D_mixture/corner_plot_def_1.png}{MoG\_5D\_corner\_def\_1}: $\mu$-deformation with $\epsilon=0.5$.
    \item \href{https://github.com/TwoSampleTests/GenerativeModelsMetrics/blob/main/results/5D_mixture/corner_plot_def_2.png}{MoG\_5D\_corner\_def\_2}: $\Sigma_{ii}$-deformation with $\epsilon=0.5$.
    \item \href{https://github.com/TwoSampleTests/GenerativeModelsMetrics/blob/main/results/5D_mixture/corner_plot_def_3.png}{MoG\_5D\_corner\_def\_3}: $\Sigma_{i\neq j}$-deformation with $\epsilon=0.5$.
    \item \href{https://github.com/TwoSampleTests/GenerativeModelsMetrics/blob/main/results/5D_mixture/corner_plot_def_4.png}{MoG\_5D\_corner\_def\_4}: $\text{pow}^{+}$-deformation with $\epsilon=0.1$.
    \item \href{https://github.com/TwoSampleTests/GenerativeModelsMetrics/blob/main/results/5D_mixture/corner_plot_def_5.png}{MoG\_5D\_corner\_def\_5}: $\text{pow}^{-}$-deformation with $\epsilon=0.1$.
    \item \href{https://github.com/TwoSampleTests/GenerativeModelsMetrics/blob/main/results/5D_mixture/corner_plot_def_6.png}{MoG\_5D\_corner\_def\_6}: $\mathcal{N}$-deformation with $\epsilon=0.5$.
    \item \href{https://github.com/TwoSampleTests/GenerativeModelsMetrics/blob/main/results/5D_mixture/corner_plot_def_7.png}{MoG\_5D\_corner\_def\_7}: $\mathcal{U}$-deformation with $\epsilon=0.5$.
\end{itemize}

\noindent The following Figures report a color plot representation of the correlation matrices for two samples $\mathbf{X}_{1}$, drawn from the reference distribution of the MoG model with $d=5$ and $q=3$, and $\mathbf{X}_{2}$, corresponding to the specified deformation. The plots are made with $10^{6}$ points per sample.

\begin{itemize}
    \item \href{https://github.com/TwoSampleTests/GenerativeModelsMetrics/blob/main/results/5D_mixture/corre_matrix_plot_def_1.pdf}{MoG\_5D\_corr\_matrix\_def\_1}: $\mu$-deformation with $\epsilon = 0.5$.
    \item \href{https://github.com/TwoSampleTests/GenerativeModelsMetrics/blob/main/results/5D_mixture/corre_matrix_plot_def_2.pdf}{MoG\_5D\_corr\_matrix\_def\_2}: $\Sigma_{ii}$-deformation with $\epsilon = 0.5$.
    \item \href{https://github.com/TwoSampleTests/GenerativeModelsMetrics/blob/main/results/5D_mixture/corre_matrix_plot_def_3.pdf}{MoG\_5D\_corr\_matrix\_def\_3}: $\Sigma_{i\neq j}$-deformation with $\epsilon = 0.5$.
    \item \href{https://github.com/TwoSampleTests/GenerativeModelsMetrics/blob/main/results/5D_mixture/corre_matrix_plot_def_4.pdf}{MoG\_5D\_corr\_matrix\_def\_4}: $\text{pow}^{+}$-deformation with $\epsilon = 0.1$.
    \item \href{https://github.com/TwoSampleTests/GenerativeModelsMetrics/blob/main/results/5D_mixture/corre_matrix_plot_def_5.pdf}{MoG\_5D\_corr\_matrix\_def\_5}: $\text{pow}^{-}$-deformation with $\epsilon = 0.1$.
    \item \href{https://github.com/TwoSampleTests/GenerativeModelsMetrics/blob/main/results/5D_mixture/corre_matrix_plot_def_6.pdf}{MoG\_5D\_corr\_matrix\_def\_6}: $\mathcal{N}$-deformation with $\epsilon = 0.5$.
    \item \href{https://github.com/TwoSampleTests/GenerativeModelsMetrics/blob/main/results/5D_mixture/corre_matrix_plot_def_7.pdf}{MoG\_5D\_corr\_matrix\_def\_7}: $\mathcal{U}$-deformation with $\epsilon = 0.5$.
\end{itemize}

\noindent The following Figures show the empirical PDF of the specified test statistic under the null hypothesis (both samples come from the same reference distribution) for the MoG model with $d=5$, $q=3$, and $n=m=(1,2,5,10)\cdot 10^{4}$. The test statistics distributions are obtained with $10^{4}$ iterations.

\begin{multicols}{3}
\begin{itemize}
    \item \href{https://github.com/TwoSampleTests/GenerativeModelsMetrics/tree/main/results/5D_mixture/5D_10K/null_hypothesis/SWD.pdf}{MoG\_5D\_10K\_null\_SW}\\
    \href{https://github.com/TwoSampleTests/GenerativeModelsMetrics/tree/main/results/5D_mixture/5D_20K/null_hypothesis/SWD.pdf}{MoG\_5D\_20K\_null\_SW}\\
    \href{https://github.com/TwoSampleTests/GenerativeModelsMetrics/tree/main/results/5D_mixture/5D_50K/null_hypothesis/SWD.pdf}{MoG\_5D\_50K\_null\_SW}\\
    \href{https://github.com/TwoSampleTests/GenerativeModelsMetrics/tree/main/results/5D_mixture/5D_100K/null_hypothesis/SWD.pdf}{MoG\_5D\_100K\_null\_SW}
    \item \href{https://github.com/TwoSampleTests/GenerativeModelsMetrics/tree/main/results/5D_mixture/5D_10K/null_hypothesis/KS.pdf}{MoG\_5D\_10K\_null\_$\overline{\mathrm{KS}}$}\\
    \href{https://github.com/TwoSampleTests/GenerativeModelsMetrics/tree/main/results/5D_mixture/5D_20K/null_hypothesis/KS.pdf}{MoG\_5D\_20K\_null\_$\overline{\mathrm{KS}}$}\\
    \href{https://github.com/TwoSampleTests/GenerativeModelsMetrics/tree/main/results/5D_mixture/5D_50K/null_hypothesis/KS.pdf}{MoG\_5D\_50K\_null\_$\overline{\mathrm{KS}}$}\\
    \href{https://github.com/TwoSampleTests/GenerativeModelsMetrics/tree/main/results/5D_mixture/5D_100K/null_hypothesis/KS.pdf}{MoG\_5D\_100K\_null\_$\overline{\mathrm{KS}}$}
    \item \href{https://github.com/TwoSampleTests/GenerativeModelsMetrics/tree/main/results/5D_mixture/5D_10K/null_hypothesis/SKS.pdf}{MoG\_5D\_10K\_null\_SKS}\\
    \href{https://github.com/TwoSampleTests/GenerativeModelsMetrics/tree/main/results/5D_mixture/5D_20K/null_hypothesis/SKS.pdf}{MoG\_5D\_20K\_null\_SKS}\\
    \href{https://github.com/TwoSampleTests/GenerativeModelsMetrics/tree/main/results/5D_mixture/5D_50K/null_hypothesis/SKS.pdf}{MoG\_5D\_50K\_null\_SKS}\\
    \href{https://github.com/TwoSampleTests/GenerativeModelsMetrics/tree/main/results/5D_mixture/5D_100K/null_hypothesis/SKS.pdf}{MoG\_5D\_100K\_null\_SKS}
    \item \href{https://github.com/TwoSampleTests/GenerativeModelsMetrics/tree/main/results/5D_mixture/5D_10K/null_hypothesis/FGD.pdf}{MoG\_5D\_10K\_null\_FGD}\\ 
    \href{https://github.com/TwoSampleTests/GenerativeModelsMetrics/tree/main/results/5D_mixture/5D_20K/null_hypothesis/FGD.pdf}{MoG\_5D\_20K\_null\_FGD}\\
    \href{https://github.com/TwoSampleTests/GenerativeModelsMetrics/tree/main/results/5D_mixture/5D_50K/null_hypothesis/FGD.pdf}{MoG\_5D\_50K\_null\_FGD}\\ 
    \href{https://github.com/TwoSampleTests/GenerativeModelsMetrics/tree/main/results/5D_mixture/5D_100K/null_hypothesis/FGD.pdf}{MoG\_5D\_100K\_null\_FGD} 
    \item \href{https://github.com/TwoSampleTests/GenerativeModelsMetrics/tree/main/results/5D_mixture/5D_10K/null_hypothesis/MMD.pdf}{MoG\_5D\_10K\_null\_MMD}\\
    \href{https://github.com/TwoSampleTests/GenerativeModelsMetrics/tree/main/results/5D_mixture/5D_20K/null_hypothesis/MMD.pdf}{MoG\_5D\_20K\_null\_MMD}\\
    \href{https://github.com/TwoSampleTests/GenerativeModelsMetrics/tree/main/results/5D_mixture/5D_50K/null_hypothesis/MMD.pdf}{MoG\_5D\_50K\_null\_MMD}\\
    \href{https://github.com/TwoSampleTests/GenerativeModelsMetrics/tree/main/results/5D_mixture/5D_100K/null_hypothesis/MMD.pdf}{MoG\_5D\_100K\_null\_MMD}\\\\
\end{itemize}
\end{multicols}

\subsubsection{List of Figures for MoG results in $20D$}
The following Figures show the 1D and 2D marginal probability distributions for two samples $\mathbf{X}_{1}$, drawn from the reference distribution of the MoG model with $d=20$ and $q=5$, and $\mathbf{X}_{2}$, corresponding to the specified deformation. The plots are made with $10^{6}$ points per sample.

\begin{itemize}
    \item \href{https://github.com/TwoSampleTests/GenerativeModelsMetrics/blob/main/results/20D_mixture/corner_plot_def_1.png}{MoG\_20D\_corner\_def\_1}: $\mu$-deformation with $\epsilon=0.5$.
    \item \href{https://github.com/TwoSampleTests/GenerativeModelsMetrics/blob/main/results/20D_mixture/corner_plot_def_2.png}{MoG\_20D\_corner\_def\_2}: $\Sigma_{ii}$-deformation with $\epsilon=0.5$.
    \item \href{https://github.com/TwoSampleTests/GenerativeModelsMetrics/blob/main/results/20D_mixture/corner_plot_def_3.png}{MoG\_20D\_corner\_def\_3}: $\Sigma_{i\neq j}$-deformation with $\epsilon=0.5$.
    \item \href{https://github.com/TwoSampleTests/GenerativeModelsMetrics/blob/main/results/20D_mixture/corner_plot_def_4.png}{MoG\_20D\_corner\_def\_4}: $\text{pow}^{+}$-deformation with $\epsilon=0.1$.
    \item \href{https://github.com/TwoSampleTests/GenerativeModelsMetrics/blob/main/results/20D_mixture/corner_plot_def_5.png}{MoG\_20D\_corner\_def\_5}: $\text{pow}^{-}$-deformation with $\epsilon=0.1$.
    \item \href{https://github.com/TwoSampleTests/GenerativeModelsMetrics/blob/main/results/20D_mixture/corner_plot_def_6.png}{MoG\_20D\_corner\_def\_6}: $\mathcal{N}$-deformation with $\epsilon=0.5$.
    \item \href{https://github.com/TwoSampleTests/GenerativeModelsMetrics/blob/main/results/20D_mixture/corner_plot_def_7.png}{MoG\_20D\_corner\_def\_7}: $\mathcal{U}$-deformation with $\epsilon=0.5$.
\end{itemize}

\noindent The following Figures report a color plot representation of the correlation matrices for two samples $\mathbf{X}_{1}$, drawn from the reference distribution of the MoG model with $d=20$ and $q=5$, and $\mathbf{X}_{2}$, corresponding to the specified deformation. The plots are made with $10^{6}$ points per sample.

\begin{itemize}
    \item \href{https://github.com/TwoSampleTests/GenerativeModelsMetrics/blob/main/results/20D_mixture/corre_matrix_plot_def_1.pdf}{MoG\_20D\_corr\_matrix\_def\_1}: $\mu$-deformation with $\epsilon = 0.5$.
    \item \href{https://github.com/TwoSampleTests/GenerativeModelsMetrics/blob/main/results/20D_mixture/corre_matrix_plot_def_2.pdf}{MoG\_20D\_corr\_matrix\_def\_2}: $\Sigma_{ii}$-deformation with $\epsilon = 0.5$.
    \item \href{https://github.com/TwoSampleTests/GenerativeModelsMetrics/blob/main/results/20D_mixture/corre_matrix_plot_def_3.pdf}{MoG\_20D\_corr\_matrix\_def\_3}: $\Sigma_{i\neq j}$-deformation with $\epsilon = 0.5$.
    \item \href{https://github.com/TwoSampleTests/GenerativeModelsMetrics/blob/main/results/20D_mixture/corre_matrix_plot_def_4.pdf}{MoG\_20D\_corr\_matrix\_def\_4}: $\text{pow}^{+}$-deformation with $\epsilon = 0.1$.
    \item \href{https://github.com/TwoSampleTests/GenerativeModelsMetrics/blob/main/results/20D_mixture/corre_matrix_plot_def_5.pdf}{MoG\_20D\_corr\_matrix\_def\_5}: $\text{pow}^{-}$-deformation with $\epsilon = 0.1$.
    \item \href{https://github.com/TwoSampleTests/GenerativeModelsMetrics/blob/main/results/20D_mixture/corre_matrix_plot_def_6.pdf}{MoG\_20D\_corr\_matrix\_def\_6}: $\mathcal{N}$-deformation with $\epsilon = 0.5$.
    \item \href{https://github.com/TwoSampleTests/GenerativeModelsMetrics/blob/main/results/20D_mixture/corre_matrix_plot_def_7.pdf}{MoG\_20D\_corr\_matrix\_def\_7}: $\mathcal{U}$-deformation with $\epsilon = 0.5$.
\end{itemize}

\noindent The following Figures show the empirical PDF of the specified test statistic under the null hypothesis (both samples come from the same reference distribution) for the MoG model with $d=20$, $q=5$, and $n=m=(1,2,5,10)\cdot 10^{4}$. The test statistics distributions are obtained with $10^{4}$ iterations.

\begin{multicols}{3}
    \begin{itemize}
        \item \href{https://github.com/TwoSampleTests/GenerativeModelsMetrics/tree/main/results/20D_mixture/20D_10K/null_hypothesis/SWD.pdf}{MoG\_20D\_10K\_null\_SW}\\
        \href{https://github.com/TwoSampleTests/GenerativeModelsMetrics/tree/main/results/20D_mixture/20D_20K/null_hypothesis/SWD.pdf}{MoG\_20D\_20K\_null\_SW}\\
        \href{https://github.com/TwoSampleTests/GenerativeModelsMetrics/tree/main/results/20D_mixture/20D_50K/null_hypothesis/SWD.pdf}{MoG\_20D\_50K\_null\_SW}\\
        \href{https://github.com/TwoSampleTests/GenerativeModelsMetrics/tree/main/results/20D_mixture/20D_100K/null_hypothesis/SWD.pdf}{MoG\_20D\_100K\_null\_SW}
        \item \href{https://github.com/TwoSampleTests/GenerativeModelsMetrics/tree/main/results/20D_mixture/20D_10K/null_hypothesis/KS.pdf}{MoG\_20D\_10K\_null\_$\overline{\mathrm{KS}}$}\\
        \href{https://github.com/TwoSampleTests/GenerativeModelsMetrics/tree/main/results/20D_mixture/20D_20K/null_hypothesis/KS.pdf}{MoG\_20D\_20K\_null\_$\overline{\mathrm{KS}}$}\\
        \href{https://github.com/TwoSampleTests/GenerativeModelsMetrics/tree/main/results/20D_mixture/20D_50K/null_hypothesis/KS.pdf}{MoG\_20D\_50K\_null\_$\overline{\mathrm{KS}}$}\\
        \href{https://github.com/TwoSampleTests/GenerativeModelsMetrics/tree/main/results/20D_mixture/20D_100K/null_hypothesis/KS.pdf}{MoG\_20D\_100K\_null\_$\overline{\mathrm{KS}}$}
        \item \href{https://github.com/TwoSampleTests/GenerativeModelsMetrics/tree/main/results/20D_mixture/20D_10K/null_hypothesis/SKS.pdf}{MoG\_20D\_10K\_null\_SKS}\\
        \href{https://github.com/TwoSampleTests/GenerativeModelsMetrics/tree/main/results/20D_mixture/20D_20K/null_hypothesis/SKS.pdf}{MoG\_20D\_20K\_null\_SKS}\\
        \href{https://github.com/TwoSampleTests/GenerativeModelsMetrics/tree/main/results/20D_mixture/20D_50K/null_hypothesis/SKS.pdf}{MoG\_20D\_50K\_null\_SKS}\\
        \href{https://github.com/TwoSampleTests/GenerativeModelsMetrics/tree/main/results/20D_mixture/20D_100K/null_hypothesis/SKS.pdf}{MoG\_20D\_100K\_null\_SKS}
        \item \href{https://github.com/TwoSampleTests/GenerativeModelsMetrics/tree/main/results/20D_mixture/20D_10K/null_hypothesis/FGD.pdf}{MoG\_20D\_10K\_null\_FGD}\\ 
        \href{https://github.com/TwoSampleTests/GenerativeModelsMetrics/tree/main/results/20D_mixture/20D_20K/null_hypothesis/FGD.pdf}{MoG\_20D\_20K\_null\_FGD}\\
        \href{https://github.com/TwoSampleTests/GenerativeModelsMetrics/tree/main/results/20D_mixture/20D_50K/null_hypothesis/FGD.pdf}{MoG\_20D\_50K\_null\_FGD}\\ 
        \href{https://github.com/TwoSampleTests/GenerativeModelsMetrics/tree/main/results/20D_mixture/20D_100K/null_hypothesis/FGD.pdf}{MoG\_20D\_100K\_null\_FGD} 
        \item \href{https://github.com/TwoSampleTests/GenerativeModelsMetrics/tree/main/results/20D_mixture/20D_10K/null_hypothesis/MMD.pdf}{MoG\_20D\_10K\_null\_MMD}\\
        \href{https://github.com/TwoSampleTests/GenerativeModelsMetrics/tree/main/results/20D_mixture/20D_20K/null_hypothesis/MMD.pdf}{MoG\_20D\_20K\_null\_MMD}\\
        \href{https://github.com/TwoSampleTests/GenerativeModelsMetrics/tree/main/results/20D_mixture/20D_50K/null_hypothesis/MMD.pdf}{MoG\_20D\_50K\_null\_MMD}\\
        \href{https://github.com/TwoSampleTests/GenerativeModelsMetrics/tree/main/results/20D_mixture/20D_100K/null_hypothesis/MMD.pdf}{MoG\_20D\_100K\_null\_MMD}\\\\
    \end{itemize}
    \end{multicols}

\subsubsection{List of Figures for MoG results in $100D$}
The following Figures show the 1D and 2D marginal probability distributions for two samples $\mathbf{X}_{1}$, drawn from the reference distribution of the MoG model with $d=100$ and $q=10$, and $\mathbf{X}_{2}$, corresponding to the specified deformation. The plots are made with $10^{6}$ points per sample.

\begin{itemize}
    \item \href{https://github.com/TwoSampleTests/GenerativeModelsMetrics/blob/main/results/100D_mixture/corner_plot_def_1.png}{MoG\_100D\_corner\_def\_1}: $\mu$-deformation with $\epsilon=0.5$.
    \item \href{https://github.com/TwoSampleTests/GenerativeModelsMetrics/blob/main/results/100D_mixture/corner_plot_def_2.png}{MoG\_100D\_corner\_def\_2}: $\Sigma_{ii}$-deformation with $\epsilon=0.5$.
    \item \href{https://github.com/TwoSampleTests/GenerativeModelsMetrics/blob/main/results/100D_mixture/corner_plot_def_3.png}{MoG\_100D\_corner\_def\_3}: $\Sigma_{i\neq j}$-deformation with $\epsilon=0.5$.
    \item \href{https://github.com/TwoSampleTests/GenerativeModelsMetrics/blob/main/results/100D_mixture/corner_plot_def_4.png}{MoG\_100D\_corner\_def\_4}: $\text{pow}^{+}$-deformation with $\epsilon=0.1$.
    \item \href{https://github.com/TwoSampleTests/GenerativeModelsMetrics/blob/main/results/100D_mixture/corner_plot_def_5.png}{MoG\_100D\_corner\_def\_5}: $\text{pow}^{-}$-deformation with $\epsilon=0.1$.
    \item \href{https://github.com/TwoSampleTests/GenerativeModelsMetrics/blob/main/results/100D_mixture/corner_plot_def_6.png}{MoG\_100D\_corner\_def\_6}: $\mathcal{N}$-deformation with $\epsilon=0.5$.
    \item \href{https://github.com/TwoSampleTests/GenerativeModelsMetrics/blob/main/results/100D_mixture/corner_plot_def_7.png}{MoG\_100D\_corner\_def\_7}: $\mathcal{U}$-deformation with $\epsilon=0.5$.
\end{itemize}

\noindent The following Figures report a color plot representation of the correlation matrices for two samples $\mathbf{X}_{1}$, drawn from the reference distribution of the MoG model with $d=100$ and $q=10$, and $\mathbf{X}_{2}$, corresponding to the specified deformation. The plots are made with $10^{6}$ points per sample.

\begin{itemize}
    \item \href{https://github.com/TwoSampleTests/GenerativeModelsMetrics/blob/main/results/100D_mixture/corre_matrix_plot_def_1.pdf}{MoG\_100D\_corr\_matrix\_def\_1}: $\mu$-deformation with $\epsilon = 0.5$.
    \item \href{https://github.com/TwoSampleTests/GenerativeModelsMetrics/blob/main/results/100D_mixture/corre_matrix_plot_def_2.pdf}{MoG\_100D\_corr\_matrix\_def\_2}: $\Sigma_{ii}$-deformation with $\epsilon = 0.5$.
    \item \href{https://github.com/TwoSampleTests/GenerativeModelsMetrics/blob/main/results/100D_mixture/corre_matrix_plot_def_3.pdf}{MoG\_100D\_corr\_matrix\_def\_3}: $\Sigma_{i\neq j}$-deformation with $\epsilon = 0.5$.
    \item \href{https://github.com/TwoSampleTests/GenerativeModelsMetrics/blob/main/results/100D_mixture/corre_matrix_plot_def_4.pdf}{MoG\_100D\_corr\_matrix\_def\_4}: $\text{pow}^{+}$-deformation with $\epsilon = 0.1$.
    \item \href{https://github.com/TwoSampleTests/GenerativeModelsMetrics/blob/main/results/100D_mixture/corre_matrix_plot_def_5.pdf}{MoG\_100D\_corr\_matrix\_def\_5}: $\text{pow}^{-}$-deformation with $\epsilon = 0.1$.
    \item \href{https://github.com/TwoSampleTests/GenerativeModelsMetrics/blob/main/results/100D_mixture/corre_matrix_plot_def_6.pdf}{MoG\_100D\_corr\_matrix\_def\_6}: $\mathcal{N}$-deformation with $\epsilon = 0.5$.
    \item \href{https://github.com/TwoSampleTests/GenerativeModelsMetrics/blob/main/results/100D_mixture/corre_matrix_plot_def_7.pdf}{MoG\_100D\_corr\_matrix\_def\_7}: $\mathcal{U}$-deformation with $\epsilon = 0.5$.
\end{itemize}

\noindent The following Figures show the empirical PDF of the specified test statistic under the null hypothesis (both samples come from the same reference distribution) for the MoG model with $d=100$, $q=10$, and $n=m=(1,2,5,10)\cdot 10^{4}$. The test statistics distributions are obtained with $10^{4}$ iterations.

\begin{multicols}{3}
    \begin{itemize}
        \item \href{https://github.com/TwoSampleTests/GenerativeModelsMetrics/tree/main/results/100D_mixture/100D_10K/null_hypothesis/SWD.pdf}{MoG\_100D\_10K\_null\_SW}\\
        \href{https://github.com/TwoSampleTests/GenerativeModelsMetrics/tree/main/results/100D_mixture/100D_20K/null_hypothesis/SWD.pdf}{MoG\_100D\_20K\_null\_SW}\\
        \href{https://github.com/TwoSampleTests/GenerativeModelsMetrics/tree/main/results/100D_mixture/100D_50K/null_hypothesis/SWD.pdf}{MoG\_100D\_50K\_null\_SW}\\
        \href{https://github.com/TwoSampleTests/GenerativeModelsMetrics/tree/main/results/100D_mixture/100D_100K/null_hypothesis/SWD.pdf}{MoG\_100D\_100K\_null\_SW}
        \item \href{https://github.com/TwoSampleTests/GenerativeModelsMetrics/tree/main/results/100D_mixture/100D_10K/null_hypothesis/KS.pdf}{MoG\_100D\_10K\_null\_$\overline{\mathrm{KS}}$}\\
        \href{https://github.com/TwoSampleTests/GenerativeModelsMetrics/tree/main/results/100D_mixture/100D_20K/null_hypothesis/KS.pdf}{MoG\_100D\_20K\_null\_$\overline{\mathrm{KS}}$}\\
        \href{https://github.com/TwoSampleTests/GenerativeModelsMetrics/tree/main/results/100D_mixture/100D_50K/null_hypothesis/KS.pdf}{MoG\_100D\_50K\_null\_$\overline{\mathrm{KS}}$}\\
        \href{https://github.com/TwoSampleTests/GenerativeModelsMetrics/tree/main/results/100D_mixture/100D_100K/null_hypothesis/KS.pdf}{MoG\_100D\_100K\_null\_$\overline{\mathrm{KS}}$}
        \item \href{https://github.com/TwoSampleTests/GenerativeModelsMetrics/tree/main/results/100D_mixture/100D_10K/null_hypothesis/SKS.pdf}{MoG\_100D\_10K\_null\_SKS}\\
        \href{https://github.com/TwoSampleTests/GenerativeModelsMetrics/tree/main/results/100D_mixture/100D_20K/null_hypothesis/SKS.pdf}{MoG\_100D\_20K\_null\_SKS}\\
        \href{https://github.com/TwoSampleTests/GenerativeModelsMetrics/tree/main/results/100D_mixture/100D_50K/null_hypothesis/SKS.pdf}{MoG\_100D\_50K\_null\_SKS}\\
        \href{https://github.com/TwoSampleTests/GenerativeModelsMetrics/tree/main/results/100D_mixture/100D_100K/null_hypothesis/SKS.pdf}{MoG\_100D\_100K\_null\_SKS}
        \item \href{https://github.com/TwoSampleTests/GenerativeModelsMetrics/tree/main/results/100D_mixture/100D_10K/null_hypothesis/FGD.pdf}{MoG\_100D\_10K\_null\_FGD}\\ 
        \href{https://github.com/TwoSampleTests/GenerativeModelsMetrics/tree/main/results/100D_mixture/100D_20K/null_hypothesis/FGD.pdf}{MoG\_100D\_20K\_null\_FGD}\\
        \href{https://github.com/TwoSampleTests/GenerativeModelsMetrics/tree/main/results/100D_mixture/100D_50K/null_hypothesis/FGD.pdf}{MoG\_100D\_50K\_null\_FGD}\\ 
        \href{https://github.com/TwoSampleTests/GenerativeModelsMetrics/tree/main/results/100D_mixture/100D_100K/null_hypothesis/FGD.pdf}{MoG\_100D\_100K\_null\_FGD} 
        \item \href{https://github.com/TwoSampleTests/GenerativeModelsMetrics/tree/main/results/100D_mixture/100D_10K/null_hypothesis/MMD.pdf}{MoG\_100D\_10K\_null\_MMD}\\
        \href{https://github.com/TwoSampleTests/GenerativeModelsMetrics/tree/main/results/100D_mixture/100D_20K/null_hypothesis/MMD.pdf}{MoG\_100D\_20K\_null\_MMD}\\
        \href{https://github.com/TwoSampleTests/GenerativeModelsMetrics/tree/main/results/100D_mixture/100D_50K/null_hypothesis/MMD.pdf}{MoG\_100D\_50K\_null\_MMD}\\
        \href{https://github.com/TwoSampleTests/GenerativeModelsMetrics/tree/main/results/100D_mixture/100D_100K/null_hypothesis/MMD.pdf}{MoG\_100D\_100K\_null\_MMD}\\\\
    \end{itemize}
    \end{multicols}

\subsection{List of Figures for the JetNet at jet level}
The following Figures show the 1D and 2D marginal probability distributions for two samples $\mathbf{X}_{1}$, extracted from the JetNet reference distribution for jet level features in the gluon dataset, and $\mathbf{X}_{2}$, corresponding to the specified deformation. The plots are made with $10^{6}$ points per sample.

\begin{itemize}
    \item \href{https://github.com/TwoSampleTests/JetNetMetrics/blob/main/results/jet_features/figures/corner_plot_def_1.png}{jet\_corner\_def\_1}: $\mu$-deformation with $\epsilon=0.5$.
    \item \href{https://github.com/TwoSampleTests/JetNetMetrics/blob/main/results/jet_features/figures/corner_plot_def_2.png}{jet\_corner\_def\_2}: $\Sigma_{ii}$-deformation with $\epsilon=0.2$.
    \item \href{https://github.com/TwoSampleTests/JetNetMetrics/blob/main/results/jet_features/figures/corner_plot_def_3.png}{jet\_corner\_def\_3}: $\Sigma_{i\neq j}$-deformation with $\epsilon=0.2$.
    \item \href{https://github.com/TwoSampleTests/JetNetMetrics/blob/main/results/jet_features/figures/corner_plot_def_4.png}{jet\_corner\_def\_4}: $\text{pow}^{+}$-deformation with $\epsilon=0.1$.
    \item \href{https://github.com/TwoSampleTests/JetNetMetrics/blob/main/results/jet_features/figures/corner_plot_def_5.png}{jet\_corner\_def\_5}: $\text{pow}^{-}$-deformation with $\epsilon=0.1$.
    \item \href{https://github.com/TwoSampleTests/JetNetMetrics/blob/main/results/jet_features/figures/corner_plot_def_6.png}{jet\_corner\_def\_6}: $\mathcal{N}$-deformation with $\epsilon=0.2$.
    \item \href{https://github.com/TwoSampleTests/JetNetMetrics/blob/main/results/jet_features/figures/corner_plot_def_7.png}{jet\_corner\_def\_7}: $\mathcal{U}$-deformation with $\epsilon=0.2$.
\end{itemize}

\noindent The following Figures report a color plot representation of the correlation matrices for two samples $\mathbf{X}_{1}$, extracted from the JetNet reference distribution for jet level features in the gluon dataset, and $\mathbf{X}_{2}$, corresponding to the specified deformation. The plots are made with $10^{6}$ points per sample.

\begin{itemize}
    \item \href{https://github.com/TwoSampleTests/JetNetMetrics/blob/main/results/jet_features/figures/corre_matrix_plot_def_1.pdf}{jet\_corr\_matrix\_def\_1}: $\mu$-deformation with $\epsilon = 0.5$.
    \item \href{https://github.com/TwoSampleTests/JetNetMetrics/blob/main/results/jet_features/figures/corre_matrix_plot_def_2.pdf}{jet\_corr\_matrix\_def\_2}: $\Sigma_{ii}$-deformation with $\epsilon = 0.2$.
    \item \href{https://github.com/TwoSampleTests/JetNetMetrics/blob/main/results/jet_features/figures/corre_matrix_plot_def_3.pdf}{jet\_corr\_matrix\_def\_3}: $\Sigma_{i\neq j}$-deformation with $\epsilon = 0.2$.
    \item \href{https://github.com/TwoSampleTests/JetNetMetrics/blob/main/results/jet_features/figures/corre_matrix_plot_def_4.pdf}{jet\_corr\_matrix\_def\_4}: $\text{pow}^{+}$-deformation with $\epsilon = 0.1$.
    \item \href{https://github.com/TwoSampleTests/JetNetMetrics/blob/main/results/jet_features/figures/corre_matrix_plot_def_5.pdf}{jet\_corr\_matrix\_def\_5}: $\text{pow}^{-}$-deformation with $\epsilon = 0.1$.
    \item \href{https://github.com/TwoSampleTests/JetNetMetrics/blob/main/results/jet_features/figures/corre_matrix_plot_def_6.pdf}{jet\_corr\_matrix\_def\_6}: $\mathcal{N}$-deformation with $\epsilon = 0.2$.
    \item \href{https://github.com/TwoSampleTests/JetNetMetrics/blob/main/results/jet_features/figures/corre_matrix_plot_def_7.pdf}{jet\_corr\_matrix\_def\_7}: $\mathcal{U}$-deformation with $\epsilon = 0.2$.
\end{itemize}

\noindent The following Figures report a summary of the marginal distributions for two samples $\mathbf{X}_{1}$, extracted from the JetNet reference distribution for jet level features in the gluon dataset, and $\mathbf{X}_{2}$, corresponding to the specified deformation. The plots are made with $10^{6}$ points per sample.

\begin{itemize}
    \item \href{https://github.com/TwoSampleTests/JetNetMetrics/blob/main/results/jet_features/figures/deformed_jet_features_1.pdf}{jet\_deformation\_summary\_1}
    \item \href{https://github.com/TwoSampleTests/JetNetMetrics/blob/main/results/jet_features/figures/deformed_jet_features_2.pdf}{jet\_deformation\_summary\_2}
\end{itemize}

\noindent The following Figures show the empirical PDF of the specified test statistic under the null hypothesis (both samples come from the same reference distribution) for the JetNet distribution for jet level features in the gluon dataset, considering $n=m=(1,2,5)\cdot 10^{4}$. The test statistics distributions are obtained with $10^{3}$ iterations.

\begin{multicols}{3}
\begin{itemize}
    \item \href{https://github.com/TwoSampleTests/JetNetMetrics/blob/main/results/jet_features/tests/10K/null_hypotheses/FGD.pdf}{jet\_10K\_null\_FGD}\\ 
    \href{https://github.com/TwoSampleTests/JetNetMetrics/blob/main/results/jet_features/tests/20K/null_hypotheses/FGD.pdf}{jet\_20K\_null\_FGD}\\
    \href{https://github.com/TwoSampleTests/JetNetMetrics/blob/main/results/jet_features/tests/50K/null_hypotheses/FGD.pdf}{jet\_50K\_null\_FGD}
    \item \href{https://github.com/TwoSampleTests/JetNetMetrics/blob/main/results/jet_features/tests/10K/null_hypotheses/KS.pdf}{jet\_10K\_null\_KS}\\ 
    \href{https://github.com/TwoSampleTests/JetNetMetrics/blob/main/results/jet_features/tests/20K/null_hypotheses/KS.pdf}{jet\_20K\_null\_KS}\\
    \href{https://github.com/TwoSampleTests/JetNetMetrics/blob/main/results/jet_features/tests/50K/null_hypotheses/KS.pdf}{jet\_50K\_null\_KS} 
    \item \href{https://github.com/TwoSampleTests/JetNetMetrics/blob/main/results/jet_features/tests/10K/null_hypotheses/MMD.pdf}{jet\_10K\_null\_MMD}\\ 
    \href{https://github.com/TwoSampleTests/JetNetMetrics/blob/main/results/jet_features/tests/20K/null_hypotheses/MMD.pdf}{jet\_20K\_null\_MMD}\\
    \href{https://github.com/TwoSampleTests/JetNetMetrics/blob/main/results/jet_features/tests/50K/null_hypotheses/MMD.pdf}{jet\_50K\_null\_MMD}
    \item \href{https://github.com/TwoSampleTests/JetNetMetrics/blob/main/results/jet_features/tests/10K/null_hypotheses/SKS.pdf}{jet\_10K\_null\_SKS}\\ 
    \href{https://github.com/TwoSampleTests/JetNetMetrics/blob/main/results/jet_features/tests/20K/null_hypotheses/SKS.pdf}{jet\_20K\_null\_SKS}\\
    \href{https://github.com/TwoSampleTests/JetNetMetrics/blob/main/results/jet_features/tests/50K/null_hypotheses/SKS.pdf}{jet\_50K\_null\_SKS}
    \item \href{https://github.com/TwoSampleTests/JetNetMetrics/blob/main/results/jet_features/tests/10K/null_hypotheses/SWD.pdf}{jet\_10K\_null\_SWD}\\ 
    \href{https://github.com/TwoSampleTests/JetNetMetrics/blob/main/results/jet_features/tests/20K/null_hypotheses/SWD.pdf}{jet\_20K\_null\_SWD}\\
    \href{https://github.com/TwoSampleTests/JetNetMetrics/blob/main/results/jet_features/tests/50K/null_hypotheses/SWD.pdf}{jet\_50K\_null\_SWD} \\\\
\end{itemize}
\end{multicols}

\subsection{List of Figures for the JetNet at particle level}


\noindent The following Figures report a color plot representation of the correlation matrices for two samples $\mathbf{X}_{1}$, extracted from the JetNet reference distribution for particle level features including the first $30$ particles (ordered in $p_{T}$) in the gluon dataset, and $\mathbf{X}_{2}$, corresponding to the specified deformation. The plots are made with $10^{6}$ points per sample.

\begin{itemize}
    \item \href{https://github.com/TwoSampleTests/JetNetMetrics/blob/main/results/particle_features_30/figures/corre_matrix_plot_def_1.pdf}{part\_30\_corr\_matrix\_def\_1}: $\mu$-deformation with $\epsilon = 0.5$.
    \item \href{https://github.com/TwoSampleTests/JetNetMetrics/blob/main/results/particle_features_30/figures/corre_matrix_plot_def_2.pdf}{part\_30\_corr\_matrix\_def\_2}: $\Sigma_{ii}$-deformation with $\epsilon = 0.2$.
    \item \href{https://github.com/TwoSampleTests/JetNetMetrics/blob/main/results/particle_features_30/figures/corre_matrix_plot_def_3.pdf}{part\_30\_corr\_matrix\_def\_3}: $\Sigma_{i\neq j}$-deformation with $\epsilon = 0.2$.
    \item \href{https://github.com/TwoSampleTests/JetNetMetrics/blob/main/results/particle_features_30/figures/corre_matrix_plot_def_4.pdf}{part\_30\_corr\_matrix\_def\_4}: $\text{pow}^{+}$-deformation with $\epsilon = 0.1$.
    \item \href{https://github.com/TwoSampleTests/JetNetMetrics/blob/main/results/particle_features_30/figures/corre_matrix_plot_def_5.pdf}{part\_30\_corr\_matrix\_def\_5}: $\text{pow}^{-}$-deformation with $\epsilon = 0.1$.
    \item \href{https://github.com/TwoSampleTests/JetNetMetrics/blob/main/results/particle_features_30/figures/corre_matrix_plot_def_6.pdf}{part\_30\_corr\_matrix\_def\_6}: $\mathcal{N}$-deformation with $\epsilon = 0.2$.
    \item \href{https://github.com/TwoSampleTests/JetNetMetrics/blob/main/results/particle_features_30/figures/corre_matrix_plot_def_7.pdf}{part\_30\_corr\_matrix\_def\_7}: $\mathcal{U}$-deformation with $\epsilon = 0.2$.
\end{itemize}

\noindent The following Figures report a summary of the marginal distributions for two samples $\mathbf{X}_{1}$, extracted from the JetNet reference distribution for particle level features including the first $30$ particles (ordered in $p_{T}$) in the gluon dataset, and $\mathbf{X}_{2}$, corresponding to the specified deformation. The plots are made with $10^{6}$ points per sample.

\begin{itemize}
    \item \href{https://github.com/TwoSampleTests/JetNetMetrics/blob/main/results/particle_features_30/figures/deformed_particle_features_1.pdf}{part\_30\_deformation\_summary\_1}
    \item \href{https://github.com/TwoSampleTests/JetNetMetrics/blob/main/results/particle_features_30/figures/deformed_particle_features_2.pdf}{part\_30\_deformation\_summary\_2}
\end{itemize}

\noindent The following Figures show the empirical PDF of the specified test statistic under the null hypothesis (both samples come from the same reference distribution) for the JetNet distribution for particle level features with $30$ particles in the gluon dataset, considering $n=m=(1,2,5)\cdot 10^{4}$. The test statistics distributions are obtained with $10^{3}$ iterations.

\begin{multicols}{3}
\begin{itemize}
    \item \href{https://github.com/TwoSampleTests/JetNetMetrics/blob/main/results/particle_features_30/tests/10K/null_hypotheses/FGD.pdf}{part\_30\_10K\_null\_FGD}\\ 
    \href{https://github.com/TwoSampleTests/JetNetMetrics/blob/main/results/particle_features_30/tests/20K/null_hypotheses/FGD.pdf}{part\_30\_20K\_null\_FGD}\\
    \href{https://github.com/TwoSampleTests/JetNetMetrics/blob/main/results/particle_features_30/tests/50K/null_hypotheses/FGD.pdf}{part\_30\_50K\_null\_FGD}
    \item \href{https://github.com/TwoSampleTests/JetNetMetrics/blob/main/results/particle_features_30/tests/10K/null_hypotheses/KS.pdf}{part\_30\_10K\_null\_KS}\\ 
    \href{https://github.com/TwoSampleTests/JetNetMetrics/blob/main/results/particle_features_30/tests/20K/null_hypotheses/KS.pdf}{part\_30\_20K\_null\_KS}\\
    \href{https://github.com/TwoSampleTests/JetNetMetrics/blob/main/results/particle_features_30/tests/50K/null_hypotheses/KS.pdf}{part\_30\_50K\_null\_KS}
    \item \href{https://github.com/TwoSampleTests/JetNetMetrics/blob/main/results/particle_features_30/tests/10K/null_hypotheses/MMD.pdf}{part\_30\_10K\_null\_MMD}\\ 
    \href{https://github.com/TwoSampleTests/JetNetMetrics/blob/main/results/particle_features_30/tests/20K/null_hypotheses/MMD.pdf}{part\_30\_20K\_null\_MMD}\\
    \href{https://github.com/TwoSampleTests/JetNetMetrics/blob/main/results/particle_features_30/tests/50K/null_hypotheses/MMD.pdf}{part\_30\_50K\_null\_MMD}
    \item \href{https://github.com/TwoSampleTests/JetNetMetrics/blob/main/results/particle_features_30/tests/10K/null_hypotheses/SKS.pdf}{part\_30\_10K\_null\_SKS}\\ 
    \href{https://github.com/TwoSampleTests/JetNetMetrics/blob/main/results/particle_features_30/tests/20K/null_hypotheses/SKS.pdf}{part\_30\_20K\_null\_SKS}\\
    \href{https://github.com/TwoSampleTests/JetNetMetrics/blob/main/results/particle_features_30/tests/50K/null_hypotheses/SKS.pdf}{part\_30\_50K\_null\_SKS}
    \item \href{https://github.com/TwoSampleTests/JetNetMetrics/blob/main/results/particle_features_30/tests/10K/null_hypotheses/SWD.pdf}{part\_30\_10K\_null\_SWD}\\ 
    \href{https://github.com/TwoSampleTests/JetNetMetrics/blob/main/results/particle_features_30/tests/20K/null_hypotheses/SWD.pdf}{part\_30\_20K\_null\_SWD}\\
    \href{https://github.com/TwoSampleTests/JetNetMetrics/blob/main/results/particle_features_30/tests/50K/null_hypotheses/SWD.pdf}{part\_30\_50K\_null\_SWD} \\\\
\end{itemize}
\end{multicols}

\section{List of Tables}\label{app:tables}

\subsection{List of Tables for the CG models}\label{app:tables_CG}

\noindent The following Tables summarize the results for the CG model with $d=5$, $20$, and $100$, for different $n=m$ number of samples. The tables show, for the different deformations considered in the CG models, the values of $\epsilon$ for which the null hypothesis is rejected at the $95\%$ and $99\%$ confidence levels, and the time in seconds needed to compute them with percent-level error given the null hypothesis distributions. The null hypothesis distribution was evaluated with $n_{\mathrm{iter}}=10^{4}$ iterations, while the value of the test statistics under the alternative hypothesis was estimated from the mean of $n_{\mathrm{iter}}=100$ iterations. The last column shows the computing time of the test statistics distributions under the null hypothesis, which is independent of $\epsilon$ for all the test statistics but $t_{\mathrm{LLR}}$.

\begin{multicols}{3}
\begin{itemize}
    \item \href{https://github.com/TwoSampleTests/GenerativeModelsMetrics/blob/main/results/5D_unimodal/5D_10K/results_table.pdf}{CG\_5D\_10K}\\
    \href{https://github.com/TwoSampleTests/GenerativeModelsMetrics/blob/main/results/5D_unimodal/5D_20K/results_table.pdf}{CG\_5D\_20K}\\
    \href{https://github.com/TwoSampleTests/GenerativeModelsMetrics/blob/main/results/5D_unimodal/5D_50K/results_table.pdf}{CG\_5D\_50K}\\
    \href{https://github.com/TwoSampleTests/GenerativeModelsMetrics/blob/main/results/5D_unimodal/5D_100K/results_table.pdf}{CG\_5D\_100K}
    \item \href{https://github.com/TwoSampleTests/GenerativeModelsMetrics/blob/main/results/20D_unimodal/20D_10K/results_table.pdf}{CG\_20D\_10K}\\
    \href{https://github.com/TwoSampleTests/GenerativeModelsMetrics/blob/main/results/20D_unimodal/20D_20K/results_table.pdf}{CG\_20D\_20K}\\
    \href{https://github.com/TwoSampleTests/GenerativeModelsMetrics/blob/main/results/20D_unimodal/20D_50K/results_table.pdf}{CG\_20D\_50K}\\
    \href{https://github.com/TwoSampleTests/GenerativeModelsMetrics/blob/main/results/20D_unimodal/20D_100K/results_table.pdf}{CG\_20D\_100K}
    \item \href{https://github.com/TwoSampleTests/GenerativeModelsMetrics/blob/main/results/100D_unimodal/100D_10K/results_table.pdf}{CG\_100D\_10K}\\
    \href{https://github.com/TwoSampleTests/GenerativeModelsMetrics/blob/main/results/100D_unimodal/100D_20K/results_table.pdf}{CG\_100D\_20K}\\
    \href{https://github.com/TwoSampleTests/GenerativeModelsMetrics/blob/main/results/100D_unimodal/100D_50K/results_table.pdf}{CG\_100D\_50K}\\
    \href{https://github.com/TwoSampleTests/GenerativeModelsMetrics/blob/main/results/100D_unimodal/100D_100K/results_table.pdf}{CG\_100D\_100K}
\end{itemize}
\end{multicols}

\subsection{List of Tables for the MoG models}
\noindent The following Tables summarize the results for the MoG model with $d=5$, $q=3$, $d=20$, $q=5$, and $d=100$, $q=10$, for different $n=m$ number of samples. The tables show, for the different deformations considered in the MoG models, the values of $\epsilon$ for which the null hypothesis is rejected at the $95\%$ and $99\%$ confidence levels, and the time in seconds needed to compute them with percent-level error given the null hypothesis distributions. The null hypothesis distribution was evaluated with $n_{\mathrm{iter}}=10^{4}$ iterations, while the value of the test statistics under the alternative hypothesis was estimated from the mean of $n_{\mathrm{iter}}=100$ iterations. The last column shows the computing time of the test statistics distributions under the null hypothesis, which is independent of $\epsilon$ for all the test statistics but $t_{\mathrm{LLR}}$.

\begin{multicols}{3}
    \begin{itemize}
        \item \href{https://github.com/TwoSampleTests/GenerativeModelsMetrics/blob/main/results/5D_mixture/5D_10K/results_table.pdf}{MoG\_5D\_10K}\\
        \href{https://github.com/TwoSampleTests/GenerativeModelsMetrics/blob/main/results/5D_mixture/5D_20K/results_table.pdf}{MoG\_5D\_20K}\\
        \href{https://github.com/TwoSampleTests/GenerativeModelsMetrics/blob/main/results/5D_mixture/5D_50K/results_table.pdf}{MoG\_5D\_50K}\\
        \href{https://github.com/TwoSampleTests/GenerativeModelsMetrics/blob/main/results/5D_mixture/5D_100K/results_table.pdf}{MoG\_5D\_100K}
        \item \href{https://github.com/TwoSampleTests/GenerativeModelsMetrics/blob/main/results/20D_mixture/20D_10K/results_table.pdf}{MoG\_20D\_10K}\\
        \href{https://github.com/TwoSampleTests/GenerativeModelsMetrics/blob/main/results/20D_mixture/20D_20K/results_table.pdf}{MoG\_20D\_20K}\\
        \href{https://github.com/TwoSampleTests/GenerativeModelsMetrics/blob/main/results/20D_mixture/20D_50K/results_table.pdf}{MoG\_20D\_50K}\\
        \href{https://github.com/TwoSampleTests/GenerativeModelsMetrics/blob/main/results/20D_mixture/20D_100K/results_table.pdf}{MoG\_20D\_100K}
        \item \href{https://github.com/TwoSampleTests/GenerativeModelsMetrics/blob/main/results/100D_mixture/100D_10K/results_table.pdf}{MoG\_100D\_10K}\\
        \href{https://github.com/TwoSampleTests/GenerativeModelsMetrics/blob/main/results/100D_mixture/100D_20K/results_table.pdf}{MoG\_100D\_20K}\\
        \href{https://github.com/TwoSampleTests/GenerativeModelsMetrics/blob/main/results/100D_mixture/100D_50K/results_table.pdf}{MoG\_100D\_50K}\\
        \href{https://github.com/TwoSampleTests/GenerativeModelsMetrics/blob/main/results/100D_mixture/100D_100K/results_table.pdf}{MoG\_100D\_100K}
    \end{itemize}
    \end{multicols}

    \subsection{List of Tables for the JetNet at jet level}

    \noindent The following Tables summarize the results for the JetNet gluon dataset at the jet level for different $n=m$ number of samples. The tables show, for the different deformations of the JetNet reference distribution for jet level features in the gluon dataset, the values of $\epsilon$ for which the null hypothesis is rejected at the $95\%$ and $99\%$ confidence levels, and the time in seconds needed to compute them with percent-level error given the null hypothesis distributions. The null hypothesis distribution was evaluated with $n_{\mathrm{iter}}=10^{3}$ iterations, while the value of the test statistics under the alternative hypothesis was estimated from the mean of $n_{\mathrm{iter}}=100$ iterations. The last column shows the computing time of the test statistics distributions under the null hypothesis, which is independent of $\epsilon$ for all the test statistics but $t_{\mathrm{LLR}}$. To better understand the effect of scaling features (to zero mean and unit standard deviation) on the performance of the tests, we also show the results for the same deformations but with the test performed on scaled features. The latter results are denoted with a suffix `scaled'.
    
    \begin{multicols}{2}
    \begin{itemize}
        \item \href{https://github.com/TwoSampleTests/JetNetMetrics/blob/main/results/jet_features/tests/10K/results_table.pdf}{jet\_10K}\\
        \href{https://github.com/TwoSampleTests/JetNetMetrics/blob/main/results/jet_features/tests/20K/results_table.pdf}{jet\_20K}\\
        \href{https://github.com/TwoSampleTests/JetNetMetrics/blob/main/results/jet_features/tests/50K/results_table.pdf}{jet\_50K}\\
        \item \href{https://github.com/TwoSampleTests/JetNetMetrics/blob/main/results/jet_features/tests/10K_preprocessed/results_table.pdf}{jet\_10K\_scaled}\\
        \href{https://github.com/TwoSampleTests/JetNetMetrics/blob/main/results/jet_features/tests/20K_preprocessed/results_table.pdf}{jet\_20K\_scaled}\\
        \href{https://github.com/TwoSampleTests/JetNetMetrics/blob/main/results/jet_features/tests/50K_preprocessed/results_table.pdf}{jet\_50K\_scaled}
    \end{itemize}
    \end{multicols}

    \subsection{List of Tables for the JetNet at particle level}

    \noindent The following Tables summarize the results for the JetNet gluon dataset at the particle level, including the first $30$ particles (ordered in $p_{T}$), for different $n=m$ number of samples. The tables show, for the different deformations of the JetNet reference distribution for particle level features in the gluon dataset, the values of $\epsilon$ for which the null hypothesis is rejected at the $95\%$ and $99\%$ confidence levels, and the time in seconds needed to compute them with percent-level error given the null hypothesis distributions. The null hypothesis distribution was evaluated with $n_{\mathrm{iter}}=10^{3}$ iterations, while the value of the test statistics under the alternative hypothesis was estimated from the mean of $n_{\mathrm{iter}}=100$ iterations. The last column shows the computing time of the test statistics distributions under the null hypothesis, which is independent of $\epsilon$ for all the test statistics but $t_{\mathrm{LLR}}$. To better understand the effect of scaling features (to zero mean and unit standard deviation) on the performance of the tests, we also show the results for the same deformations but with the test performed on scaled features. The latter results are denoted with a suffix `scaled'.
    
    \begin{multicols}{2}
        \begin{itemize}
            \item \href{https://github.com/TwoSampleTests/JetNetMetrics/blob/main/results/particle_features_30/tests/10K/results_table.pdf}{part\_30\_10K}\\
            \href{https://github.com/TwoSampleTests/JetNetMetrics/blob/main/results/particle_features_30/tests/20K/results_table.pdf}{part\_30\_20K}\\
            \href{https://github.com/TwoSampleTests/JetNetMetrics/blob/main/results/particle_features_30/tests/50K/results_table.pdf}{part\_30\_50K}\\
            \item \href{https://github.com/TwoSampleTests/JetNetMetrics/blob/main/results/particle_features_30/tests/10K_preprocessed/results_table.pdf}{part\_30\_10K\_scaled}\\
            \href{https://github.com/TwoSampleTests/JetNetMetrics/blob/main/results/particle_features_30/tests/20K_preprocessed/results_table.pdf}{part\_30\_20K\_scaled}\\
            \href{https://github.com/TwoSampleTests/JetNetMetrics/blob/main/results/particle_features_30/tests/50K_preprocessed/results_table.pdf}{part\_30\_50K\_scaled}
        \end{itemize}
        \end{multicols}
    
\twocolumn
\clearpage
\bibliographystyle{mine}
\bibliography{references}

\end{document}

%% file: Tables/MoG_20D_50K_results_table.tex
\begin{tabular}{l|llr|llr}
	\toprule
	\multicolumn{7}{c}{{\bf MoG model with $\mathbf{d=20}$, $\mathbf{q=5}$, and $\mathbf{n=m=5\cdot 10^{4}}$}} \\
	\toprule
	\multicolumn{1}{c}{} & \multicolumn{3}{c}{$\mu$-deformation} & \multicolumn{3}{c}{$\Sigma_{ii}$-deformation} \\
	Statistic & $\epsilon_{95\%\mathrm{CL}}$ & $\epsilon_{99\%\mathrm{CL}}$ & $t$ (s) & $\epsilon_{95\%\mathrm{CL}}$ & $\epsilon_{99\%\mathrm{CL}}$ & $t$ (s) \\
	\midrule
	$t_{\mathrm{SW}}$ & $0.04957_{-0.02}^{+0.018}$ & $0.06694_{-0.017}^{+0.017}$ & $3023$ & $0.01679_{-0.0063}^{+0.005}$ & $0.02315_{-0.005}^{+0.0045}$ & $3197$ \\
	$t_{\overline{\mathrm{KS}}}$ & ${\mathbf{0.00482_{-0.0018}^{+0.0013}}}$ & ${\mathbf{0.00667_{-0.0013}^{+0.0011}}}$ & $2966$ & ${\mathbf{0.00175_{-0.00068}^{+0.00052}}}$ & ${\mathbf{0.00248_{-0.00052}^{+0.00042}}}$ & $3185$ \\
	$t_{\mathrm{SKS}}$ & $0.03647_{-0.014}^{+0.011}$ & $0.04821_{-0.012}^{+0.011}$ & ${\mathbf{2899}}$ & $0.01329_{-0.0043}^{+0.003}$ & $0.01759_{-0.003}^{+0.0025}$ & ${\mathbf{3022}}$ \\
	$t_{\mathrm{FGD}}$ & $0.05778_{-0.027}^{+0.026}$ & $0.0787_{-0.021}^{+0.023}$ & $4047$ & $0.01945_{-0.0081}^{+0.0063}$ & $0.02651_{-0.0056}^{+0.0053}$ & $4507$ \\
	$t_{\mathrm{MMD}}$ & $0.04425_{-0.018}^{+0.019}$ & $0.06215_{-0.015}^{+0.017}$ & $10204$ & $0.00923_{-0.0051}^{+0.0058}$ & $0.01305_{-0.0044}^{+0.0053}$ & $11217$ \\
	$t_{\mathrm{LLR}}$ & $0.00021_{-0.00014}^{+0.00013}$ & $0.0003_{-0.00014}^{+0.00013}$ & $5911$ & $0.00007_{-0.00004}^{+0.00005}$ & $0.0001_{-0.00004}^{+0.00005}$ & $6304$ \\
	\toprule
	\multicolumn{1}{c}{} & \multicolumn{3}{c}{$\Sigma_{i\neq j}$-deformation} & \multicolumn{3}{c}{$\rm{pow}_{+}$-deformation} \\
	Statistic & $\epsilon_{95\%\mathrm{CL}}$ & $\epsilon_{99\%\mathrm{CL}}$ & $t$ (s) & $\epsilon_{95\%\mathrm{CL}}$ & $\epsilon_{99\%\mathrm{CL}}$ & $t$ (s) \\
	\midrule
	$t_{\mathrm{SW}}$ & $0.02162_{-0.008}^{+0.0056}$ & $0.02935_{-0.0055}^{+0.0045}$ & ${\mathbf{3410}}$ & $0.00581_{-0.0022}^{+0.0017}$ & $0.00798_{-0.0017}^{+0.0015}$ & ${\mathbf{3157}}$ \\
	$t_{\overline{\mathrm{KS}}}$ & $1.00146_{-0.00031}^{+0.00074}$ & $1.00238_{-0.00031}^{+0.00055}$ & $3967$ & ${\mathbf{0.0004_{-0.00017}^{+0.00015}}}$ & ${\mathbf{0.00059_{-0.00014}^{+0.00013}}}$ & $3363$ \\
	$t_{\mathrm{SKS}}$ & $0.02306_{-0.0088}^{+0.0071}$ & $0.03079_{-0.0072}^{+0.0062}$ & $3553$ & $0.0043_{-0.0013}^{+0.0009}$ & $0.00565_{-0.0009}^{+0.00074}$ & $3193$ \\
	$t_{\mathrm{FGD}}$ & ${\mathbf{0.00551_{-0.002}^{+0.0015}}}$ & ${\mathbf{0.00748_{-0.0013}^{+0.0013}}}$ & $6327$ & $0.00702_{-0.0028}^{+0.0021}$ & $0.00965_{-0.0019}^{+0.0016}$ & $4870$ \\
	$t_{\mathrm{MMD}}$ & $0.01723_{-0.0072}^{+0.008}$ & $0.02431_{-0.0064}^{+0.0069}$ & $11450$ & $0.00332_{-0.0017}^{+0.0018}$ & $0.00467_{-0.0014}^{+0.0017}$ & $11801$ \\
	$t_{\mathrm{LLR}}$ & - & - & - & $0.00002_{-0.00001}^{+0.00001}$ & $0.00002_{-0.00001}^{+0.00001}$ & $6877$ \\
	\toprule
	\multicolumn{1}{c}{} & \multicolumn{3}{c}{$\rm{pow}_{-}$-deformation} & \multicolumn{3}{c}{$\mathcal{N}$-deformation} \\
	Statistic & $\epsilon_{95\%\mathrm{CL}}$ & $\epsilon_{99\%\mathrm{CL}}$ & $t$ (s) & $\epsilon_{95\%\mathrm{CL}}$ & $\epsilon_{99\%\mathrm{CL}}$ & $t$ (s) \\
	\midrule
	$t_{\mathrm{SW}}$ & $0.00604_{-0.0023}^{+0.0017}$ & $0.00825_{-0.0018}^{+0.0016}$ & ${\mathbf{3051}}$ & $0.19318_{-0.039}^{+0.025}$ & $0.22704_{-0.026}^{+0.019}$ & ${\mathbf{2403}}$ \\
	$t_{\overline{\mathrm{KS}}}$ & ${\mathbf{0.00042_{-0.00018}^{+0.00015}}}$ & ${\mathbf{0.00061_{-0.00015}^{+0.00013}}}$ & $3372$ & ${\mathbf{0.00751_{-0.0024}^{+0.002}}}$ & ${\mathbf{0.00993_{-0.002}^{+0.0018}}}$ & $2934$ \\
	$t_{\mathrm{SKS}}$ & $0.00441_{-0.0014}^{+0.00092}$ & $0.00574_{-0.00094}^{+0.00077}$ & $3324$ & $0.15874_{-0.034}^{+0.023}$ & $0.18473_{-0.023}^{+0.019}$ & $2726$ \\
	$t_{\mathrm{FGD}}$ & $0.00722_{-0.0027}^{+0.0021}$ & $0.00987_{-0.0019}^{+0.0016}$ & $4892$ & $0.18095_{-0.038}^{+0.023}$ & $0.21269_{-0.02}^{+0.016}$ & $3756$ \\
	$t_{\mathrm{MMD}}$ & $0.00353_{-0.0015}^{+0.0016}$ & $0.00494_{-0.0012}^{+0.0014}$ & $11418$ & $0.43531_{-0.11}^{+0.066}$ & $0.51609_{-0.054}^{+0.045}$ & $8642$ \\
	$t_{\mathrm{LLR}}$ & $0.00002_{-0.00001}^{+0.00001}$ & $0.00002_{-0.00001}^{+0.00001}$ & $6991$ & - & - & - \\
	\toprule
	\multicolumn{1}{c}{} & \multicolumn{3}{c}{$\mathcal{U}$-deformation} & \multicolumn{3}{c}{Timing} \\
	Statistic & $\epsilon_{95\%\mathrm{CL}}$ & $\epsilon_{99\%\mathrm{CL}}$ & $t$ (s) & $t^{\mathrm{null}}$ (s) \\
	\midrule
	$t_{\mathrm{SW}}$ & $0.33394_{-0.068}^{+0.044}$ & $0.39248_{-0.044}^{+0.033}$ & ${\mathbf{2354}}$ & $338$ \\
	$t_{\overline{\mathrm{KS}}}$ & ${\mathbf{0.01211_{-0.0035}^{+0.003}}}$ & ${\mathbf{0.01575_{-0.003}^{+0.0027}}}$ & $2835$ & ${\mathbf{155}}$ \\
	$t_{\mathrm{SKS}}$ & $0.27395_{-0.059}^{+0.041}$ & $0.3188_{-0.04}^{+0.033}$ & $2601$ & $509$ \\
	$t_{\mathrm{FGD}}$ & $0.31409_{-0.07}^{+0.04}$ & $0.36919_{-0.036}^{+0.027}$ & $3643$ & $2795$ \\
	$t_{\mathrm{MMD}}$ & $0.75353_{-0.18}^{+0.12}$ & $0.89336_{-0.098}^{+0.078}$ & $7700$ & $13860$ \\
	$t_{\mathrm{LLR}}$ & - & - & - & - \\
	\bottomrule
\end{tabular}

%% file: Tables/CG_20D_50K_results_table.tex
\begin{tabular}{l|llr|llr}
	\toprule
	\multicolumn{7}{c}{{\bf CG model with $\mathbf{d=20}$ and $\mathbf{n=m=5\cdot 10^{4}}$}} \\
	\toprule
	\multicolumn{1}{c}{} & \multicolumn{3}{c}{$\mu$-deformation} & \multicolumn{3}{c}{$\Sigma_{ii}$-deformation} \\
	Statistic & $\epsilon_{95\%\mathrm{CL}}$ & $\epsilon_{99\%\mathrm{CL}}$ & $t$ (s) & $\epsilon_{95\%\mathrm{CL}}$ & $\epsilon_{99\%\mathrm{CL}}$ & $t$ (s) \\
	\midrule
	$t_{\mathrm{SW}}$ & $0.04948_{-0.021}^{+0.022}$ & $0.06621_{-0.02}^{+0.021}$ & $571$ & $0.02059_{-0.0078}^{+0.0066}$ & $0.02732_{-0.0065}^{+0.0061}$ & $617$ \\
	$t_{\overline{\mathrm{KS}}}$ & ${\mathbf{0.04811_{-0.021}^{+0.022}}}$ & $0.06605_{-0.02}^{+0.021}$ & ${\mathbf{407}}$ & $0.02898_{-0.012}^{+0.011}$ & $0.04029_{-0.01}^{+0.0097}$ & ${\mathbf{434}}$ \\
	$t_{\mathrm{SKS}}$ & $0.04841_{-0.021}^{+0.021}$ & ${\mathbf{0.06372_{-0.02}^{+0.02}}}$ & $655$ & $0.02623_{-0.01}^{+0.0087}$ & $0.03417_{-0.0086}^{+0.0082}$ & $694$ \\
	$t_{\mathrm{FGD}}$ & $0.05029_{-0.022}^{+0.026}$ & $0.06539_{-0.02}^{+0.024}$ & $1886$ & ${\mathbf{0.01695_{-0.007}^{+0.007}}}$ & ${\mathbf{0.02215_{-0.0059}^{+0.0065}}}$ & $1994$ \\
	$t_{\mathrm{MMD}}$ & $0.0596_{-0.02}^{+0.028}$ & $0.08041_{-0.02}^{+0.026}$ & $7733$ & $0.02325_{-0.0079}^{+0.011}$ & $0.03109_{-0.0079}^{+0.01}$ & $8173$ \\
	$t_{\mathrm{LLR}}$ & $0.00556_{-0.003}^{+0.0031}$ & $0.00795_{-0.003}^{+0.003}$ & $2441$ & $0.00153_{-0.00098}^{+0.001}$ & $0.0022_{-0.00099}^{+0.00098}$ & $3081$ \\
	\toprule
	\multicolumn{1}{c}{} & \multicolumn{3}{c}{$\Sigma_{i\neq j}$-deformation} & \multicolumn{3}{c}{$\rm{pow}_{+}$-deformation} \\
	Statistic & $\epsilon_{95\%\mathrm{CL}}$ & $\epsilon_{99\%\mathrm{CL}}$ & $t$ (s) & $\epsilon_{95\%\mathrm{CL}}$ & $\epsilon_{99\%\mathrm{CL}}$ & $t$ (s) \\
	\midrule
	$t_{\mathrm{SW}}$ & $0.02783_{-0.0099}^{+0.0077}$ & $0.03884_{-0.0076}^{+0.0064}$ & ${\mathbf{1073}}$ & $0.0046_{-0.0019}^{+0.0017}$ & $0.00614_{-0.0017}^{+0.0016}$ & $642$ \\
	$t_{\overline{\mathrm{KS}}}$ & $1.02831_{-0.015}^{+0.015}$ & $1.04211_{-0.012}^{+0.0046}$ & $1401$ & $0.00602_{-0.0024}^{+0.002}$ & $0.00806_{-0.0019}^{+0.0019}$ & ${\mathbf{459}}$ \\
	$t_{\mathrm{SKS}}$ & $0.03839_{-0.014}^{+0.011}$ & $0.05106_{-0.012}^{+0.01}$ & $1172$ & $0.00505_{-0.002}^{+0.0017}$ & $0.00646_{-0.0017}^{+0.0016}$ & $747$ \\
	$t_{\mathrm{FGD}}$ & ${\mathbf{0.00483_{-0.0014}^{+0.0012}}}$ & ${\mathbf{0.00631_{-0.0011}^{+0.0011}}}$ & $3433$ & $0.00419_{-0.0018}^{+0.0019}$ & $0.0054_{-0.0015}^{+0.0017}$ & $2765$ \\
	$t_{\mathrm{MMD}}$ & $0.03094_{-0.013}^{+0.017}$ & $0.04245_{-0.013}^{+0.016}$ & $8963$ & ${\mathbf{0.00358_{-0.0012}^{+0.0018}}}$ & ${\mathbf{0.00483_{-0.0013}^{+0.0016}}}$ & $8839$ \\
	$t_{\mathrm{LLR}}$ & - & - & - & $0.00042_{-0.00026}^{+0.00025}$ & $0.00061_{-0.00025}^{+0.00025}$ & $2919$ \\
	\toprule
	\multicolumn{1}{c}{} & \multicolumn{3}{c}{$\rm{pow}_{-}$-deformation} & \multicolumn{3}{c}{$\mathcal{N}$-deformation} \\
	Statistic & $\epsilon_{95\%\mathrm{CL}}$ & $\epsilon_{99\%\mathrm{CL}}$ & $t$ (s) & $\epsilon_{95\%\mathrm{CL}}$ & $\epsilon_{99\%\mathrm{CL}}$ & $t$ (s) \\
	\midrule
	$t_{\mathrm{SW}}$ & $0.00455_{-0.0017}^{+5}$ & $0.00609_{-0.0015}^{+5}$ & $587$ & $0.28641_{-0.065}^{+0.047}$ & $0.33654_{-0.046}^{+0.037}$ & $535$ \\
	$t_{\overline{\mathrm{KS}}}$ & $0.00575_{-0.0022}^{+0.002}$ & $0.00773_{-0.0019}^{+0.0018}$ & ${\mathbf{461}}$ & $0.32182_{-0.08}^{+0.055}$ & $0.3832_{-0.054}^{+0.045}$ & ${\mathbf{393}}$ \\
	$t_{\mathrm{SKS}}$ & $0.00487_{-0.0019}^{+0.0017}$ & $0.00632_{-0.0017}^{+0.0016}$ & $750$ & $0.28237_{-0.066}^{+0.046}$ & $0.32811_{-0.048}^{+0.038}$ & $612$ \\
	$t_{\mathrm{FGD}}$ & $0.00411_{-0.0015}^{+0.0017}$ & $0.0054_{-0.0014}^{+0.0015}$ & $2758$ & ${\mathbf{0.16992_{-0.03}^{+0.02}}}$ & ${\mathbf{0.1944_{-0.018}^{+0.014}}}$ & $2132$ \\
	$t_{\mathrm{MMD}}$ & ${\mathbf{0.00346_{-0.0014}^{+0.0019}}}$ & ${\mathbf{0.00477_{-0.0014}^{+0.0018}}}$ & $8990$ & $0.73852_{-0.091}^{+0.086}$ & $0.85602_{-0.062}^{+0.075}$ & $5790$ \\
	$t_{\mathrm{LLR}}$ & $0.00042_{-0.00026}^{+0.00025}$ & $0.0006_{-0.00025}^{+0.00025}$ & $2930$ & - & - & - \\
	\toprule
	\multicolumn{1}{c}{} & \multicolumn{3}{c}{$\mathcal{U}$-deformation} & \multicolumn{3}{c}{Timing} \\
	Statistic & $\epsilon_{95\%\mathrm{CL}}$ & $\epsilon_{99\%\mathrm{CL}}$ & $t$ (s) & $t^{\mathrm{null}}$ (s) \\
	\midrule
	$t_{\mathrm{SW}}$ & $0.49513_{-0.11}^{+0.079}$ & $0.5818_{-0.078}^{+0.063}$ & $512$ & $313$ \\
	$t_{\overline{\mathrm{KS}}}$ & $0.55562_{-0.14}^{+0.096}$ & $0.65585_{-0.089}^{+0.083}$ & ${\mathbf{378}}$ & ${\mathbf{127}}$ \\
	$t_{\mathrm{SKS}}$ & $0.48849_{-0.11}^{+0.085}$ & $0.56476_{-0.079}^{+0.072}$ & $582$ & $480$ \\
	$t_{\mathrm{FGD}}$ & ${\mathbf{0.2926_{-0.05}^{+0.036}}}$ & ${\mathbf{0.33697_{-0.034}^{+0.025}}}$ & $2042$ & $3821$ \\
	$t_{\mathrm{MMD}}$ & $1.28521_{-0.17}^{+0.15}$ & $1.49004_{-0.12}^{+0.11}$ & $6502$ & $13843$ \\
	$t_{\mathrm{LLR}}$ & - & - & - & - \\
	\bottomrule
\end{tabular}

%% file: Tables/jet_features_50K_results_table.tex
\begin{tabular}{l|llr|llr}
	\toprule
	\multicolumn{7}{c}{{\bf Jet features with $\mathbf{n=m=5\cdot 10^{4}}$}} \\
	\toprule
	\multicolumn{1}{c}{} & \multicolumn{3}{c}{$\mu$-deformation} & \multicolumn{3}{c}{$\Sigma_{ii}$-deformation} \\
	Statistic & $\epsilon_{95\%\mathrm{CL}}$ & $\epsilon_{99\%\mathrm    {CL}}$ & $t$ (s) & $\epsilon_{95\%\mathrm{CL}}$ & $\epsilon_{99\%\mathrm{CL}}$ & $t$ (s) \\
	\midrule
	$t_{\mathrm{SW}}$ & $0.03049_{-0.013}^{+0.019}$ & $0.04713_{-0.015}^{+0.015}$ & ${\mathbf{1108}}$ & $0.04623_{-0.025}^{+0.017}$ & $0.06323_{-0.015}^{+0.019}$ & ${\mathbf{1141}}$ \\
	$t_{\overline{\mathrm{KS}}}$ & ${\mathbf{0.01585_{-0.0063}^{+0.0043}}}$ & ${\mathbf{0.01927_{-0.0056}^{+0.0043}}}$ & $17004$ & ${\mathbf{0.02085_{-0.008}^{+0.0064}}}$ & ${\mathbf{0.02567_{-0.0075}^{+0.006}}}$ & $21589$ \\
	$t_{\mathrm{SKS}}$ & $0.02815_{-0.014}^{+0.013}$ & $0.03444_{-0.014}^{+0.012}$ & $35328$ & $0.04838_{-0.019}^{+0.018}$ & $0.06304_{-0.02}^{+0.016}$ & $27128$ \\
	$t_{\mathrm{FGD}}$ & $0.03986_{-0.013}^{+0.025}$ & $0.06157_{-0.017}^{+0.019}$ & $11779$ & $0.04333_{-0.023}^{+0.028}$ & $0.05934_{-0.022}^{+0.027}$ & $18470$ \\
	$t_{\mathrm{MMD}}$ & $0.04941_{-0.021}^{+0.034}$ & $0.0712_{-0.022}^{+0.03}$ & $78077$ & $0.07669_{-0.035}^{+0.072}$ & $0.11237_{-0.035}^{+0.068}$ & $71427$ \\
	\toprule
	\multicolumn{1}{c}{} & \multicolumn{3}{c}{$\Sigma_{i\neq j}$-deformation} & \multicolumn{3}{c}{$\rm{pow}_{+}$-deformation} \\
	Statistic & $\epsilon_{95\%\mathrm{CL}}$ & $\epsilon_{99\%\mathrm{CL}}$ & $t$ (s) & $\epsilon_{95\%\mathrm{CL}}$ & $\epsilon^{\rm   {pow}_{+}}_{99\%\mathrm{CL}}$ & $t$ (s) \\
	\midrule
	$t_{\mathrm{SW}}$ & $0.30801_{-0.11}^{+0.08}$ & $0.45956_{-0.063}^{+0.071}$ & ${\mathbf{1033}}$ & $0.02535_{-0.011}^{+0.0077}$ & $0.03745_{-0.0084}^{+0.0066}$ & ${\mathbf{1028}}$ \\
	$t_{\overline{\mathrm{KS}}}$ & $1.01892_{-0.01}^{+0.0084}$ & $1.02245_{-0.0035}^{+0.011}$ & $19934$ & ${\mathbf{0.0232_{-0.011}^{+0.0074}}}$ & ${\mathbf{0.02698_{-0.0092}^{+0.01}}}$ & $35049$ \\
	$t_{\mathrm{SKS}}$ & $0.2959_{-0.12}^{+0.12}$ & $0.40074_{-0.054}^{+0.11}$ & $32727$ & $0.02709_{-0.012}^{+0.014}$ & $0.03452_{-0.012}^{+0.017}$ & $28409$ \\
	$t_{\mathrm{FGD}}$ & ${\mathbf{0.22063_{-0.082}^{+0.053}}}$ & ${\mathbf{0.29862_{-0.052}^{+0.045}}}$ & $13459$ & $0.02454_{-0.014}^{+0.015}$ & $0.0321_{-0.012}^{+0.017}$ & $11640$ \\
	$t_{\mathrm{MMD}}$ & $0.80374_{-0.28}^{+0.26}$ & $1.05932_{-0.1}^{+0.078}$ & $31136$ & $0.02933_{-0.015}^{+0.019}$ & $0.03749_{-0.016}^{+0.021}$ & $54684$ \\
	\toprule
	\multicolumn{1}{c}{} & \multicolumn{3}{c}{$\rm{pow}_{-}$-deformation} & \multicolumn{3}{c}{$\mathcal{N}$-deformation} \\
	Statistic & $\epsilon_{95\%\mathrm{CL}}$ & $\epsilon^  {\rm{pow}_{-}}_{99\%\mathrm{CL}}$ & $t$ (s) & $\epsilon_{95\%\mathrm{CL}}$ & $\epsilon^    {\mathcal{N}}_{99\%\mathrm{CL}}$ & $t$ (s) \\
	\midrule
	$t_{\mathrm{SW}}$ & $0.02553_{-0.0088}^{+0.0078}$ & $0.03665_{-0.0068}^{+0.0074}$ & ${\mathbf{1080}}$ & $0.12904_{-0.034}^{+0.029}$ & $0.16235_{-0.025}^{+0.02}$ & ${\mathbf{981}}$ \\
	$t_{\overline{\mathrm{KS}}}$ & ${\mathbf{0.02125_{-0.0092}^{+0.01}}}$ & ${\mathbf{0.02649_{-0.009}^{+0.0074}}}$ & $15925$ & ${\mathbf{0.10579_{-0.019}^{+0.014}}}$ & ${\mathbf{0.11672_{-0.016}^{+0.012}}}$ & $28786$ \\
	$t_{\mathrm{SKS}}$ & $0.02682_{-0.012}^{+0.012}$ & $0.03607_{-0.012}^{+0.01}$ & $47622$ & $0.11163_{-0.023}^{+0.022}$ & $0.12765_{-0.023}^{+0.017}$ & $38615$ \\
	$t_{\mathrm{FGD}}$ & $0.02511_{-0.012}^{+0.017}$ & $0.03353_{-0.01}^{+0.016}$ & $18451$ & $0.16887_{-0.052}^{+0.046}$ & $0.19783_{-0.036}^{+0.043}$ & $13634$ \\
	$t_{\mathrm{MMD}}$ & $0.03_{-0.014}^{+0.02}$ & $0.04112_{-0.012}^{+0.021}$ & $39156$ & $0.25305_{-0.11}^{+0.085}$ & $0.29551_{-0.073}^{+0.081}$ & $52861$ \\
	\toprule
	\multicolumn{1}{c}{} & \multicolumn{3}{c}{$\mathcal{U}$-deformation} & \multicolumn{3}{c}{Timing} \\
	Statistic & $\epsilon_{95\%\mathrm{CL}}$ & $\epsilon^    {\mathcal{U}}_{99\%\mathrm{CL}}$ & $t$ (s) & $t^{\mathrm{null}}$ (s) \\
	\midrule
	$t_{\mathrm{SW}}$ & $0.22631_{-0.064}^{+0.05}$ & $0.27734_{-0.039}^{+0.044}$ & ${\mathbf{916}}$ & ${\mathbf{129}}$ \\
	$t_{\overline{\mathrm{KS}}}$ & ${\mathbf{0.18246_{-0.032}^{+0.022}}}$ & ${\mathbf{0.19931_{-0.027}^{+0.018}}}$ & $32276$ & $1907$ \\
	$t_{\mathrm{SKS}}$ & $0.18837_{-0.048}^{+0.033}$ & $0.21334_{-0.029}^{+0.027}$ & $38491$ & $4382$ \\
	$t_{\mathrm{FGD}}$ & $0.27796_{-0.074}^{+0.1}$ & $0.34469_{-0.062}^{+0.068}$ & $19098$ & $1794$ \\
	$t_{\mathrm{MMD}}$ & $0.49303_{-0.18}^{+0.16}$ & $0.57279_{-0.11}^{+0.12}$ & $55838$ & $3504$ \\
	\bottomrule
\end{tabular}

%% file: Tables/jet_features_50K_scaled_results_table.tex
\begin{tabular}{l|llr|llr}
	\toprule
	\multicolumn{7}{c}{{\bf Scaled Jet features with $\mathbf{n=m=5\cdot 10^{4}}$}} \\
	\toprule
	\multicolumn{1}{c}{} & \multicolumn{3}{c}{$\mu$-deformation} & \multicolumn{3}{c}{$\Sigma_{ii}$-deformation} \\
	Statistic & $\epsilon_{95\%\mathrm{CL}}$ & $\epsilon_{99\%\mathrm    {CL}}$ & $t$ (s) & $\epsilon_{95\%\mathrm{CL}}$ & $\epsilon_{99\%\mathrm{CL}}$ & $t$ (s) \\
	\midrule
	$t_{\mathrm{SW}}$ & $0.01623_{-0.0069}^{+0.0045}$ & $0.02098_{-0.0059}^{+0.0049}$ & $12410$ & $0.02089_{-0.008}^{+0.0073}$ & $0.02834_{-0.0079}^{+0.0077}$ & ${\mathbf{1054}}$ \\
	$t_{\overline{\mathrm{KS}}}$ & $0.01585_{-0.0063}^{+0.0043}$ & $0.01927_{-0.0056}^{+0.0043}$ & $17174$ & ${\mathbf{0.02085_{-0.008}^{+0.0064}}}$ & ${\mathbf{0.02567_{-0.0075}^{+0.006}}}$ & $38871$ \\
	$t_{\mathrm{SKS}}$ & ${\mathbf{0.0113_{-0.005}^{+0.0044}}}$ & ${\mathbf{0.0141_{-0.0045}^{+0.0037}}}$ & $32620$ & $0.02254_{-0.0099}^{+0.0074}$ & $0.02773_{-0.0089}^{+0.0073}$ & $28803$ \\
	$t_{\mathrm{FGD}}$ & $0.02106_{-0.0079}^{+0.0062}$ & $0.02659_{-0.0069}^{+0.0058}$ & ${\mathbf{11583}}$ & $0.02133_{-0.0097}^{+0.0078}$ & $0.02741_{-0.008}^{+0.0071}$ & $14254$ \\
	$t_{\mathrm{MMD}}$ & $0.06739_{-0.021}^{+0.013}$ & $0.08802_{-0.011}^{+0.013}$ & $46972$ & $0.0318_{-0.0083}^{+0.015}$ & $0.04328_{-0.012}^{+0.014}$ & $28709$ \\
	\toprule
	\multicolumn{1}{c}{} & \multicolumn{3}{c}{$\Sigma_{i\neq j}$-deformation} & \multicolumn{3}{c}{$\rm{pow}_{+}$-deformation} \\
	Statistic & $\epsilon_{95\%\mathrm{CL}}$ & $\epsilon_{99\%\mathrm{CL}}$ & $t$ (s) & $\epsilon_{95\%\mathrm{CL}}$ & $\epsilon^{\rm   {pow}_{+}}_{99\%\mathrm{CL}}$ & $t$ (s) \\
	\midrule
	$t_{\mathrm{SW}}$ & $0.0503_{-0.019}^{+0.016}$ & $0.07052_{-0.014}^{+0.015}$ & ${\mathbf{1008}}$ & $0.02465_{-0.0081}^{+0.011}$ & $0.03314_{-0.0095}^{+0.0099}$ & ${\mathbf{1025}}$ \\
	$t_{\overline{\mathrm{KS}}}$ & $1.02009_{-0.001}^{+0.0072}$ & $1.02812_{-0.008}^{+0.003}$ & $16410$ & $0.0232_{-0.011}^{+0.0074}$ & $0.02698_{-0.0092}^{+0.01}$ & $35198$ \\
	$t_{\mathrm{SKS}}$ & $0.06201_{-0.029}^{+0.02}$ & $0.07573_{-0.024}^{+0.02}$ & $35383$ & $0.0402_{-0.015}^{+0.015}$ & $0.04921_{-0.015}^{+0.015}$ & $47807$ \\
	$t_{\mathrm{FGD}}$ & ${\mathbf{0.00627_{-0.0018}^{+0.0016}}}$ & ${\mathbf{0.00809_{-0.0018}^{+0.0015}}}$ & $14008$ & $0.02237_{-0.011}^{+0.013}$ & $0.0281_{-0.0084}^{+0.011}$ & $24967$ \\
	$t_{\mathrm{MMD}}$ & $0.0794_{-0.031}^{+0.039}$ & $0.112_{-0.026}^{+0.031}$ & $29620$ & ${\mathbf{0.01898_{-0.0094}^{+0.012}}}$ & ${\mathbf{0.02472_{-0.0076}^{+0.012}}}$ & $66075$ \\
	\toprule
	\multicolumn{1}{c}{} & \multicolumn{3}{c}{$\rm{pow}_{-}$-deformation} & \multicolumn{3}{c}{$\mathcal{N}$-deformation} \\
	Statistic & $\epsilon_{95\%\mathrm{CL}}$ & $\epsilon^  {\rm{pow}_{-}}_{99\%\mathrm{CL}}$ & $t$ (s) & $\epsilon_{95\%\mathrm{CL}}$ & $\epsilon^    {\mathcal{N}}_{99\%\mathrm{CL}}$ & $t$ (s) \\
	\midrule
	$t_{\mathrm{SW}}$ & $0.02527_{-0.011}^{+0.011}$ & $0.03513_{-0.01}^{+0.0084}$ & ${\mathbf{993}}$ & $0.11836_{-0.028}^{+0.027}$ & $0.14062_{-0.026}^{+0.018}$ & ${\mathbf{910}}$ \\
	$t_{\overline{\mathrm{KS}}}$ & ${\mathbf{0.02125_{-0.0092}^{+0.01}}}$ & ${\mathbf{0.02649_{-0.009}^{+0.0074}}}$ & $16472$ & $0.10579_{-0.019}^{+0.014}$ & $0.11672_{-0.016}^{+0.012}$ & $31727$ \\
	$t_{\mathrm{SKS}}$ & $0.03986_{-0.017}^{+0.013}$ & $0.04873_{-0.013}^{+0.013}$ & $27407$ & $0.08577_{-0.028}^{+0.024}$ & $0.10148_{-0.026}^{+0.021}$ & $25899$ \\
	$t_{\mathrm{FGD}}$ & $0.02163_{-0.0097}^{+0.015}$ & $0.02954_{-0.0087}^{+0.014}$ & $12892$ & ${\mathbf{0.07833_{-0.019}^{+0.0094}}}$ & ${\mathbf{0.08847_{-0.0069}^{+0.0084}}}$ & $13246$ \\
	$t_{\mathrm{MMD}}$ & $0.02133_{-0.0086}^{+0.013}$ & $0.02924_{-0.0081}^{+0.011}$ & $68458$ & $0.26032_{-0.057}^{+0.037}$ & $0.29897_{-0.036}^{+0.028}$ & $42149$ \\
	\toprule
	\multicolumn{1}{c}{} & \multicolumn{3}{c}{$\mathcal{U}$-deformation} & \multicolumn{3}{c}{Timing} \\
	Statistic & $\epsilon_{95\%\mathrm{CL}}$ & $\epsilon^    {\mathcal{U}}_{99\%\mathrm{CL}}$ & $t$ (s) & $t^{\mathrm{null}}$ (s) \\
	\midrule
	$t_{\mathrm{SW}}$ & $0.20487_{-0.048}^{+0.042}$ & $0.2434_{-0.035}^{+0.032}$ & ${\mathbf{877}}$ & ${\mathbf{123}}$ \\
	$t_{\overline{\mathrm{KS}}}$ & $0.18018_{-0.035}^{+0.024}$ & $0.19884_{-0.027}^{+0.018}$ & $25630$ & $1913$ \\
	$t_{\mathrm{SKS}}$ & $0.14529_{-0.056}^{+0.04}$ & $0.1719_{-0.048}^{+0.035}$ & $42277$ & $4383$ \\
	$t_{\mathrm{FGD}}$ & ${\mathbf{0.13545_{-0.032}^{+0.014}}}$ & ${\mathbf{0.15299_{-0.012}^{+0.015}}}$ & $12782$ & $1787$ \\
	$t_{\mathrm{MMD}}$ & $0.45177_{-0.091}^{+0.066}$ & $0.52083_{-0.047}^{+0.05}$ & $56078$ & $3504$ \\
	\bottomrule
\end{tabular}

%% file: Tables/particle_features_50K_results_table.tex
\begin{tabular}{l|llr|llr}
	\toprule
	\multicolumn{7}{c}{{\bf Particle features with $\mathbf{n=m=5\cdot 10^{4}}$}} \\
	\toprule
	\multicolumn{1}{c}{} & \multicolumn{3}{c}{$\mu$-deformation} & \multicolumn{3}{c}{$\Sigma_{ii}$-deformation} \\
	Statistic & $\epsilon_{95\%\mathrm{CL}}$ & $\epsilon_{99\%\mathrm    {CL}}$ & $t$ (s) & $\epsilon_{95\%\mathrm{CL}}$ & $\epsilon_{99\%\mathrm{CL}}$ & $t$ (s) \\
	\midrule
	$t_{\mathrm{SW}}$ & $0.02633_{-0.013}^{+0.0098}$ & $0.03714_{-0.0097}^{+0.0084}$ & ${\mathbf{849}}$ & $0.02913_{-0.0079}^{+0.012}$ & $0.04108_{-0.011}^{+0.0093}$ & ${\mathbf{824}}$ \\
	$t_{\overline{\mathrm{KS}}}$ & ${\mathbf{0.0_{-0.0}^{+0.0045}}}$ & ${\mathbf{0.00771_{-0.0068}^{+0.0022}}}$ & $49525$ & ${\mathbf{0.0_{-0.0}^{+0.013}}}$ & ${\mathbf{0.01904_{-0.011}^{+0.0086}}}$ & $55017$ \\
	$t_{\mathrm{SKS}}$ & $0.01592_{-0.0061}^{+0.0046}$ & $0.02334_{-0.0059}^{+0.0058}$ & $17572$ & $0.02735_{-0.01}^{+0.0049}$ & $0.03362_{-0.0071}^{+0.0081}$ & $24987$ \\
	$t_{\mathrm{FGD}}$ & $0.04749_{-0.024}^{+0.012}$ & $0.06462_{-0.013}^{+0.013}$ & $30820$ & $0.04004_{-0.012}^{+0.017}$ & $0.0556_{-0.016}^{+0.018}$ & $25551$ \\
	$t_{\mathrm{MMD}}$ & $0.1396_{-0.065}^{+0.1}$ & $0.21274_{-0.065}^{+0.071}$ & $18527$ & $0.06988_{-0.031}^{+0.048}$ & $0.0986_{-0.036}^{+0.037}$ & $33217$ \\
	\toprule
	\multicolumn{1}{c}{} & \multicolumn{3}{c}{$\Sigma_{i\neq j}$-deformation} & \multicolumn{3}{c}{$\rm{pow}_{+}$-deformation} \\
	Statistic & $\epsilon_{95\%\mathrm{CL}}$ & $\epsilon_{99\%\mathrm{CL}}$ & $t$ (s) & $\epsilon_{95\%\mathrm{CL}}$ & $\epsilon^{\rm   {pow}_{+}}_{99\%\mathrm{CL}}$ & $t$ (s) \\
	\midrule
	$t_{\mathrm{SW}}$ & $0.04883_{-0.015}^{+0.012}$ & $0.06979_{-0.016}^{+0.0097}$ & ${\mathbf{1966}}$ & $0.02745_{-0.0075}^{+0.011}$ & $0.03872_{-0.011}^{+0.0088}$ & ${\mathbf{806}}$ \\
	$t_{\overline{\mathrm{KS}}}$ & $0.99933_{-0.014}^{+0.0085}$ & $1.01732_{-0.0079}^{+0.006}$ & $11225$ & ${\mathbf{0.0_{-0.0}^{+0.0066}}}$ & ${\mathbf{0.01141_{-0.011}^{+0.0073}}}$ & $46010$ \\
	$t_{\mathrm{SKS}}$ & $0.04267_{-0.012}^{+0.018}$ & $0.06018_{-0.013}^{+0.014}$ & $29568$ & $0.03594_{-0.016}^{+0.021}$ & $0.05069_{-0.014}^{+0.011}$ & $24821$ \\
	$t_{\mathrm{FGD}}$ & ${\mathbf{0.02641_{-0.012}^{+0.0058}}}$ & ${\mathbf{0.03966_{-0.0079}^{+0.006}}}$ & $28408$ & $0.02459_{-0.012}^{+0.013}$ & $0.03501_{-0.012}^{+0.013}$ & $25798$ \\
	$t_{\mathrm{MMD}}$ & $0.25965_{-0.086}^{+0.068}$ & $0.35327_{-0.056}^{+0.073}$ & $16061$ & $0.02054_{-0.0071}^{+0.014}$ & $0.02657_{-0.0091}^{+0.013}$ & $26195$ \\
	\toprule
	\multicolumn{1}{c}{} & \multicolumn{3}{c}{$\rm{pow}_{-}$-deformation} & \multicolumn{3}{c}{$\mathcal{N}$-deformation} \\
	Statistic & $\epsilon_{95\%\mathrm{CL}}$ & $\epsilon^  {\rm{pow}_{-}}_{99\%\mathrm{CL}}$ & $t$ (s) & $\epsilon_{95\%\mathrm{CL}}$ & $\epsilon^    {\mathcal{N}}_{99\%\mathrm{CL}}$ & $t$ (s) \\
	\midrule
	$t_{\mathrm{SW}}$ & $0.02745_{-0.014}^{+0.011}$ & $0.03872_{-0.01}^{+0.0088}$ & ${\mathbf{809}}$ & $0.10733_{-0.026}^{+0.022}$ & $0.13357_{-0.016}^{+0.016}$ & ${\mathbf{691}}$ \\
	$t_{\overline{\mathrm{KS}}}$ & ${\mathbf{0.0_{-0.0}^{+0.0095}}}$ & ${\mathbf{0.01323_{-0.0085}^{+0.0069}}}$ & $45685$ & ${\mathbf{0.0656_{-0.053}^{+0.016}}}$ & ${\mathbf{0.08707_{-0.016}^{+0.013}}}$ & $7484$ \\
	$t_{\mathrm{SKS}}$ & $0.03777_{-0.017}^{+0.015}$ & $0.04837_{-0.015}^{+0.014}$ & $15966$ & $0.08456_{-0.013}^{+0.013}$ & $0.09935_{-0.01}^{+0.0089}$ & $18276$ \\
	$t_{\mathrm{FGD}}$ & $0.02241_{-0.01}^{+0.019}$ & $0.0353_{-0.011}^{+0.019}$ & $26549$ & $0.14608_{-0.038}^{+0.034}$ & $0.1758_{-0.021}^{+0.023}$ & $23330$ \\
	$t_{\mathrm{MMD}}$ & $0.02077_{-0.01}^{+0.017}$ & $0.02939_{-0.0089}^{+0.016}$ & $20263$ & $0.33827_{-0.089}^{+0.088}$ & $0.37964_{-0.073}^{+0.091}$ & $19908$ \\
	\toprule
	\multicolumn{1}{c}{} & \multicolumn{3}{c}{$\mathcal{U}$-deformation} & \multicolumn{3}{c}{Timing} \\
	Statistic & $\epsilon_{95\%\mathrm{CL}}$ & $\epsilon^    {\mathcal{U}}_{99\%\mathrm{CL}}$ & $t$ (s) & $t^{\mathrm{null}}$ (s) \\
	\midrule
	$t_{\mathrm{SW}}$ & $0.1889_{-0.046}^{+0.038}$ & $0.2351_{-0.04}^{+0.028}$ & ${\mathbf{625}}$ & ${\mathbf{150}}$ \\
	$t_{\overline{\mathrm{KS}}}$ & ${\mathbf{0.10693_{-0.086}^{+0.022}}}$ & ${\mathbf{0.14193_{-0.019}^{+0.021}}}$ & $13565$ & $2126$ \\
	$t_{\mathrm{SKS}}$ & $0.14453_{-0.028}^{+0.02}$ & $0.17795_{-0.023}^{+0.0057}$ & $17723$ & $4818$ \\
	$t_{\mathrm{FGD}}$ & $0.25168_{-0.065}^{+0.053}$ & $0.30289_{-0.036}^{+0.04}$ & $23243$ & $7351$ \\
	$t_{\mathrm{MMD}}$ & $0.58039_{-0.15}^{+0.17}$ & $0.6876_{-0.14}^{+0.14}$ & $25557$ & $3880$ \\
	\bottomrule
\end{tabular}

%% file: Tables/particle_features_50K_scaled_results_table.tex
\begin{tabular}{l|llr|llr}
	\toprule
	\multicolumn{7}{c}{{\bf Scaled Particle features with $\mathbf{n=m=5\cdot 10^{4}}$}} \\
	\toprule
	\multicolumn{1}{c}{} & \multicolumn{3}{c}{$\mu$-deformation} & \multicolumn{3}{c}{$\Sigma_{ii}$-deformation} \\
	Statistic & $\epsilon_{95\%\mathrm{CL}}$ & $\epsilon_{99\%\mathrm    {CL}}$ & $t$ (s) & $\epsilon_{95\%\mathrm{CL}}$ & $\epsilon_{99\%\mathrm{CL}}$ & $t$ (s) \\
	\midrule
	$t_{\mathrm{SW}}$ & $0.01334_{-0.0046}^{+0.0038}$ & $0.01815_{-0.0029}^{+0.0037}$ & ${\mathbf{1116}}$ & $0.0166_{-0.0063}^{+0.0059}$ & $0.02125_{-0.0034}^{+0.006}$ & ${\mathbf{1079}}$ \\
	$t_{\overline{\mathrm{KS}}}$ & ${\mathbf{0.0_{-0.0}^{+0.0045}}}$ & ${\mathbf{0.00771_{-0.0049}^{+0.0022}}}$ & $58835$ & ${\mathbf{0.0_{-0.0}^{+0.013}}}$ & $0.01904_{-0.011}^{+0.0086}$ & $62555$ \\
	$t_{\mathrm{SKS}}$ & $0.01275_{-0.0043}^{+0.0034}$ & $0.01734_{-0.0028}^{+0.0036}$ & $18356$ & $0.02131_{-0.0073}^{+0.007}$ & $0.02899_{-0.0047}^{+0.006}$ & $26542$ \\
	$t_{\mathrm{FGD}}$ & $0.01627_{-0.006}^{+0.003}$ & $0.02025_{-0.0047}^{+0.0024}$ & $39057$ & $0.01469_{-0.0057}^{+0.0034}$ & ${\mathbf{0.01805_{-0.0051}^{+0.0043}}}$ & $27175$ \\
	$t_{\mathrm{MMD}}$ & $0.01613_{-0.0058}^{+0.0049}$ & $0.02141_{-0.0035}^{+0.0032}$ & $22841$ & $0.01606_{-0.0066}^{+0.0074}$ & $0.02089_{-0.0061}^{+0.0055}$ & $33730$ \\
	\toprule
	\multicolumn{1}{c}{} & \multicolumn{3}{c}{$\Sigma_{i\neq j}$-deformation} & \multicolumn{3}{c}{$\rm{pow}_{+}$-deformation} \\
	Statistic & $\epsilon_{95\%\mathrm{CL}}$ & $\epsilon_{99\%\mathrm{CL}}$ & $t$ (s) & $\epsilon_{95\%\mathrm{CL}}$ & $\epsilon^{\rm   {pow}_{+}}_{99\%\mathrm{CL}}$ & $t$ (s) \\
	\midrule
	$t_{\mathrm{SW}}$ & $0.03128_{-0.011}^{+0.0068}$ & $0.04152_{-0.0067}^{+0.0061}$ & ${\mathbf{1424}}$ & $0.01894_{-0.0072}^{+0.0055}$ & $0.02425_{-0.0039}^{+0.0068}$ & ${\mathbf{1006}}$ \\
	$t_{\overline{\mathrm{KS}}}$ & $0.99134_{-0.0078}^{+0.016}$ & $1.01532_{-0.004}^{+0.008}$ & $11987$ & ${\mathbf{0.0_{-0.0}^{+0.0066}}}$ & ${\mathbf{0.01141_{-0.011}^{+0.0073}}}$ & $49091$ \\
	$t_{\mathrm{SKS}}$ & $0.03809_{-0.013}^{+0.011}$ & $0.0515_{-0.0083}^{+0.011}$ & $27313$ & $0.03552_{-0.013}^{+0.0055}$ & $0.04366_{-0.0066}^{+0.01}$ & $15487$ \\
	$t_{\mathrm{FGD}}$ & ${\mathbf{0.0026_{-0.00089}^{+0.00076}}}$ & ${\mathbf{0.00345_{-0.00076}^{+0.00052}}}$ & $33338$ & $0.01534_{-0.0062}^{+0.0052}$ & $0.01886_{-0.0055}^{+0.0045}$ & $24241$ \\
	$t_{\mathrm{MMD}}$ & $0.01919_{-0.0079}^{+0.011}$ & $0.02614_{-0.0065}^{+0.0089}$ & $20604$ & $0.01896_{-0.008}^{+0.0074}$ & $0.02428_{-0.0071}^{+0.0068}$ & $27198$ \\
	\toprule
	\multicolumn{1}{c}{} & \multicolumn{3}{c}{$\rm{pow}_{-}$-deformation} & \multicolumn{3}{c}{$\mathcal{N}$-deformation} \\
	Statistic & $\epsilon_{95\%\mathrm{CL}}$ & $\epsilon^  {\rm{pow}_{-}}_{99\%\mathrm{CL}}$ & $t$ (s) & $\epsilon_{95\%\mathrm{CL}}$ & $\epsilon^    {\mathcal{N}}_{99\%\mathrm{CL}}$ & $t$ (s) \\
	\midrule
	$t_{\mathrm{SW}}$ & $0.01909_{-0.0077}^{+0.0073}$ & $0.02693_{-0.0068}^{+0.0061}$ & ${\mathbf{1006}}$ & $0.10868_{-0.017}^{+0.02}$ & $0.1277_{-0.022}^{+0.011}$ & ${\mathbf{886}}$ \\
	$t_{\overline{\mathrm{KS}}}$ & ${\mathbf{0.0_{-0.0}^{+0.0095}}}$ & ${\mathbf{0.01323_{-0.0085}^{+0.0069}}}$ & $45323$ & $0.0656_{-0.049}^{+0.016}$ & $0.08707_{-0.019}^{+0.013}$ & $22186$ \\
	$t_{\mathrm{SKS}}$ & $0.0356_{-0.013}^{+0.0093}$ & $0.04726_{-0.011}^{+0.007}$ & $22261$ & $0.10733_{-0.017}^{+0.022}$ & $0.13357_{-0.026}^{+0.016}$ & $24344$ \\
	$t_{\mathrm{FGD}}$ & $0.01543_{-0.0065}^{+0.007}$ & $0.01852_{-0.0042}^{+0.0068}$ & $24968$ & ${\mathbf{0.04853_{-0.0075}^{+0.0071}}}$ & ${\mathbf{0.05702_{-0.006}^{+0.0051}}}$ & $24273$ \\
	$t_{\mathrm{MMD}}$ & $0.01859_{-0.0081}^{+0.0085}$ & $0.02501_{-0.0064}^{+0.0083}$ & $27960$ & $0.26953_{-0.052}^{+0.035}$ & $0.30333_{-0.011}^{+0.029}$ & $19782$ \\
	\toprule
	\multicolumn{1}{c}{} & \multicolumn{3}{c}{$\mathcal{U}$-deformation} & \multicolumn{3}{c}{Timing} \\
	Statistic & $\epsilon_{95\%\mathrm{CL}}$ & $\epsilon^    {\mathcal{U}}_{99\%\mathrm{CL}}$ & $t$ (s) & $t^{\mathrm{null}}$ (s) \\
	\midrule
	$t_{\mathrm{SW}}$ & $0.19116_{-0.03}^{+0.035}$ & $0.22462_{-0.039}^{+0.02}$ & ${\mathbf{774}}$ & ${\mathbf{133}}$ \\
	$t_{\overline{\mathrm{KS}}}$ & $0.10693_{-0.086}^{+0.022}$ & $0.14193_{-0.019}^{+0.021}$ & $10646$ & $1972$ \\
	$t_{\mathrm{SKS}}$ & $0.19116_{-0.03}^{+0.035}$ & $0.22462_{-0.039}^{+0.02}$ & $16154$ & $4379$ \\
	$t_{\mathrm{FGD}}$ & ${\mathbf{0.08969_{-0.016}^{+0.006}}}$ & ${\mathbf{0.10215_{-0.016}^{+0.0042}}}$ & $21825$ & $6689$ \\
	$t_{\mathrm{MMD}}$ & $0.48398_{-0.088}^{+0.032}$ & $0.5512_{-0.058}^{+0.022}$ & $14676$ & $3605$ \\
	\bottomrule
\end{tabular}